%% file: acl2023.tex
\pdfoutput=1

\documentclass[11pt]{article}

\usepackage{ACL2023}

\usepackage{pgf}
\usepackage{import}
\usepackage{tikz}
\usepackage{bm}
\usepackage[outline]{contour}
\usepackage{float}
\usepackage{amsthm}

\newcommand\taskname[1]{\tikz[baseline=-1mm] \node[white,draw,inner sep=1mm,fill=gray,rounded corners] {\small\sffamily\uppercase{#1}};}

\usepackage{times}
\usepackage{latexsym}
\usepackage{tikz}
\usetikzlibrary{calc}
\input{math_commands.tex}

\usepackage{microtype}
\usepackage{mathtools}
\usepackage{bbm}
\usepackage{graphicx}
\usepackage[capitalize,noabbrev]{cleveref}
\usepackage{subfigure}
\usepackage{booktabs} %
\usepackage[noend]{algpseudocode}
\usepackage{amsthm,amssymb,amsmath}
\usepackage{amsmath}
\usepackage{glossaries}
\usepackage{listings}
\usepackage{wrapfig}
\usepackage{xspace}
\usepackage{stfloats}
\usepackage{algorithm2e}
\usepackage{enumitem}
\usepackage[font=small]{caption}

\DeclareMathOperator*{\Ex}{\mathbb{E}}

\newcommand{\cbox}[3]{%
    #2%
}

\newcommand\pto{\mathrel{\ooalign{\hfil$\mapstochar\mkern5mu$\hfil\cr$\to$\cr}}}

\usepackage[T1]{fontenc}

\usepackage[utf8]{inputenc}

\usepackage{microtype}

\usepackage{inconsolata}

\newcommand{\speaker}{\textsc{speaker}\xspace}
\newcommand{\listener}{\textsc{listener}\xspace}
\newcommand{\xspeaker}{\textsf{S}}
\newcommand{\xlistener}{\textsf{L}}
\newcommand{\object}{m}
\newcommand{\utterance}{u}

\newcommand\colorPatch[1]{%
\definecolor{temp}{HTML}{#1}%
\tikz[baseline=1mm] \filldraw[draw=black!50,very thin,anchor=base,rounded corners=3pt,fill=temp] (0,0) rectangle (1.6, .65);%
}

\newcommand{\ourmethod}{\textsc{ReCo}\xspace}
\newcommand{\ourmethodfull}{Regularized Conventions\xspace}
  
\newcommand{\colorContext}[4]{
  \framebox{\colorPatch{#1}} & \colorPatch{#2} & \colorPatch{#3} & \emph{#4}}

\renewcommand\vec[1]{\bm{#1}}

\title{
\ourmethodfull: \\ Equilibrium Computation as a Model of Pragmatic Reasoning
}

\author{Athul Paul Jacob \\
  \texttt{apjacob@mit.edu} \\\And
  Gabriele Farina \\
  \texttt{gfarina@mit.edu} \\\And
  Jacob Andreas \\
  \texttt{jda@mit.edu}}

\begin{document}

\maketitle

\begin{abstract}
  We present a model of pragmatic language understanding, where utterances are produced and understood by searching for \emph{regularized equilibria} of signaling games.
  In this model (which we call \ourmethod, for \ourmethodfull), speakers and listeners search for contextually appropriate utterance--meaning mappings that are both close to game-theoretically optimal conventions and close to a shared, ``default'' semantics. By characterizing pragmatic communication as equilibrium search, we obtain principled sampling algorithms and formal guarantees about the trade-off between communicative success and naturalness.
  Across several datasets capturing real and idealized human judgments about pragmatic implicatures, \ourmethod matches or improves upon predictions made by best response and rational speech act models of language understanding.
\end{abstract}

\begin{figure*}[b]
  \centering
  \includegraphics[width=\textwidth]{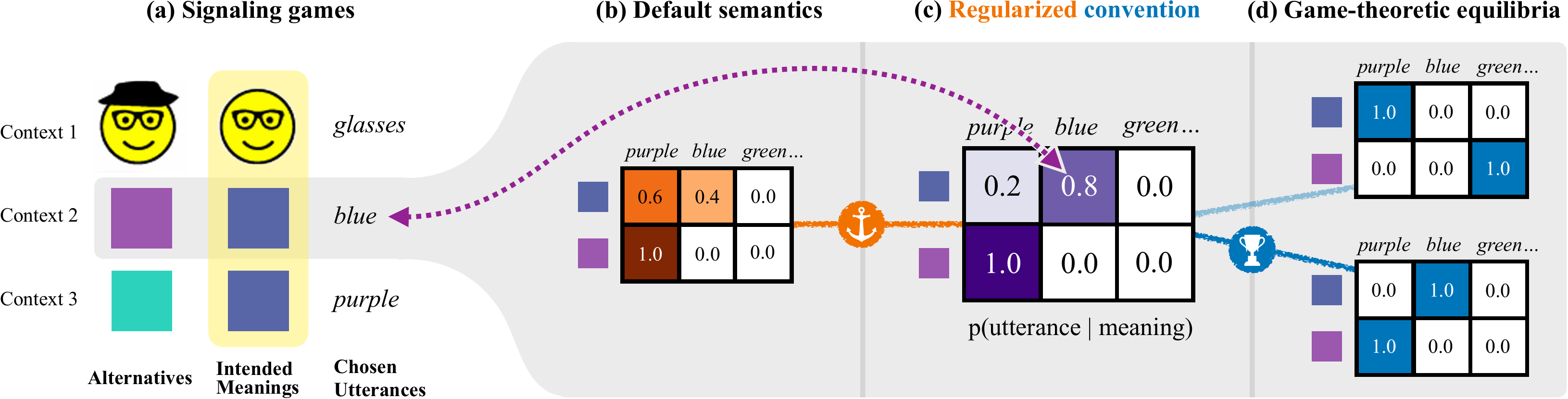}
  \caption{The \ourmethod model. To communicate (or resolve) an intended meaning from a set of possibilities \textbf{(a)}, language users search for a joint distributions over utterances and interpretations that is close to a distribution encoding ``default semantics'' \textbf{(b)} and close to some (game-theoretically) optimal signaling convention \textbf{(d)}. The resulting ``regularized conventions'' \textbf{(c)} predict human judgments on a variety of implicature tasks.}
  \label{fig:teaser}
\end{figure*}

\input{text/0_introduction}

\input{text/1_background}
\input{text/2_methods}

\input{text/3_experiments}

\input{text/4_conclusion}

\section*{Acknowledgements}

Thanks to Jennifer Hu for helpful suggestions about experiment design. This work was supported by the National Science Foundation under grant IIS-2212310.


\input{acl2023.bbl}
\end{document}

%% file: math_commands.tex
\usepackage{amsmath,amsfonts,bm}

\def\eqref#1{equation~\ref{#1}}

\def\1{\bm{1}}

\def\vc{{\bm{c}}}

\def\vp{{\bm{p}}}

\def\mL{{\mathbf{L}}}

\def\mP{{\mathbf{P}}}

\def\mS{{\mathbf{S}}}

\DeclareMathAlphabet{\mathsfit}{\encodingdefault}{\sfdefault}{m}{sl}
\SetMathAlphabet{\mathsfit}{bold}{\encodingdefault}{\sfdefault}{bx}{n}

\newcommand{\defeq}{\coloneqq}

%% file: text/0_introduction.tex
\section{Introduction}

Meaning in language is fluid: speakers can use the word \emph{blue} to pick out a color that in other contexts would be described as \emph{purple}, or identify a friend as \emph{the one with glasses} even in a room in which everyone is wearing glasses (\Cref{fig:teaser}). Such context-dependent meanings can arise as \textbf{conventions} within groups of language-users communicating repeatedly to solve a shared task \cite{hawkins2017convention}. But remarkably, they can also arise \emph{without any interaction at all}, between pairs of language users who share only common knowledge of words' default meanings.

What makes this kind of context-dependent language use possible?
Almost all existing computational models of pragmatics are implemented as \textbf{iterated response} procedures, in which listeners interpret utterances by reasoning about the possible intentions of less-sophisticated speakers (or vice-versa) \cite{golland2010game, degen2023rational}. These models have been successful at explaining a number of aspects of pragmatic language use. But they can be challenging to fit to real data: because they specify behavior in terms of an algorithm that speakers and listeners implement, rather than an objective that they optimize, iterated response models can be highly sensitive to low-level details of initialization and runtime.

We present an alternative model of pragmatic understanding based on \textbf{equilibrium search} rather than iterated response.
In this model (which we call \ourmethodfull, or \ourmethod), speakers and listeners solve communicative tasks like the ones in \Cref{fig:teaser} by searching for utterance--meaning mappings that are both close some to a (game-theoretically) optimal convention and close to a set of default semantics. In \Cref{fig:teaser}, for example, this ``regularized convention'' assigns high probability to the use of \emph{blue} to signal the intended color, and low (but nonzero) probability to the use of \emph{purple} instead. This strategy is both close to one of many optimal strategies (in which every utterance arbitrarily, but uniquely, picks out one color), and close to color terms' standard interpretation (in which the target color is improbably, but not impossibly, described as \emph{blue}).

\ourmethod is by no means the first application of game-theoretic tools to model pragmatic language understanding \cite{parikh2000communication, franke2013game, jager2012game}---many iterated response models (e.g. \citealp{franke2009interpretation}) also have a game-theoretic foundation. But by leveraging recently developed algorithmic tools for computing regularized equilibria of games, \ourmethod makes it possible to efficiently learn models of pragmatic communication from data, and to provide formal guarantees about their communicative success and deviation from default semantic. The algorithms that compute these equilibria turn out to have a very similar structure to probabilistic iterated response methods \cite{frank2012predicting}, offering a possible bridge between algorithmic characterizations of pragmatic reasoning  and \ourmethod's optimality-based characterization.

Most importantly, \ourmethod gives a good fit to human data: on classic exemplars of
pragmatic implicature, reference tasks eliciting graded human judgments, and tasks featuring perceptually complex meaning spaces, its predictions match or modestly outperform standard iterated response models. These results highlight the usefulness of modern game-theoretic tools in modeling language production and comprehension.

%% file: text/1_background.tex
\section{Background and Preliminaries}
\label{sec:background}

Consider again the example in \Cref{fig:teaser}. We want to understand the process by which a \speaker might use \emph{blue} to refer to the second color in the second row, and by which a \listener might resolve it correctly.

\subsection{Signaling Games}
\label{ssec:signaling}
This problem has often been formulated as a signalling game \cite{lewis1971convention}, which features two players: the \speaker and the \listener. In this game, a \textbf{target meaning} (representing a communicative need) is first sampled from a space of possible meanings $\object \in M$ with probability $p(m)$. To communicate this meaning, the \speaker produces an \textbf{utterance} according to a policy $\pi_\xspeaker(\utterance \mid \object)$. Finally, the \listener produces an \textbf{interpretation} according to a policy $\pi_\xlistener(\object' \mid \utterance)$. 

Informally, communication is successful if the \listener's interpretation is the same as the \speaker's intended meaning.
More formally (and somewhat more generally), we may define communicative success in terms of \textbf{rewards}.
Consider any (meaning, utterance, interpretation) combination $(\object,\utterance,\object')$. The \speaker's reward (or ``payoff'') $r_\xspeaker(\object,\utterance,\object')$ in this interaction is the sum of:
\begin{itemize}
    \item a \emph{cost} $-c(u)$ that the \speaker incurs for producing utterance $u$ (all else equal, they may for example prefer short utterances); and
    \item a \emph{success measure}, equal to 1 only when $\object'$ matches the target $\object$, that is, $\mathbf{1}[\object'=\object]$.
\end{itemize}
Together,
\[
    r_\xspeaker(\object,\utterance,\object') \defeq -c(u) + \mathbf{1}[\object'=\object].
\]
Most models assume that the \listener's reward $r_\xlistener(\object,\utterance,\object')$ depends only on communicative success:
\[
    r_\xlistener(\object,\utterance,\object') = \mathbf{1}[\object'=\object].
\]

Having specified rewards for all interactions, the \emph{expected utility} of each player given policies $(\pi_\xspeaker, \pi_\xlistener)$ for the \speaker and \listener respectively is defined as the expected payoff when the meanings $\object$ are sampled from a prior distribution $p(\object)$, and agents sample from their policies:
\begin{equation}\label{eq:ubar}
    \bar u_i(\pi_\xspeaker, \pi_\xlistener) \defeq \Ex_{\substack{\object \sim p \\ u \sim \pi_\xspeaker(\cdot \mid \object) \\ \object' \sim \pi_\xlistener(\cdot \mid \utterance)}} p_i(\object,\utterance,\object')
\end{equation}
for $i \in \{\xspeaker, \xlistener\}$.

\subsection{Computing Policies for Signaling Games}
\label{ssec:solving}

How should a \speaker and \listener communicate to maximize the probability of success? We call a pair of policies, for the \speaker and for the \listener a \emph{Nash equilibrium} if nobody wants to deviate.
Notice that there may in general be multiple such policies: in \Cref{fig:teaser}, for example, there is one equilibrium policy in which the intended meaning is called \emph{blue} and the alternative is called \emph{purple}, but another equilibrium policy in which the former is called \emph{purple} and the latter called \emph{green} (in clear violation of those words' standard use in English!). When a \speaker and \listener communicate for the first time, how can they ensure that their policies are compatible?

\paragraph{Iterated response methods}
A popular family of approaches attempt to ensure communicative success \emph{algorithmically}. These approaches typically begin from an assumption that \speaker{}s' and \listener{}s' common knowledge of language consists of a \textbf{literal semantics} (which assigns context-independent meanings to utterances). Agents then derive policies by computing behaviors likely to be successful given an interlocutor communicating literally, or given an interlocutor themself attempting to respond to a literal communicator.
Approaches in this family involve (Iterated) Best Response (I)BR \cite{jager2007game,franke2009signal,franke2009interpretation} and Rational Speech Acts (RSA) \cite{frank2012predicting}. 

(I)BR is an iterative algorithm in which speakers (listeners) alternatingly compute the highest-utility action keeping the listener's (speaker's) policy fixed:
\begin{align*}
    \pi^{(t+1)}_\xlistener(\object' \mid \utterance) &= \mathbf{1}\left[\object' = \arg\max_\object \pi^{(t)}_\xspeaker(\utterance \mid \object)\right] \\
    \pi^{(t+1)}_\xspeaker(\utterance \mid \object) &= \mathbf{1}\left[\utterance = \arg\max_{\utterance'} \pi^{(t)}_\xlistener(\object \mid \utterance')\right]
\end{align*}
RSA frames communication as one in which Bayesian listeners and speakers reason recursively about each other's beliefs in order to choose utterances and meanings:
\begin{align*}
    \pi^{(t)}_\xlistener(\object \mid \utterance) &\propto \pi^{(t)}_\xspeaker(\utterance \mid \object) \cdot p(\object) \\
    \pi^{(t)}_\xspeaker(\utterance \mid \object) &\propto \big({\pi^{(t)}_\xlistener(\object \mid \utterance) / c(\utterance)}\big)^{\alpha}
\end{align*}
In both approaches, ``good'' policies are obtained by assuming that speakers and listeners will run a specific algorithm from a specific starting point (rather than generically optimizing a known objective). As a result, a key feature of both algorithms is its sensitivity to the choice of initial ($t=0$) policy; their convergence behavior remains poorly understood.

\paragraph{piKL-Hedge and regularized no-regret dynamics}
A set of principled techniques for solving games comes from the vast literature of online optimization and learning in games. Hedge~\cite{littlestone1994weighted,freund1997decision} is a popular iterative algorithm in this family that converges to a coarse correlated equilibrium \cite{hannan1957approximation} and to a Nash equilibrium in the special case of two-player zero-sum games. However, in many cases, the equilibria that is of interest is one that is close to certain \textbf{anchor policies} -- which Hedge does not guarantee.

In order to sidestep this issue while retaining the appealing properties of learning in games, \citet{jacob2022modeling} introduced \textbf{piKL-Hedge}. piKL-Hedge has been used in the context of board games, like Diplomacy \cite{meta2022human, bakhtin2022mastering} to find equilibria that are close to human imitation learned anchor policies. Recently, piKL-Hedge has been used in the context of language models, with the objective of increasing consensus between disriminative and generative approaches to language model generation \cite{jacob2023consensus}.

%% file: text/2_methods.tex
\section{Our Approach: Pragmatic Inference as Regularized Equilibrium Search}
\label{sec:model}

Building on this past work, the key idea underlying the \ourmethod model is to define an objective that makes it possible to directly optimize for both communicative success and adherence to shared background knowledge of language.
As noted in \cref{ssec:solving}, simply searching for high-utility equilibria of signaling games is unlikely to predict the behavior of human language users, or result in successful communication with new interlocutors: instead, we must guide inference toward policies that \emph{look like natural language}. In \ourmethod, we do so by optimizing utilities of the following form:
\begin{align*}
    \tilde u_\xspeaker(\pi_\xspeaker, \pi_\xlistener)  & \defeq \bar u_\xspeaker(\pi_\xspeaker, \pi_\xlistener) -\lambda_\xspeaker \cdot \mathrm{D}_\mathrm{KL}(\pi_\xspeaker \,\|\, \tau_\xspeaker),     \\
    \tilde u_\xlistener(\pi_\xspeaker, \pi_\xlistener) & \defeq \bar u_\xlistener(\pi_\xspeaker, \pi_\xlistener) -\lambda_\xlistener \cdot \mathrm{D}_\mathrm{KL}(\pi_\xlistener \,\|\, \tau_\xlistener).
\end{align*}
Here $\tau_\xspeaker$ and $\tau_\xlistener$ represent the \speaker{}'s and \listener{}'s prior knowledge of language (independent of any specific communicative goal or context). We refer to these policies as the \textbf{default semantics} in the language used for communication. They play a similar role to the literal semantics used by RSA and other iterated response models. But here, we need not assume that they correspond specifically to literal semantics---instead, they model agents' prior expectations about how utterances are likely to be produced and interpreted in general by pragmatic language users.

The regularization parameters $\lambda_\xspeaker$ and $\lambda_\xlistener$ control the amount of regularization towards the default semantics $\tau_\xspeaker, \tau_\xlistener$. When the value of $\lambda_i$ is large, Player $i\in\{\xspeaker,\xlistener\}$ will be regularized towards only considering policies extremely close to $\tau_i$; conversely, when $\lambda_i$ is close to zero, the player will not be penalized for adopting semantics that differ significantly from $\tau_i$.

\subsection{Notation and representation of policies}

In this subsection, we lay down the notation and representation details for the policies produced by our algorithm. 
Each agent's \emph{policy} consists of a mapping from that player's observations to a distribution over actions. In order to provide a compact description of the algorithm, as well as an efficient vectorized implementation, we represent such a mapping as a row-stochastic matrix, with rows indexed by observations and columns indexed by actions. For the
\speaker{}, the set of observations coincides with the set of meanings available in a given communicative context, and the set of actions coincides with the set of possible utterances. For the \listener{}, observations are utterances and actions are meanings. See \cref{fig:scalars} for examples. We denote with $\mathbf{S}^{(t)} \in \mathbb{R}^{M \times U}$ the policy of the speaker at time $t$, and with $\mathbf{L}^{(t)} \in \mathbb{R}^{U \times M}$ that of the listener represented in this matrix form. Similar, we will also represent the anchor policies (\emph{i.e.}, default semantics) $\tau_\xspeaker,\tau_\xlistener$ in this representation as matrices $\vec{\tau}_\xspeaker \in \mathbb{R}^{M \times U}$ and $\vec{\tau}_\xlistener \in \mathbb{R}^{U\times M}$. Instances of such matrix objects can be seen in \Cref{fig:scalars}.

\subsection{\ourmethod: Computation of Approximate Convention-Regularized Equilibria}

Given the regularized utilities $\tilde u_\xspeaker$ and $u_\xlistener$ defined above, we use the piKL-Hedge algorithm \cite{jacob2022modeling} to progressively refine the \speaker's and \listener's policy toward equilibrium (in the sense of \cref{ssec:solving}).
Intuitively, piKL-Hedge performs a variant of projected gradient ascent in the geometry of entropic regularization where projections are equivalent to softmax (normalized exponentiation). In order to apply piKL-Hedge, we start by computing the gradients of the unregularized utility functions $\bar u_\xspeaker, \bar u_\xlistener$ defined in (\ref{eq:ubar}).

Let $\vp \in \mathbb{R}^M$ be the vector whose entries correspond to $p(\object)$, the prior distribution over meanings. Similarly, we let $\vc \in \mathbb{R}^U$ denote the vector of utterance costs. Finally, let $\mP \in \mathbb{R}^{M\times M}$ be the diagonal matrix whose diagonal equals $\vp$. With this notation, it is straightforward to verify that the gradient of the unregularized utility function $\bar u_\xspeaker$ of the \speaker player, as a function of the matrix-form policies $\mS, \mL$, is given by
\begin{equation}\label{eq:nabla S}
    \nabla_\mS (\mL) \defeq -\vp \vc^\top + \mP \mL^\top \in \mathbb{R}^{M\times U}.
\end{equation}
Similarly, for the \listener player we have
\begin{equation}\label{eq:nabla L}
    \nabla_\mL (\mS) \defeq \mS^\top \mP \in \mathbb{R}^{U\times M}.
\end{equation}
With the above gradients, piKL-Hedge prescribes the following dynamics: first, at time $0$, set $\bar\mS^{(0)}=\bar\mL^{(0)} \defeq \vec{0}$; then, at each time $t \ge 0$, the next policy $\mS^{(t+1)},\mL^{(t+1)}$ is chosen according to the update rules:
\begin{align*}
    \mathbf{S}^{(t+1)}       & \overset{\text{row}}{\propto} \exp\left\{ \frac{\nabla_{\mathbf{S}}(\bar{\mathbf{L}}^{(t)}) + \lambda_\xspeaker \log \vec{\tau}_\xspeaker}{1/(\eta_\xspeaker t) + \lambda_\xspeaker} \right\}, \\
    \mathbf{L}^{(t+1)}       & \overset{\text{row}}{\propto} \exp\left\{ \frac{\nabla_{\mathbf{L}}(\bar{\mathbf{S}}^{(t)})^\top + \lambda_\xlistener \log \vec{\tau}_\xlistener}{1/(\eta_\xlistener t) + \lambda_\xlistener} \right\}, \\
    \bar{\mathbf{S}}^{(t+1)} & = \frac{t}{t+1}\bar{\mathbf{S}}^{(t)} + \frac{1}{t+1}\mathbf{S}^{(t+1)},                                                                      \\
    \bar{\mathbf{L}}^{(t+1)} & = \frac{t}{t+1}\bar{\mathbf{L}}^{(t)} + \frac{1}{t+1}\mathbf{L}^{(t+1)},
\end{align*}
where $\overset{\text{row}}{\propto}$ denotes row-wise proportionality and exponentiation is intended elementwise. 

piKL-Hedge dynamics have
strong guarantees, including the following:
\begin{itemize}
\item the average correlated distribution of play of \speaker and \listener converges to the set of coarse-correlated equilibria of the game defined by the regularized utilities $\tilde u_\xspeaker, \tilde u_\xlistener$;
\item for any $i\in\{\xspeaker,\xlistener\}$, the average policy of Player $i$ lies
within a distance of roughly $1/\lambda_i$ from the default semantics $\vec{\tau}_i$;
\item the policies produced by piKL-Hedge guarantee that the player's regret will remain bounded by a functions whose growth is
logarithmic in the number of training steps.
\end{itemize}

\subsection{Special Case: Uniform Priors, No Costs}

When the prior over the objects is uniform, and utterance costs are all set to zero, the gradients $\nabla_{\mathbf{S}}(\mathbf{L})$ and $\nabla_{\mathbf{L}}(\mathbf{S})$, defined in (\ref{eq:nabla S}) and (\ref{eq:nabla L}), simplify into
\[
    \nabla_{\mathbf{S}}(\mathbf{L}) = \frac{1}{|M|} \mathbf{L},\quad
    \nabla_{\mathbf{L}}(\mathbf{S}) = \frac{1}{|M|} \mathbf{S}.
\]
Hence, piKL-Hedge reduces to the simple algorithm that repeatedly updates and renormalizes policy matrices according to
\begin{align*}
    \mathbf{S}^{(t+1)}       & \overset{\text{row}}{\propto} \exp\left\{ \frac{(\bar{\mathbf{L}}^{(t)})^\top + \hat\lambda_\xspeaker \log \vec{\tau}_\xspeaker}{1/(\hat\eta_\xspeaker t) + \hat\lambda_\xspeaker} \right\}, \\
    \mathbf{L}^{(t+1)}       & \overset{\text{row}}{\propto} \exp\left\{ \frac{(\bar{\mathbf{S}}^{(t)})^\top + \hat\lambda_\xlistener \log \vec{\tau}_\xlistener}{1/(\hat\eta_\xlistener t) + \hat\lambda_\xlistener} \right\},
\end{align*}
where we let $\hat\lambda_i \defeq |M|\lambda_i$ and $\hat\eta_i \defeq \eta_i / |M|$ for all $i\in\{\xspeaker,\xlistener\}$.

The above procedure is similar to the Rational Speech Acts model \cite{frank2012predicting}, a widely used probabilistic iterated response model of pragmatics. In particular, using the same matrix notation from above, we may express RSA (with $\alpha=1.0$) as:
\begin{align*}
    \bar\mL^{(0)} &= \vec{\tau}_\xlistener\\
    \mathbf{S}^{(t+1)}       & \overset{\text{row}}{\propto} (\bar{\mathbf{L}}^{(t)})^\top, \\
    \bar{\mathbf{S}}^{(t+1)} & = \mathbf{S}^{(t+1)},                                 \\
    \mathbf{L}^{(t+1)}       & \overset{\text{row}}{\propto} (\bar{\mathbf{S}}^{(t)})^\top, \\
    \bar{\mathbf{L}}^{(t+1)} & = \mathbf{L}^{(t+1)}.
\end{align*}
Thus, it is also possible to interpret \ourmethod as an RSA variant in which (1) the final policy at level $t$ is a weighted average of policies computed at lower levels, (2) both speakers and listeners downweight actions that are low-probability under the default semantics. \\

Having defined the \ourmethod objective and procedures for optimizing it, the remainder of this paper evaluates whether \ourmethod can successfully predict human judgments across standard test-beds for pragmatic implicature.

%% file: text/3_experiments.tex
\begin{figure}[t]
    \begin{figure}[H]
      \begin{tikzpicture}[overlay]
        \fill[rounded corners,black!10] (.1, 0) rectangle +(3.93,8.2);
        \fill[rounded corners,black!10] (4.19, 0) rectangle +(3.65,8.2);
        \fill[rounded corners,black!30] (.1, 7.9) rectangle +(3.93,.6);
        \fill[black!30] (.1, 7.9) rectangle +(3.93,.3);
        \fill[rounded corners,black!30] (4.19, 7.9) rectangle +(3.65,.6);
        \fill[black!30] (4.19, 7.9) rectangle +(3.65,.3);
        \node[black] at (2,8.16) {\small Default semantics};
        \node[black] at (6.05,8.16) {\small Regularized convention};
      \end{tikzpicture}
      \scalebox{.68}{\import{viz}{scalars_T2000_0.001_0.001.pgf}}%
    \end{figure}%

    \caption{Quantity implicatures in \ourmethod. (Left) Matrices representing conditional probabilities that represent the default semantics $\tau_\xspeaker$ and $\tau_\xlistener$. (Right) Matrices representing conditional probabilities that represent the resulting regularized conventions $\pi_\xspeaker$ and $\pi_\xlistener$. In this setting, \ourmethod is able to predict the correct set of interpretations.}
    \label{fig:scalars}
\end{figure}

\begin{figure}[t]
    \begin{figure}[H]
      \begin{tikzpicture}[overlay]
        \fill[rounded corners,black!10] (.1, 0) rectangle +(3.93,7.8);
        \fill[rounded corners,black!10] (4.19, 0) rectangle +(3.65,7.8);
        \fill[rounded corners,black!30] (.1, 7.5) rectangle +(3.93,.6);
        \fill[black!30] (.1, 7.5) rectangle +(3.93,.3);
        \fill[rounded corners,black!30] (4.19, 7.5) rectangle +(3.65,.6);
        \fill[black!30] (4.19, 7.5) rectangle +(3.65,.3);
        \node[black] at (2,7.76) {\small Default semantics};
        \node[black] at (6.05,7.76) {\small Regularized convention};
      \end{tikzpicture}
      \scalebox{.85}{\import{viz}{mimplicatures_T2000_0.001_0.001.pgf}}%
    \end{figure}%
{}
    \caption{Manner implicatures in \ourmethod. (Left) Matrices representing conditional probabilities that represent the default semantics $\tau_\xspeaker$ and $\tau_\xlistener$. (Right) Matrices representing conditional probabilities that represent the resulting regularized conventions $\pi_\xspeaker$ and $\pi_\xlistener$. By incorporate prior probabilities of meanings and costs for utterances, \ourmethod is able to predict the correct set of interpretations.}
    \label{fig:mimplicatures}
\end{figure}

\section{Two Model Problems: Q-implicature and M-implicature}

We begin with two simple, widely studied ``model problems'' in pragmatics: Quantity implicature and Manner implicature.
The experiments in this section aim to demonstrate that \ourmethod makes predictions that agree qualitatively with key motivating examples in theories of pragmatics.

\subsection{Quantity Implicature}

Quantity (or ``scalar'') implicatures are those in which a weak assertion is interpreted to mean that a stronger assertion does not hold. (For example, \emph{Avery ate some of the cookies} $\pto$ \emph{Avery did not eat all of the cookies}, where $\pto$ denotes pragmatic implication; \cite{huang1991neo}). The reference game we use as a model of scalar implicature is adopted from \citet{jager2012game}; its associated default semantics is shown in \Cref{fig:scalars}. Here, the utterances \emph{none}, \emph{some}, and \emph{all} are used to communicate meanings \texttt{none}, \texttt{some (not all)}, and \texttt{all}. \emph{Some} can (literally) denote \emph{all} (as we may felicitously say \emph{Avery ate some of the cookies; in fact, Avery ate all of them}), but is generally understood to \emph{implicate} \texttt{not all}.
The policy found by \ourmethod is shown in \cref{fig:scalars}, where it can be seen that this prediction is recovered by \ourmethod.

\subsection{Manner Implicature}

Another important class of implicatures are Manner implicatures, in (a subclass of) which an atypical utterance is used to denote that a situation occurred in an atypical way (\emph{I started the car} $\pto$ \emph{The car started normally}; but \emph{I got the car to start} $\pto$ \emph{The car started abnormally}; \citealp{levinson2000presumptive}). The reference game we adopt as a model of such implicatures is due to \citet{bergen2016pragmatic}. In this model, we assume we have two utterances (\emph{short} and \emph{long}) and two meanings (\texttt{freq} and \texttt{rare}) satisfying the following properties: (1) \texttt{freq} is more often the intended meaning than \texttt{rare}, (2) \emph{long} is more costly to communicate than \emph{short}, but (3) either \emph{long} or \emph{short} may, by default, denote \texttt{freq} or \texttt{rare}. In such situations, \emph{short} is understood to implicate \texttt{freq} and \emph{long} to implicate \texttt{rare}; as noted by \citet{bergen2016pragmatic}, RSA and related theories require substantial modification to derive these predictions.

When using \ourmethod to perform equilibrium search with these costs and priors, it natively predicts the correct set of interpretations (\Cref{fig:mimplicatures}).

\section{Graded Human Judgments}

\begin{table}[t]
    \centering
    \scalebox{.75}{\begin{tabular}{@{}lccccc@{}} %
            \toprule
                                                   & Literal                                 & BR                                      &                                          &                                          &                                          \\
                                                   & \listener                               & \!\speaker                              & RSA                                      & RD-RSA                                   & \ourmethod                               \\
            \midrule
            \makebox[1.2cm][l]{\taskname{all}}     & \cbox{1,1,1}{73.57\%}{+0.00\%}          & \cbox{1,1,1}{90.04\%}{+16.47\%}         & \cbox{1,1,1}{95.07\%}{+21.50\%}          & \cbox{1,1,1}{94.98\%}{+21.41\%}          & \cbox{1,1,1}{\textbf{95.96\%}}{+22.39\%} \\
            \midrule
            \makebox[1.2cm][l]{\taskname{simple}}  & \cbox{1,1,1}{70.10\%}{+0.00\%}          & \cbox{1,1,1}{88.16\%}{+18.07\%}         & \cbox{1,1,1}{\textbf{96.02\%}}{+25.92\%} & \cbox{1,1,1}{\textbf{96.02\%}}{+25.92\%} & \cbox{1,1,1}{\textbf{96.02\%}}{+25.92\%} \\
            \makebox[1.2cm][l]{\taskname{complex}} & \cbox{1,1,1}{83.86\%}{+0.00\%}          & \cbox{1,1,1}{97.83\%}{+13.97\%}         & \cbox{1,1,1}{94.74\%}{+10.88\%}          & \cbox{1,1,1}{94.35\%}{+10.49\%}          & \cbox{1,1,1}{\textbf{98.18\%}}{+14.32\%} \\
            \makebox[1.2cm][l]{\taskname{twins}}   & \cbox{1,1,1}{97.61\%}{+0.00\%}          & \cbox{1,1,1}{93.43\%}{-4.18\%}          & \cbox{1,1,1}{97.61\%}{-0.00\%}           & \cbox{1,1,1}{\textbf{98.98\%}}{+1.37\%}  & \cbox{1,1,1}{97.61\%}{-0.00\%}           \\
            \makebox[1.2cm][l]{\taskname{oddman}}  & \cbox{1,1,1}{\textbf{94.97\%}}{+0.00\%} & \cbox{1,1,1}{\textbf{94.97\%}}{-0.00\%} & \cbox{1,1,1}{\textbf{94.97\%}}{+0.00\%}  & \cbox{1,1,1}{\textbf{94.97\%}}{+0.00\%}  & \cbox{1,1,1}{\textbf{94.97\%}}{+0.00\%}  \\
            \bottomrule
        \end{tabular}}
    \caption{Correlation across different methods with graded human judgements in four reference games \citet{frank2016rational} (with the best hyperparameter settings). \ourmethod performs better than the alternatives in \taskname{all}.}
    \label{tab:graded}
\end{table}

We next study a family of four reference tasks introduced by \citet{frank2016rational}, which we refer to as \taskname{simple}, \taskname{complex}, \taskname{twins} and \taskname{oddman}. We refer readers to the original work for the default meanings that define each of these tasks. \citeauthor{frank2016rational} gathered graded human judgments about the likelihood that particular utterances might carry particular meanings. \ourmethod, like RSA-family models, captures probabilistic associations between utterances and meanings, we may evaluate the quality of its predictions in terms of correlations with human judgments.

\begin{figure*}[t]
    \centering
    \def\sc{.79}
    \scalebox{\sc}{\import{plots}{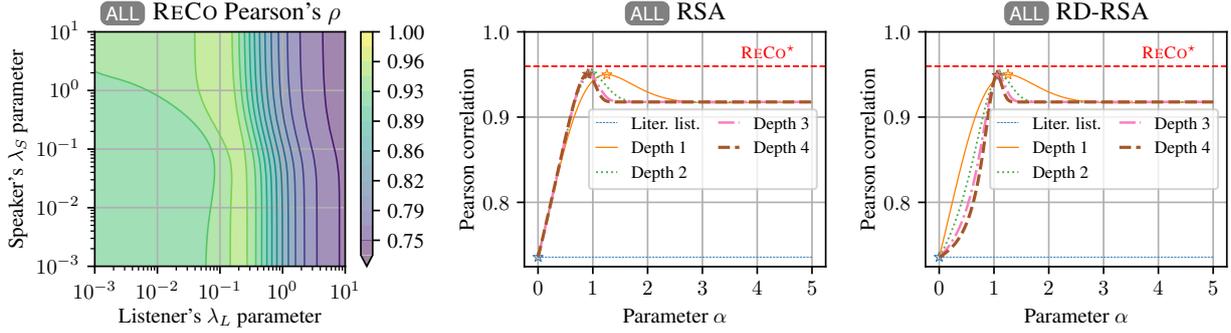}}%
    \scalebox{\sc}{\import{plots}{rsa_all_startlistener_rd0.pgf}}%
    \scalebox{\sc}{\import{plots}{rsa_all_startlistener_rd1.pgf}}%
    \caption{Pearson's correlation $\rho$ on the full dataset of graded human judgments from \cite{frank2016rational}. (Left) Correlation for \ourmethod as a function of $\lambda_\xlistener$ and $\lambda_\xspeaker$ represented as a contour plot. (Middle) Correlation between RSA at different levels of $\alpha$ and recursive depth (Right) Correlation between RD-RSA at different levels of $\alpha$ and recursive depth. (Middle, Right) \ourmethod with the best setting of $\lambda_\xlistener$ and $\lambda_\xspeaker$ is indicated with a red dashed line. Stars indicate the best $\alpha$ value at different depths.}
    \label{fig:plots_all}
\end{figure*}

Comparisons between \ourmethod, RSA, BR \speaker (i.e., best-response to a literal speaker) and RD-RSA \cite{zaslavsky2021rate} are shown in \Cref{tab:graded}, with additional information about parameters in \Cref{fig:plots_all}. In these figures, \taskname{all} denotes correlations computed across all four tasks. It can be seen that \ourmethod modestly improves upon the best predictions of RSA across a range of speaker parameters.

\begin{figure*}[t]
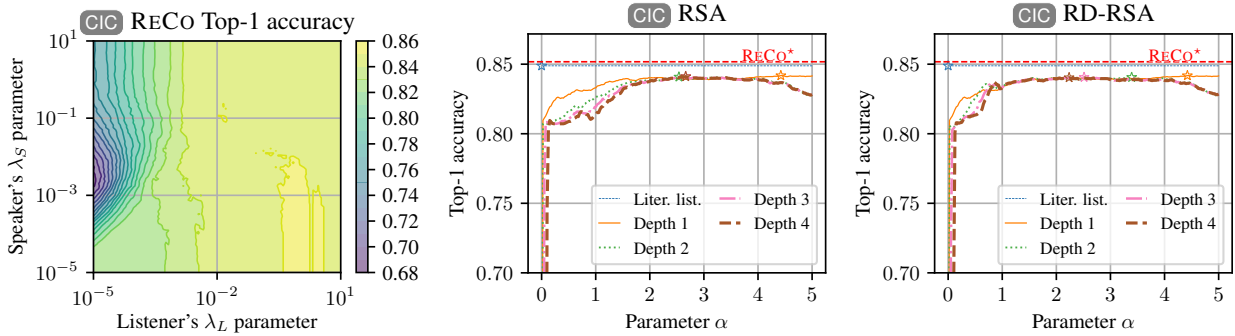

    \centering
    \def\sc{.78}
    \scalebox{\sc}{\import{plots}{pikl_cic_val_li1.pgf}}%
    \scalebox{\sc}{\import{plots}{rsa_cic_val_startlistener_rd0.pgf}}%
    \scalebox{\sc}{\import{plots}{rsa_cic_val_startlistener_rd1.pgf}}%
    \caption{Top-1 accuracy of predicting meanings on the validation set of the Colors in Context task \cite{monroe2017colors}. (Left) Accuracy for \ourmethod as a function of $\lambda_\xlistener$ and $\lambda_\xspeaker$ represented as a contour plot. (Middle) Accuracy of RSA at different levels of $\alpha$ and recursive depth (Right) Accuracy of RD-RSA at different levels of $\alpha$ and recursive depth. (Middle, Right) \ourmethod with the best setting of $\lambda_\xlistener$ and $\lambda_\xspeaker$ is indicated with a red dashed line. Stars indicate the best $\alpha$ value at different depths.}
    \label{fig:plots_cic}
\end{figure*}

\section{Complex Referents and Utterances}

\begin{table}
    \centering
    \setlength{\tabcolsep}{4pt}
    \scalebox{.8}{\begin{tabular}[c]{r@{. \ } ccc l}
        \toprule
        & \multicolumn{3}{c}{\bf Context} & \bf Utterance                                     \\
        \midrule
        1                           & \colorContext{5866A7}{2DD2BC}{C23D5A}{purple} \\
        2                           & \colorContext{5866A7}{9953AC}{2DD2A6}{blue}   \\
        3                           & \colorContext{3884C7}{02F9FD}{9E6461}{blue}   \\
        \bottomrule
    \end{tabular}}
    \caption{Example of the Colors in Context task \cite{monroe2017colors}. The \speaker produces an utterance to refer to the reference color (the one within the black box) subject to the context to a \listener. As in \Cref{fig:teaser}, notice how context affects the utterance.}
    \label{tab:cic}
\end{table}
\begin{table}[t]
    \setlength{\tabcolsep}{1.5mm}
    \scalebox{.8}{\begin{tabular}{@{}lccccc@{}} %
            \toprule
                                                      & Literal                        & BR                             &                                &                                &                                         \\
                                                      & \listener                      & \!\speaker                     & RSA                            & RD-RSA                         & \ourmethod                              \\
            \midrule
            \makebox[1.2cm][l]{\taskname{cic} (val.)} & \cbox{1,1,1}{84.88\%}{+0.00\%} & \cbox{1,1,1}{75.90\%}{-8.98\%} & \cbox{1,1,1}{84.18\%}{-0.70\%} & \cbox{1,1,1}{84.18\%}{-0.70\%} & \cbox{1,1,1}{\textbf{85.17\%}}{+0.29\%} \\
            \makebox[1.2cm][l]{\taskname{cic} (test)} & \cbox{1,1,1}{83.34\%}{+0.00\%} & \cbox{1,1,1}{74.28\%}{-9.06\%} & \cbox{1,1,1}{83.41\%}{+0.07\%} & \cbox{1,1,1}{83.41\%}{+0.07\%} & \cbox{1,1,1}{\textbf{83.62\%}}{+0.28\%} \\
            \bottomrule
        \end{tabular}}
    \caption{Performance of different models on Colors in Context \cite{monroe2017colors}. All approaches aside from BR perform well on this task -- as even literal models have access to all three referents. Note that, \ourmethod performs best.}
    \label{tab:cic_results}
\end{table}

Our final experiments focus on Colors in Context (\taskname{cic}), a dataset of color reference tasks like the one in \Cref{fig:teaser} featuring a more complex space of meanings and a larger space of utterances. Another example from the dataset (introduced by \citealp{monroe2017colors}) is given in \cref{tab:cic}.
For this task, we use human-generated utterances collected by the authors across 948 games yielding a total of 46,994 utterances. We divide this data into an 80\% / 10\% / 20\% train / validation / test splits. Here, we evaluate models by measuring the accuracy with which they can infer the intended meaning produced by a human \speaker.

\paragraph{Base models} Following past work \cite{monroe2017colors}, we first train a transformer-based literal listener as a  model that takes in the three colors and a natural language utterance, and uses these to predict the index of the referent.
We also train a transformer-based speaker model, which takes in the context and target referent and generates a natural language utterance.

\paragraph{Candidate utterances} The set of utterances are produced by first sampling 5 candidate utterances for each of the 3 possible targets from the speaker model along with the produced utterance, for a total of 16 candidates.

\paragraph{}
Results are shown in \cref{fig:plots_cic} and \cref{tab:cic_results}. As with past work \cite{mcdowell2019learning, monroe2017colors}, all models aside from BR perform well (even the literal listener); \ourmethod matches (or perhaps slightly improves upon) these results.

%% file: viz/mimplicatures_T2000_0.001_0.001.pgf
\begingroup%
\makeatletter%
\begin{pgfpicture}%
\pgfpathrectangle{\pgfpointorigin}{\pgfqpoint{3.444262in}{3.400000in}}%
\pgfusepath{use as bounding box, clip}%
\begin{pgfscope}%
\pgfsetbuttcap%
\pgfsetmiterjoin%
\pgfsetlinewidth{0.000000pt}%
\definecolor{currentstroke}{rgb}{0.000000,0.000000,0.000000}%
\pgfsetstrokecolor{currentstroke}%
\pgfsetstrokeopacity{0.000000}%
\pgfsetdash{}{0pt}%
\pgfpathmoveto{\pgfqpoint{0.000000in}{0.000000in}}%
\pgfpathlineto{\pgfqpoint{3.444262in}{0.000000in}}%
\pgfpathlineto{\pgfqpoint{3.444262in}{3.400000in}}%
\pgfpathlineto{\pgfqpoint{0.000000in}{3.400000in}}%
\pgfpathlineto{\pgfqpoint{0.000000in}{0.000000in}}%
\pgfpathclose%
\pgfusepath{}%
\end{pgfscope}%
\begin{pgfscope}%
\pgfsetbuttcap%
\pgfsetmiterjoin%
\pgfsetlinewidth{0.000000pt}%
\definecolor{currentstroke}{rgb}{0.000000,0.000000,0.000000}%
\pgfsetstrokecolor{currentstroke}%
\pgfsetstrokeopacity{0.000000}%
\pgfsetdash{}{0pt}%
\pgfpathmoveto{\pgfqpoint{0.665061in}{2.172358in}}%
\pgfpathlineto{\pgfqpoint{1.596372in}{2.172358in}}%
\pgfpathlineto{\pgfqpoint{1.596372in}{3.103668in}}%
\pgfpathlineto{\pgfqpoint{0.665061in}{3.103668in}}%
\pgfpathlineto{\pgfqpoint{0.665061in}{2.172358in}}%
\pgfpathclose%
\pgfusepath{}%
\end{pgfscope}%
\begin{pgfscope}%
\pgfpathrectangle{\pgfqpoint{0.665061in}{2.172358in}}{\pgfqpoint{0.931311in}{0.931311in}}%
\pgfusepath{clip}%
\pgfsys@transformshift{0.665061in}{2.172358in}%
\pgftext[left,bottom]{\includegraphics[interpolate=true,width=0.940000in,height=0.940000in]{mimplicatures_T2000_0.001_0.001-img0.png}}%
\end{pgfscope}%
\begin{pgfscope}%
\pgfsetbuttcap%
\pgfsetroundjoin%
\definecolor{currentfill}{rgb}{0.000000,0.000000,0.000000}%
\pgfsetfillcolor{currentfill}%
\pgfsetlinewidth{0.803000pt}%
\definecolor{currentstroke}{rgb}{0.000000,0.000000,0.000000}%
\pgfsetstrokecolor{currentstroke}%
\pgfsetdash{}{0pt}%
\pgfsys@defobject{currentmarker}{\pgfqpoint{0.000000in}{-0.048611in}}{\pgfqpoint{0.000000in}{0.000000in}}{%
\pgfpathmoveto{\pgfqpoint{0.000000in}{0.000000in}}%
\pgfpathlineto{\pgfqpoint{0.000000in}{-0.048611in}}%
\pgfusepath{stroke,fill}%
}%
\begin{pgfscope}%
\pgfsys@transformshift{0.897889in}{2.172358in}%
\pgfsys@useobject{currentmarker}{}%
\end{pgfscope}%
\end{pgfscope}%
\begin{pgfscope}%
\definecolor{textcolor}{rgb}{0.000000,0.000000,0.000000}%
\pgfsetstrokecolor{textcolor}%
\pgfsetfillcolor{textcolor}%
\pgftext[x=0.897889in,y=1.984858in,,base]{\color{textcolor}\rmfamily\fontsize{10.000000}{12.000000}\selectfont \textit{short}}%
\end{pgfscope}%
\begin{pgfscope}%
\pgfsetbuttcap%
\pgfsetroundjoin%
\definecolor{currentfill}{rgb}{0.000000,0.000000,0.000000}%
\pgfsetfillcolor{currentfill}%
\pgfsetlinewidth{0.803000pt}%
\definecolor{currentstroke}{rgb}{0.000000,0.000000,0.000000}%
\pgfsetstrokecolor{currentstroke}%
\pgfsetdash{}{0pt}%
\pgfsys@defobject{currentmarker}{\pgfqpoint{0.000000in}{-0.048611in}}{\pgfqpoint{0.000000in}{0.000000in}}{%
\pgfpathmoveto{\pgfqpoint{0.000000in}{0.000000in}}%
\pgfpathlineto{\pgfqpoint{0.000000in}{-0.048611in}}%
\pgfusepath{stroke,fill}%
}%
\begin{pgfscope}%
\pgfsys@transformshift{1.363545in}{2.172358in}%
\pgfsys@useobject{currentmarker}{}%
\end{pgfscope}%
\end{pgfscope}%
\begin{pgfscope}%
\definecolor{textcolor}{rgb}{0.000000,0.000000,0.000000}%
\pgfsetstrokecolor{textcolor}%
\pgfsetfillcolor{textcolor}%
\pgftext[x=1.363545in,y=1.984858in,,base]{\color{textcolor}\rmfamily\fontsize{10.000000}{12.000000}\selectfont \textit{long}}%
\end{pgfscope}%
\begin{pgfscope}%
\definecolor{textcolor}{rgb}{0.000000,0.000000,0.000000}%
\pgfsetstrokecolor{textcolor}%
\pgfsetfillcolor{textcolor}%
\pgftext[x=1.130717in,y=1.899234in,,top]{\color{textcolor}\rmfamily\fontsize{10.000000}{12.000000}\selectfont ...produce utterance}%
\end{pgfscope}%
\begin{pgfscope}%
\pgfsetbuttcap%
\pgfsetroundjoin%
\definecolor{currentfill}{rgb}{0.000000,0.000000,0.000000}%
\pgfsetfillcolor{currentfill}%
\pgfsetlinewidth{0.803000pt}%
\definecolor{currentstroke}{rgb}{0.000000,0.000000,0.000000}%
\pgfsetstrokecolor{currentstroke}%
\pgfsetdash{}{0pt}%
\pgfsys@defobject{currentmarker}{\pgfqpoint{-0.048611in}{0.000000in}}{\pgfqpoint{-0.000000in}{0.000000in}}{%
\pgfpathmoveto{\pgfqpoint{-0.000000in}{0.000000in}}%
\pgfpathlineto{\pgfqpoint{-0.048611in}{0.000000in}}%
\pgfusepath{stroke,fill}%
}%
\begin{pgfscope}%
\pgfsys@transformshift{0.665061in}{2.870841in}%
\pgfsys@useobject{currentmarker}{}%
\end{pgfscope}%
\end{pgfscope}%
\begin{pgfscope}%
\definecolor{textcolor}{rgb}{0.000000,0.000000,0.000000}%
\pgfsetstrokecolor{textcolor}%
\pgfsetfillcolor{textcolor}%
\pgftext[x=0.290061in, y=2.823758in, left, base]{\color{textcolor}\rmfamily\fontsize{10.000000}{12.000000}\selectfont \texttt{freq}}%
\end{pgfscope}%
\begin{pgfscope}%
\pgfsetbuttcap%
\pgfsetroundjoin%
\definecolor{currentfill}{rgb}{0.000000,0.000000,0.000000}%
\pgfsetfillcolor{currentfill}%
\pgfsetlinewidth{0.803000pt}%
\definecolor{currentstroke}{rgb}{0.000000,0.000000,0.000000}%
\pgfsetstrokecolor{currentstroke}%
\pgfsetdash{}{0pt}%
\pgfsys@defobject{currentmarker}{\pgfqpoint{-0.048611in}{0.000000in}}{\pgfqpoint{-0.000000in}{0.000000in}}{%
\pgfpathmoveto{\pgfqpoint{-0.000000in}{0.000000in}}%
\pgfpathlineto{\pgfqpoint{-0.048611in}{0.000000in}}%
\pgfusepath{stroke,fill}%
}%
\begin{pgfscope}%
\pgfsys@transformshift{0.665061in}{2.405185in}%
\pgfsys@useobject{currentmarker}{}%
\end{pgfscope}%
\end{pgfscope}%
\begin{pgfscope}%
\definecolor{textcolor}{rgb}{0.000000,0.000000,0.000000}%
\pgfsetstrokecolor{textcolor}%
\pgfsetfillcolor{textcolor}%
\pgftext[x=0.290061in, y=2.358103in, left, base]{\color{textcolor}\rmfamily\fontsize{10.000000}{12.000000}\selectfont \texttt{rare}}%
\end{pgfscope}%
\begin{pgfscope}%
\definecolor{textcolor}{rgb}{0.000000,0.000000,0.000000}%
\pgfsetstrokecolor{textcolor}%
\pgfsetfillcolor{textcolor}%
\pgftext[x=0.234506in,y=2.638013in,,bottom,rotate=90.000000]{\color{textcolor}\rmfamily\fontsize{10.000000}{12.000000}\selectfont Given meaning...}%
\end{pgfscope}%
\begin{pgfscope}%
\pgfpathrectangle{\pgfqpoint{0.665061in}{2.172358in}}{\pgfqpoint{0.931311in}{0.931311in}}%
\pgfusepath{clip}%
\pgfsetrectcap%
\pgfsetroundjoin%
\pgfsetlinewidth{0.803000pt}%
\definecolor{currentstroke}{rgb}{0.000000,0.000000,0.000000}%
\pgfsetstrokecolor{currentstroke}%
\pgfsetdash{}{0pt}%
\pgfpathmoveto{\pgfqpoint{1.130717in}{2.172358in}}%
\pgfpathlineto{\pgfqpoint{1.130717in}{3.103668in}}%
\pgfusepath{stroke}%
\end{pgfscope}%
\begin{pgfscope}%
\pgfpathrectangle{\pgfqpoint{0.665061in}{2.172358in}}{\pgfqpoint{0.931311in}{0.931311in}}%
\pgfusepath{clip}%
\pgfsetrectcap%
\pgfsetroundjoin%
\pgfsetlinewidth{0.803000pt}%
\definecolor{currentstroke}{rgb}{0.000000,0.000000,0.000000}%
\pgfsetstrokecolor{currentstroke}%
\pgfsetdash{}{0pt}%
\pgfpathmoveto{\pgfqpoint{0.665061in}{2.638013in}}%
\pgfpathlineto{\pgfqpoint{1.596372in}{2.638013in}}%
\pgfusepath{stroke}%
\end{pgfscope}%
\begin{pgfscope}%
\pgfsetrectcap%
\pgfsetmiterjoin%
\pgfsetlinewidth{1.505625pt}%
\definecolor{currentstroke}{rgb}{0.000000,0.000000,0.000000}%
\pgfsetstrokecolor{currentstroke}%
\pgfsetdash{}{0pt}%
\pgfpathmoveto{\pgfqpoint{0.665061in}{2.172357in}}%
\pgfpathlineto{\pgfqpoint{0.665061in}{3.103668in}}%
\pgfusepath{stroke}%
\end{pgfscope}%
\begin{pgfscope}%
\pgfsetrectcap%
\pgfsetmiterjoin%
\pgfsetlinewidth{1.505625pt}%
\definecolor{currentstroke}{rgb}{0.000000,0.000000,0.000000}%
\pgfsetstrokecolor{currentstroke}%
\pgfsetdash{}{0pt}%
\pgfpathmoveto{\pgfqpoint{1.596372in}{2.172357in}}%
\pgfpathlineto{\pgfqpoint{1.596372in}{3.103668in}}%
\pgfusepath{stroke}%
\end{pgfscope}%
\begin{pgfscope}%
\pgfsetrectcap%
\pgfsetmiterjoin%
\pgfsetlinewidth{1.505625pt}%
\definecolor{currentstroke}{rgb}{0.000000,0.000000,0.000000}%
\pgfsetstrokecolor{currentstroke}%
\pgfsetdash{}{0pt}%
\pgfpathmoveto{\pgfqpoint{0.665061in}{2.172358in}}%
\pgfpathlineto{\pgfqpoint{1.596372in}{2.172358in}}%
\pgfusepath{stroke}%
\end{pgfscope}%
\begin{pgfscope}%
\pgfsetrectcap%
\pgfsetmiterjoin%
\pgfsetlinewidth{1.505625pt}%
\definecolor{currentstroke}{rgb}{0.000000,0.000000,0.000000}%
\pgfsetstrokecolor{currentstroke}%
\pgfsetdash{}{0pt}%
\pgfpathmoveto{\pgfqpoint{0.665061in}{3.103668in}}%
\pgfpathlineto{\pgfqpoint{1.596372in}{3.103668in}}%
\pgfusepath{stroke}%
\end{pgfscope}%
\begin{pgfscope}%
\definecolor{textcolor}{rgb}{0.000000,0.000000,0.000000}%
\pgfsetstrokecolor{textcolor}%
\pgfsetfillcolor{textcolor}%
\pgftext[x=0.897889in,y=2.870841in,,]{\color{textcolor}\rmfamily\fontsize{10.000000}{12.000000}\selectfont 0.50}%
\end{pgfscope}%
\begin{pgfscope}%
\definecolor{textcolor}{rgb}{0.000000,0.000000,0.000000}%
\pgfsetstrokecolor{textcolor}%
\pgfsetfillcolor{textcolor}%
\pgftext[x=1.363545in,y=2.870841in,,]{\color{textcolor}\rmfamily\fontsize{10.000000}{12.000000}\selectfont 0.50}%
\end{pgfscope}%
\begin{pgfscope}%
\definecolor{textcolor}{rgb}{0.000000,0.000000,0.000000}%
\pgfsetstrokecolor{textcolor}%
\pgfsetfillcolor{textcolor}%
\pgftext[x=0.897889in,y=2.405185in,,]{\color{textcolor}\rmfamily\fontsize{10.000000}{12.000000}\selectfont 0.50}%
\end{pgfscope}%
\begin{pgfscope}%
\definecolor{textcolor}{rgb}{0.000000,0.000000,0.000000}%
\pgfsetstrokecolor{textcolor}%
\pgfsetfillcolor{textcolor}%
\pgftext[x=1.363545in,y=2.405185in,,]{\color{textcolor}\rmfamily\fontsize{10.000000}{12.000000}\selectfont 0.50}%
\end{pgfscope}%
\begin{pgfscope}%
\definecolor{textcolor}{rgb}{0.000000,0.000000,0.000000}%
\pgfsetstrokecolor{textcolor}%
\pgfsetfillcolor{textcolor}%
\pgftext[x=1.130717in,y=3.187002in,,base]{\color{textcolor}\rmfamily\fontsize{12.000000}{14.400000}\selectfont \speaker}%
\end{pgfscope}%
\begin{pgfscope}%
\pgfsetbuttcap%
\pgfsetmiterjoin%
\pgfsetlinewidth{0.000000pt}%
\definecolor{currentstroke}{rgb}{0.000000,0.000000,0.000000}%
\pgfsetstrokecolor{currentstroke}%
\pgfsetstrokeopacity{0.000000}%
\pgfsetdash{}{0pt}%
\pgfpathmoveto{\pgfqpoint{2.412951in}{2.172358in}}%
\pgfpathlineto{\pgfqpoint{3.344262in}{2.172358in}}%
\pgfpathlineto{\pgfqpoint{3.344262in}{3.103668in}}%
\pgfpathlineto{\pgfqpoint{2.412951in}{3.103668in}}%
\pgfpathlineto{\pgfqpoint{2.412951in}{2.172358in}}%
\pgfpathclose%
\pgfusepath{}%
\end{pgfscope}%
\begin{pgfscope}%
\pgfpathrectangle{\pgfqpoint{2.412951in}{2.172358in}}{\pgfqpoint{0.931311in}{0.931311in}}%
\pgfusepath{clip}%
\pgfsys@transformshift{2.412951in}{2.172358in}%
\pgftext[left,bottom]{\includegraphics[interpolate=true,width=0.940000in,height=0.940000in]{mimplicatures_T2000_0.001_0.001-img1.png}}%
\end{pgfscope}%
\begin{pgfscope}%
\pgfsetbuttcap%
\pgfsetroundjoin%
\definecolor{currentfill}{rgb}{0.000000,0.000000,0.000000}%
\pgfsetfillcolor{currentfill}%
\pgfsetlinewidth{0.803000pt}%
\definecolor{currentstroke}{rgb}{0.000000,0.000000,0.000000}%
\pgfsetstrokecolor{currentstroke}%
\pgfsetdash{}{0pt}%
\pgfsys@defobject{currentmarker}{\pgfqpoint{0.000000in}{-0.048611in}}{\pgfqpoint{0.000000in}{0.000000in}}{%
\pgfpathmoveto{\pgfqpoint{0.000000in}{0.000000in}}%
\pgfpathlineto{\pgfqpoint{0.000000in}{-0.048611in}}%
\pgfusepath{stroke,fill}%
}%
\begin{pgfscope}%
\pgfsys@transformshift{2.645779in}{2.172358in}%
\pgfsys@useobject{currentmarker}{}%
\end{pgfscope}%
\end{pgfscope}%
\begin{pgfscope}%
\definecolor{textcolor}{rgb}{0.000000,0.000000,0.000000}%
\pgfsetstrokecolor{textcolor}%
\pgfsetfillcolor{textcolor}%
\pgftext[x=2.645779in,y=1.984858in,,base]{\color{textcolor}\rmfamily\fontsize{10.000000}{12.000000}\selectfont \textit{short}}%
\end{pgfscope}%
\begin{pgfscope}%
\pgfsetbuttcap%
\pgfsetroundjoin%
\definecolor{currentfill}{rgb}{0.000000,0.000000,0.000000}%
\pgfsetfillcolor{currentfill}%
\pgfsetlinewidth{0.803000pt}%
\definecolor{currentstroke}{rgb}{0.000000,0.000000,0.000000}%
\pgfsetstrokecolor{currentstroke}%
\pgfsetdash{}{0pt}%
\pgfsys@defobject{currentmarker}{\pgfqpoint{0.000000in}{-0.048611in}}{\pgfqpoint{0.000000in}{0.000000in}}{%
\pgfpathmoveto{\pgfqpoint{0.000000in}{0.000000in}}%
\pgfpathlineto{\pgfqpoint{0.000000in}{-0.048611in}}%
\pgfusepath{stroke,fill}%
}%
\begin{pgfscope}%
\pgfsys@transformshift{3.111434in}{2.172358in}%
\pgfsys@useobject{currentmarker}{}%
\end{pgfscope}%
\end{pgfscope}%
\begin{pgfscope}%
\definecolor{textcolor}{rgb}{0.000000,0.000000,0.000000}%
\pgfsetstrokecolor{textcolor}%
\pgfsetfillcolor{textcolor}%
\pgftext[x=3.111434in,y=1.984858in,,base]{\color{textcolor}\rmfamily\fontsize{10.000000}{12.000000}\selectfont \textit{long}}%
\end{pgfscope}%
\begin{pgfscope}%
\definecolor{textcolor}{rgb}{0.000000,0.000000,0.000000}%
\pgfsetstrokecolor{textcolor}%
\pgfsetfillcolor{textcolor}%
\pgftext[x=2.878606in,y=1.899234in,,top]{\color{textcolor}\rmfamily\fontsize{10.000000}{12.000000}\selectfont ...produce utterance}%
\end{pgfscope}%
\begin{pgfscope}%
\pgfsetbuttcap%
\pgfsetroundjoin%
\definecolor{currentfill}{rgb}{0.000000,0.000000,0.000000}%
\pgfsetfillcolor{currentfill}%
\pgfsetlinewidth{0.803000pt}%
\definecolor{currentstroke}{rgb}{0.000000,0.000000,0.000000}%
\pgfsetstrokecolor{currentstroke}%
\pgfsetdash{}{0pt}%
\pgfsys@defobject{currentmarker}{\pgfqpoint{-0.048611in}{0.000000in}}{\pgfqpoint{-0.000000in}{0.000000in}}{%
\pgfpathmoveto{\pgfqpoint{-0.000000in}{0.000000in}}%
\pgfpathlineto{\pgfqpoint{-0.048611in}{0.000000in}}%
\pgfusepath{stroke,fill}%
}%
\begin{pgfscope}%
\pgfsys@transformshift{2.412951in}{2.870841in}%
\pgfsys@useobject{currentmarker}{}%
\end{pgfscope}%
\end{pgfscope}%
\begin{pgfscope}%
\definecolor{textcolor}{rgb}{0.000000,0.000000,0.000000}%
\pgfsetstrokecolor{textcolor}%
\pgfsetfillcolor{textcolor}%
\pgftext[x=2.037951in, y=2.823758in, left, base]{\color{textcolor}\rmfamily\fontsize{10.000000}{12.000000}\selectfont \texttt{freq}}%
\end{pgfscope}%
\begin{pgfscope}%
\pgfsetbuttcap%
\pgfsetroundjoin%
\definecolor{currentfill}{rgb}{0.000000,0.000000,0.000000}%
\pgfsetfillcolor{currentfill}%
\pgfsetlinewidth{0.803000pt}%
\definecolor{currentstroke}{rgb}{0.000000,0.000000,0.000000}%
\pgfsetstrokecolor{currentstroke}%
\pgfsetdash{}{0pt}%
\pgfsys@defobject{currentmarker}{\pgfqpoint{-0.048611in}{0.000000in}}{\pgfqpoint{-0.000000in}{0.000000in}}{%
\pgfpathmoveto{\pgfqpoint{-0.000000in}{0.000000in}}%
\pgfpathlineto{\pgfqpoint{-0.048611in}{0.000000in}}%
\pgfusepath{stroke,fill}%
}%
\begin{pgfscope}%
\pgfsys@transformshift{2.412951in}{2.405185in}%
\pgfsys@useobject{currentmarker}{}%
\end{pgfscope}%
\end{pgfscope}%
\begin{pgfscope}%
\definecolor{textcolor}{rgb}{0.000000,0.000000,0.000000}%
\pgfsetstrokecolor{textcolor}%
\pgfsetfillcolor{textcolor}%
\pgftext[x=2.037951in, y=2.358103in, left, base]{\color{textcolor}\rmfamily\fontsize{10.000000}{12.000000}\selectfont \texttt{rare}}%
\end{pgfscope}%
\begin{pgfscope}%
\pgfpathrectangle{\pgfqpoint{2.412951in}{2.172358in}}{\pgfqpoint{0.931311in}{0.931311in}}%
\pgfusepath{clip}%
\pgfsetrectcap%
\pgfsetroundjoin%
\pgfsetlinewidth{0.803000pt}%
\definecolor{currentstroke}{rgb}{0.000000,0.000000,0.000000}%
\pgfsetstrokecolor{currentstroke}%
\pgfsetdash{}{0pt}%
\pgfpathmoveto{\pgfqpoint{2.878606in}{2.172358in}}%
\pgfpathlineto{\pgfqpoint{2.878606in}{3.103668in}}%
\pgfusepath{stroke}%
\end{pgfscope}%
\begin{pgfscope}%
\pgfpathrectangle{\pgfqpoint{2.412951in}{2.172358in}}{\pgfqpoint{0.931311in}{0.931311in}}%
\pgfusepath{clip}%
\pgfsetrectcap%
\pgfsetroundjoin%
\pgfsetlinewidth{0.803000pt}%
\definecolor{currentstroke}{rgb}{0.000000,0.000000,0.000000}%
\pgfsetstrokecolor{currentstroke}%
\pgfsetdash{}{0pt}%
\pgfpathmoveto{\pgfqpoint{2.412951in}{2.638013in}}%
\pgfpathlineto{\pgfqpoint{3.344262in}{2.638013in}}%
\pgfusepath{stroke}%
\end{pgfscope}%
\begin{pgfscope}%
\pgfsetrectcap%
\pgfsetmiterjoin%
\pgfsetlinewidth{1.505625pt}%
\definecolor{currentstroke}{rgb}{0.000000,0.000000,0.000000}%
\pgfsetstrokecolor{currentstroke}%
\pgfsetdash{}{0pt}%
\pgfpathmoveto{\pgfqpoint{2.412951in}{2.172357in}}%
\pgfpathlineto{\pgfqpoint{2.412951in}{3.103668in}}%
\pgfusepath{stroke}%
\end{pgfscope}%
\begin{pgfscope}%
\pgfsetrectcap%
\pgfsetmiterjoin%
\pgfsetlinewidth{1.505625pt}%
\definecolor{currentstroke}{rgb}{0.000000,0.000000,0.000000}%
\pgfsetstrokecolor{currentstroke}%
\pgfsetdash{}{0pt}%
\pgfpathmoveto{\pgfqpoint{3.344262in}{2.172357in}}%
\pgfpathlineto{\pgfqpoint{3.344262in}{3.103668in}}%
\pgfusepath{stroke}%
\end{pgfscope}%
\begin{pgfscope}%
\pgfsetrectcap%
\pgfsetmiterjoin%
\pgfsetlinewidth{1.505625pt}%
\definecolor{currentstroke}{rgb}{0.000000,0.000000,0.000000}%
\pgfsetstrokecolor{currentstroke}%
\pgfsetdash{}{0pt}%
\pgfpathmoveto{\pgfqpoint{2.412951in}{2.172358in}}%
\pgfpathlineto{\pgfqpoint{3.344262in}{2.172358in}}%
\pgfusepath{stroke}%
\end{pgfscope}%
\begin{pgfscope}%
\pgfsetrectcap%
\pgfsetmiterjoin%
\pgfsetlinewidth{1.505625pt}%
\definecolor{currentstroke}{rgb}{0.000000,0.000000,0.000000}%
\pgfsetstrokecolor{currentstroke}%
\pgfsetdash{}{0pt}%
\pgfpathmoveto{\pgfqpoint{2.412951in}{3.103668in}}%
\pgfpathlineto{\pgfqpoint{3.344262in}{3.103668in}}%
\pgfusepath{stroke}%
\end{pgfscope}%
\begin{pgfscope}%
\definecolor{textcolor}{rgb}{0.000000,0.000000,0.000000}%
\pgfsetstrokecolor{textcolor}%
\pgfsetfillcolor{textcolor}%
\pgftext[x=2.645779in,y=2.870841in,,]{\color{textcolor}\rmfamily\fontsize{10.000000}{12.000000}\selectfont 0.00}%
\end{pgfscope}%
\begin{pgfscope}%
\definecolor{textcolor}{rgb}{1.000000,1.000000,1.000000}%
\pgfsetstrokecolor{textcolor}%
\pgfsetfillcolor{textcolor}%
\pgftext[x=3.111434in,y=2.870841in,,]{\color{textcolor}\rmfamily\fontsize{10.000000}{12.000000}\selectfont 1.00}%
\end{pgfscope}%
\begin{pgfscope}%
\definecolor{textcolor}{rgb}{1.000000,1.000000,1.000000}%
\pgfsetstrokecolor{textcolor}%
\pgfsetfillcolor{textcolor}%
\pgftext[x=2.645779in,y=2.405185in,,]{\color{textcolor}\rmfamily\fontsize{10.000000}{12.000000}\selectfont 1.00}%
\end{pgfscope}%
\begin{pgfscope}%
\definecolor{textcolor}{rgb}{0.000000,0.000000,0.000000}%
\pgfsetstrokecolor{textcolor}%
\pgfsetfillcolor{textcolor}%
\pgftext[x=3.111434in,y=2.405185in,,]{\color{textcolor}\rmfamily\fontsize{10.000000}{12.000000}\selectfont 0.00}%
\end{pgfscope}%
\begin{pgfscope}%
\definecolor{textcolor}{rgb}{0.000000,0.000000,0.000000}%
\pgfsetstrokecolor{textcolor}%
\pgfsetfillcolor{textcolor}%
\pgftext[x=2.878606in,y=3.187002in,,base]{\color{textcolor}\rmfamily\fontsize{12.000000}{14.400000}\selectfont \speaker}%
\end{pgfscope}%
\begin{pgfscope}%
\pgfsetbuttcap%
\pgfsetmiterjoin%
\pgfsetlinewidth{0.000000pt}%
\definecolor{currentstroke}{rgb}{0.000000,0.000000,0.000000}%
\pgfsetstrokecolor{currentstroke}%
\pgfsetstrokeopacity{0.000000}%
\pgfsetdash{}{0pt}%
\pgfpathmoveto{\pgfqpoint{0.665061in}{0.497358in}}%
\pgfpathlineto{\pgfqpoint{1.596372in}{0.497358in}}%
\pgfpathlineto{\pgfqpoint{1.596372in}{1.428668in}}%
\pgfpathlineto{\pgfqpoint{0.665061in}{1.428668in}}%
\pgfpathlineto{\pgfqpoint{0.665061in}{0.497358in}}%
\pgfpathclose%
\pgfusepath{}%
\end{pgfscope}%
\begin{pgfscope}%
\pgfpathrectangle{\pgfqpoint{0.665061in}{0.497358in}}{\pgfqpoint{0.931311in}{0.931311in}}%
\pgfusepath{clip}%
\pgfsys@transformshift{0.665061in}{0.497358in}%
\pgftext[left,bottom]{\includegraphics[interpolate=true,width=0.940000in,height=0.940000in]{mimplicatures_T2000_0.001_0.001-img2.png}}%
\end{pgfscope}%
\begin{pgfscope}%
\pgfsetbuttcap%
\pgfsetroundjoin%
\definecolor{currentfill}{rgb}{0.000000,0.000000,0.000000}%
\pgfsetfillcolor{currentfill}%
\pgfsetlinewidth{0.803000pt}%
\definecolor{currentstroke}{rgb}{0.000000,0.000000,0.000000}%
\pgfsetstrokecolor{currentstroke}%
\pgfsetdash{}{0pt}%
\pgfsys@defobject{currentmarker}{\pgfqpoint{0.000000in}{-0.048611in}}{\pgfqpoint{0.000000in}{0.000000in}}{%
\pgfpathmoveto{\pgfqpoint{0.000000in}{0.000000in}}%
\pgfpathlineto{\pgfqpoint{0.000000in}{-0.048611in}}%
\pgfusepath{stroke,fill}%
}%
\begin{pgfscope}%
\pgfsys@transformshift{0.897889in}{0.497358in}%
\pgfsys@useobject{currentmarker}{}%
\end{pgfscope}%
\end{pgfscope}%
\begin{pgfscope}%
\definecolor{textcolor}{rgb}{0.000000,0.000000,0.000000}%
\pgfsetstrokecolor{textcolor}%
\pgfsetfillcolor{textcolor}%
\pgftext[x=0.897889in,y=0.309858in,,base]{\color{textcolor}\rmfamily\fontsize{10.000000}{12.000000}\selectfont \texttt{freq}}%
\end{pgfscope}%
\begin{pgfscope}%
\pgfsetbuttcap%
\pgfsetroundjoin%
\definecolor{currentfill}{rgb}{0.000000,0.000000,0.000000}%
\pgfsetfillcolor{currentfill}%
\pgfsetlinewidth{0.803000pt}%
\definecolor{currentstroke}{rgb}{0.000000,0.000000,0.000000}%
\pgfsetstrokecolor{currentstroke}%
\pgfsetdash{}{0pt}%
\pgfsys@defobject{currentmarker}{\pgfqpoint{0.000000in}{-0.048611in}}{\pgfqpoint{0.000000in}{0.000000in}}{%
\pgfpathmoveto{\pgfqpoint{0.000000in}{0.000000in}}%
\pgfpathlineto{\pgfqpoint{0.000000in}{-0.048611in}}%
\pgfusepath{stroke,fill}%
}%
\begin{pgfscope}%
\pgfsys@transformshift{1.363545in}{0.497358in}%
\pgfsys@useobject{currentmarker}{}%
\end{pgfscope}%
\end{pgfscope}%
\begin{pgfscope}%
\definecolor{textcolor}{rgb}{0.000000,0.000000,0.000000}%
\pgfsetstrokecolor{textcolor}%
\pgfsetfillcolor{textcolor}%
\pgftext[x=1.363545in,y=0.309858in,,base]{\color{textcolor}\rmfamily\fontsize{10.000000}{12.000000}\selectfont \texttt{rare}}%
\end{pgfscope}%
\begin{pgfscope}%
\definecolor{textcolor}{rgb}{0.000000,0.000000,0.000000}%
\pgfsetstrokecolor{textcolor}%
\pgfsetfillcolor{textcolor}%
\pgftext[x=1.130717in,y=0.224234in,,top]{\color{textcolor}\rmfamily\fontsize{10.000000}{12.000000}\selectfont ...guess meaning}%
\end{pgfscope}%
\begin{pgfscope}%
\pgfsetbuttcap%
\pgfsetroundjoin%
\definecolor{currentfill}{rgb}{0.000000,0.000000,0.000000}%
\pgfsetfillcolor{currentfill}%
\pgfsetlinewidth{0.803000pt}%
\definecolor{currentstroke}{rgb}{0.000000,0.000000,0.000000}%
\pgfsetstrokecolor{currentstroke}%
\pgfsetdash{}{0pt}%
\pgfsys@defobject{currentmarker}{\pgfqpoint{-0.048611in}{0.000000in}}{\pgfqpoint{-0.000000in}{0.000000in}}{%
\pgfpathmoveto{\pgfqpoint{-0.000000in}{0.000000in}}%
\pgfpathlineto{\pgfqpoint{-0.048611in}{0.000000in}}%
\pgfusepath{stroke,fill}%
}%
\begin{pgfscope}%
\pgfsys@transformshift{0.665061in}{1.195841in}%
\pgfsys@useobject{currentmarker}{}%
\end{pgfscope}%
\end{pgfscope}%
\begin{pgfscope}%
\definecolor{textcolor}{rgb}{0.000000,0.000000,0.000000}%
\pgfsetstrokecolor{textcolor}%
\pgfsetfillcolor{textcolor}%
\pgftext[x=0.279789in, y=1.148758in, left, base]{\color{textcolor}\rmfamily\fontsize{10.000000}{12.000000}\selectfont \textit{short}}%
\end{pgfscope}%
\begin{pgfscope}%
\pgfsetbuttcap%
\pgfsetroundjoin%
\definecolor{currentfill}{rgb}{0.000000,0.000000,0.000000}%
\pgfsetfillcolor{currentfill}%
\pgfsetlinewidth{0.803000pt}%
\definecolor{currentstroke}{rgb}{0.000000,0.000000,0.000000}%
\pgfsetstrokecolor{currentstroke}%
\pgfsetdash{}{0pt}%
\pgfsys@defobject{currentmarker}{\pgfqpoint{-0.048611in}{0.000000in}}{\pgfqpoint{-0.000000in}{0.000000in}}{%
\pgfpathmoveto{\pgfqpoint{-0.000000in}{0.000000in}}%
\pgfpathlineto{\pgfqpoint{-0.048611in}{0.000000in}}%
\pgfusepath{stroke,fill}%
}%
\begin{pgfscope}%
\pgfsys@transformshift{0.665061in}{0.730185in}%
\pgfsys@useobject{currentmarker}{}%
\end{pgfscope}%
\end{pgfscope}%
\begin{pgfscope}%
\definecolor{textcolor}{rgb}{0.000000,0.000000,0.000000}%
\pgfsetstrokecolor{textcolor}%
\pgfsetfillcolor{textcolor}%
\pgftext[x=0.320896in, y=0.683103in, left, base]{\color{textcolor}\rmfamily\fontsize{10.000000}{12.000000}\selectfont \textit{long}}%
\end{pgfscope}%
\begin{pgfscope}%
\definecolor{textcolor}{rgb}{0.000000,0.000000,0.000000}%
\pgfsetstrokecolor{textcolor}%
\pgfsetfillcolor{textcolor}%
\pgftext[x=0.224234in,y=0.963013in,,bottom,rotate=90.000000]{\color{textcolor}\rmfamily\fontsize{10.000000}{12.000000}\selectfont Given utterance...}%
\end{pgfscope}%
\begin{pgfscope}%
\pgfpathrectangle{\pgfqpoint{0.665061in}{0.497358in}}{\pgfqpoint{0.931311in}{0.931311in}}%
\pgfusepath{clip}%
\pgfsetrectcap%
\pgfsetroundjoin%
\pgfsetlinewidth{0.803000pt}%
\definecolor{currentstroke}{rgb}{0.000000,0.000000,0.000000}%
\pgfsetstrokecolor{currentstroke}%
\pgfsetdash{}{0pt}%
\pgfpathmoveto{\pgfqpoint{0.665061in}{0.963013in}}%
\pgfpathlineto{\pgfqpoint{1.596372in}{0.963013in}}%
\pgfusepath{stroke}%
\end{pgfscope}%
\begin{pgfscope}%
\pgfpathrectangle{\pgfqpoint{0.665061in}{0.497358in}}{\pgfqpoint{0.931311in}{0.931311in}}%
\pgfusepath{clip}%
\pgfsetrectcap%
\pgfsetroundjoin%
\pgfsetlinewidth{0.803000pt}%
\definecolor{currentstroke}{rgb}{0.000000,0.000000,0.000000}%
\pgfsetstrokecolor{currentstroke}%
\pgfsetdash{}{0pt}%
\pgfpathmoveto{\pgfqpoint{1.130717in}{0.497358in}}%
\pgfpathlineto{\pgfqpoint{1.130717in}{1.428668in}}%
\pgfusepath{stroke}%
\end{pgfscope}%
\begin{pgfscope}%
\pgfsetrectcap%
\pgfsetmiterjoin%
\pgfsetlinewidth{1.505625pt}%
\definecolor{currentstroke}{rgb}{0.000000,0.000000,0.000000}%
\pgfsetstrokecolor{currentstroke}%
\pgfsetdash{}{0pt}%
\pgfpathmoveto{\pgfqpoint{0.665061in}{0.497357in}}%
\pgfpathlineto{\pgfqpoint{0.665061in}{1.428668in}}%
\pgfusepath{stroke}%
\end{pgfscope}%
\begin{pgfscope}%
\pgfsetrectcap%
\pgfsetmiterjoin%
\pgfsetlinewidth{1.505625pt}%
\definecolor{currentstroke}{rgb}{0.000000,0.000000,0.000000}%
\pgfsetstrokecolor{currentstroke}%
\pgfsetdash{}{0pt}%
\pgfpathmoveto{\pgfqpoint{1.596372in}{0.497357in}}%
\pgfpathlineto{\pgfqpoint{1.596372in}{1.428668in}}%
\pgfusepath{stroke}%
\end{pgfscope}%
\begin{pgfscope}%
\pgfsetrectcap%
\pgfsetmiterjoin%
\pgfsetlinewidth{1.505625pt}%
\definecolor{currentstroke}{rgb}{0.000000,0.000000,0.000000}%
\pgfsetstrokecolor{currentstroke}%
\pgfsetdash{}{0pt}%
\pgfpathmoveto{\pgfqpoint{0.665061in}{0.497358in}}%
\pgfpathlineto{\pgfqpoint{1.596372in}{0.497358in}}%
\pgfusepath{stroke}%
\end{pgfscope}%
\begin{pgfscope}%
\pgfsetrectcap%
\pgfsetmiterjoin%
\pgfsetlinewidth{1.505625pt}%
\definecolor{currentstroke}{rgb}{0.000000,0.000000,0.000000}%
\pgfsetstrokecolor{currentstroke}%
\pgfsetdash{}{0pt}%
\pgfpathmoveto{\pgfqpoint{0.665061in}{1.428668in}}%
\pgfpathlineto{\pgfqpoint{1.596372in}{1.428668in}}%
\pgfusepath{stroke}%
\end{pgfscope}%
\begin{pgfscope}%
\definecolor{textcolor}{rgb}{0.000000,0.000000,0.000000}%
\pgfsetstrokecolor{textcolor}%
\pgfsetfillcolor{textcolor}%
\pgftext[x=0.897889in,y=1.195841in,,]{\color{textcolor}\rmfamily\fontsize{10.000000}{12.000000}\selectfont 0.50}%
\end{pgfscope}%
\begin{pgfscope}%
\definecolor{textcolor}{rgb}{0.000000,0.000000,0.000000}%
\pgfsetstrokecolor{textcolor}%
\pgfsetfillcolor{textcolor}%
\pgftext[x=1.363545in,y=1.195841in,,]{\color{textcolor}\rmfamily\fontsize{10.000000}{12.000000}\selectfont 0.50}%
\end{pgfscope}%
\begin{pgfscope}%
\definecolor{textcolor}{rgb}{0.000000,0.000000,0.000000}%
\pgfsetstrokecolor{textcolor}%
\pgfsetfillcolor{textcolor}%
\pgftext[x=0.897889in,y=0.730185in,,]{\color{textcolor}\rmfamily\fontsize{10.000000}{12.000000}\selectfont 0.50}%
\end{pgfscope}%
\begin{pgfscope}%
\definecolor{textcolor}{rgb}{0.000000,0.000000,0.000000}%
\pgfsetstrokecolor{textcolor}%
\pgfsetfillcolor{textcolor}%
\pgftext[x=1.363545in,y=0.730185in,,]{\color{textcolor}\rmfamily\fontsize{10.000000}{12.000000}\selectfont 0.50}%
\end{pgfscope}%
\begin{pgfscope}%
\definecolor{textcolor}{rgb}{0.000000,0.000000,0.000000}%
\pgfsetstrokecolor{textcolor}%
\pgfsetfillcolor{textcolor}%
\pgftext[x=1.130717in,y=1.512002in,,base]{\color{textcolor}\rmfamily\fontsize{12.000000}{14.400000}\selectfont \listener}%
\end{pgfscope}%
\begin{pgfscope}%
\pgfsetbuttcap%
\pgfsetmiterjoin%
\pgfsetlinewidth{0.000000pt}%
\definecolor{currentstroke}{rgb}{0.000000,0.000000,0.000000}%
\pgfsetstrokecolor{currentstroke}%
\pgfsetstrokeopacity{0.000000}%
\pgfsetdash{}{0pt}%
\pgfpathmoveto{\pgfqpoint{2.412951in}{0.497358in}}%
\pgfpathlineto{\pgfqpoint{3.344262in}{0.497358in}}%
\pgfpathlineto{\pgfqpoint{3.344262in}{1.428668in}}%
\pgfpathlineto{\pgfqpoint{2.412951in}{1.428668in}}%
\pgfpathlineto{\pgfqpoint{2.412951in}{0.497358in}}%
\pgfpathclose%
\pgfusepath{}%
\end{pgfscope}%
\begin{pgfscope}%
\pgfpathrectangle{\pgfqpoint{2.412951in}{0.497358in}}{\pgfqpoint{0.931311in}{0.931311in}}%
\pgfusepath{clip}%
\pgfsys@transformshift{2.412951in}{0.497358in}%
\pgftext[left,bottom]{\includegraphics[interpolate=true,width=0.940000in,height=0.940000in]{mimplicatures_T2000_0.001_0.001-img3.png}}%
\end{pgfscope}%
\begin{pgfscope}%
\pgfsetbuttcap%
\pgfsetroundjoin%
\definecolor{currentfill}{rgb}{0.000000,0.000000,0.000000}%
\pgfsetfillcolor{currentfill}%
\pgfsetlinewidth{0.803000pt}%
\definecolor{currentstroke}{rgb}{0.000000,0.000000,0.000000}%
\pgfsetstrokecolor{currentstroke}%
\pgfsetdash{}{0pt}%
\pgfsys@defobject{currentmarker}{\pgfqpoint{0.000000in}{-0.048611in}}{\pgfqpoint{0.000000in}{0.000000in}}{%
\pgfpathmoveto{\pgfqpoint{0.000000in}{0.000000in}}%
\pgfpathlineto{\pgfqpoint{0.000000in}{-0.048611in}}%
\pgfusepath{stroke,fill}%
}%
\begin{pgfscope}%
\pgfsys@transformshift{2.645779in}{0.497358in}%
\pgfsys@useobject{currentmarker}{}%
\end{pgfscope}%
\end{pgfscope}%
\begin{pgfscope}%
\definecolor{textcolor}{rgb}{0.000000,0.000000,0.000000}%
\pgfsetstrokecolor{textcolor}%
\pgfsetfillcolor{textcolor}%
\pgftext[x=2.645779in,y=0.309858in,,base]{\color{textcolor}\rmfamily\fontsize{10.000000}{12.000000}\selectfont \texttt{freq}}%
\end{pgfscope}%
\begin{pgfscope}%
\pgfsetbuttcap%
\pgfsetroundjoin%
\definecolor{currentfill}{rgb}{0.000000,0.000000,0.000000}%
\pgfsetfillcolor{currentfill}%
\pgfsetlinewidth{0.803000pt}%
\definecolor{currentstroke}{rgb}{0.000000,0.000000,0.000000}%
\pgfsetstrokecolor{currentstroke}%
\pgfsetdash{}{0pt}%
\pgfsys@defobject{currentmarker}{\pgfqpoint{0.000000in}{-0.048611in}}{\pgfqpoint{0.000000in}{0.000000in}}{%
\pgfpathmoveto{\pgfqpoint{0.000000in}{0.000000in}}%
\pgfpathlineto{\pgfqpoint{0.000000in}{-0.048611in}}%
\pgfusepath{stroke,fill}%
}%
\begin{pgfscope}%
\pgfsys@transformshift{3.111434in}{0.497358in}%
\pgfsys@useobject{currentmarker}{}%
\end{pgfscope}%
\end{pgfscope}%
\begin{pgfscope}%
\definecolor{textcolor}{rgb}{0.000000,0.000000,0.000000}%
\pgfsetstrokecolor{textcolor}%
\pgfsetfillcolor{textcolor}%
\pgftext[x=3.111434in,y=0.309858in,,base]{\color{textcolor}\rmfamily\fontsize{10.000000}{12.000000}\selectfont \texttt{rare}}%
\end{pgfscope}%
\begin{pgfscope}%
\definecolor{textcolor}{rgb}{0.000000,0.000000,0.000000}%
\pgfsetstrokecolor{textcolor}%
\pgfsetfillcolor{textcolor}%
\pgftext[x=2.878606in,y=0.224234in,,top]{\color{textcolor}\rmfamily\fontsize{10.000000}{12.000000}\selectfont ...guess meaning}%
\end{pgfscope}%
\begin{pgfscope}%
\pgfsetbuttcap%
\pgfsetroundjoin%
\definecolor{currentfill}{rgb}{0.000000,0.000000,0.000000}%
\pgfsetfillcolor{currentfill}%
\pgfsetlinewidth{0.803000pt}%
\definecolor{currentstroke}{rgb}{0.000000,0.000000,0.000000}%
\pgfsetstrokecolor{currentstroke}%
\pgfsetdash{}{0pt}%
\pgfsys@defobject{currentmarker}{\pgfqpoint{-0.048611in}{0.000000in}}{\pgfqpoint{-0.000000in}{0.000000in}}{%
\pgfpathmoveto{\pgfqpoint{-0.000000in}{0.000000in}}%
\pgfpathlineto{\pgfqpoint{-0.048611in}{0.000000in}}%
\pgfusepath{stroke,fill}%
}%
\begin{pgfscope}%
\pgfsys@transformshift{2.412951in}{1.195841in}%
\pgfsys@useobject{currentmarker}{}%
\end{pgfscope}%
\end{pgfscope}%
\begin{pgfscope}%
\definecolor{textcolor}{rgb}{0.000000,0.000000,0.000000}%
\pgfsetstrokecolor{textcolor}%
\pgfsetfillcolor{textcolor}%
\pgftext[x=2.027679in, y=1.148758in, left, base]{\color{textcolor}\rmfamily\fontsize{10.000000}{12.000000}\selectfont \textit{short}}%
\end{pgfscope}%
\begin{pgfscope}%
\pgfsetbuttcap%
\pgfsetroundjoin%
\definecolor{currentfill}{rgb}{0.000000,0.000000,0.000000}%
\pgfsetfillcolor{currentfill}%
\pgfsetlinewidth{0.803000pt}%
\definecolor{currentstroke}{rgb}{0.000000,0.000000,0.000000}%
\pgfsetstrokecolor{currentstroke}%
\pgfsetdash{}{0pt}%
\pgfsys@defobject{currentmarker}{\pgfqpoint{-0.048611in}{0.000000in}}{\pgfqpoint{-0.000000in}{0.000000in}}{%
\pgfpathmoveto{\pgfqpoint{-0.000000in}{0.000000in}}%
\pgfpathlineto{\pgfqpoint{-0.048611in}{0.000000in}}%
\pgfusepath{stroke,fill}%
}%
\begin{pgfscope}%
\pgfsys@transformshift{2.412951in}{0.730185in}%
\pgfsys@useobject{currentmarker}{}%
\end{pgfscope}%
\end{pgfscope}%
\begin{pgfscope}%
\definecolor{textcolor}{rgb}{0.000000,0.000000,0.000000}%
\pgfsetstrokecolor{textcolor}%
\pgfsetfillcolor{textcolor}%
\pgftext[x=2.068786in, y=0.683103in, left, base]{\color{textcolor}\rmfamily\fontsize{10.000000}{12.000000}\selectfont \textit{long}}%
\end{pgfscope}%
\begin{pgfscope}%
\pgfpathrectangle{\pgfqpoint{2.412951in}{0.497358in}}{\pgfqpoint{0.931311in}{0.931311in}}%
\pgfusepath{clip}%
\pgfsetrectcap%
\pgfsetroundjoin%
\pgfsetlinewidth{0.803000pt}%
\definecolor{currentstroke}{rgb}{0.000000,0.000000,0.000000}%
\pgfsetstrokecolor{currentstroke}%
\pgfsetdash{}{0pt}%
\pgfpathmoveto{\pgfqpoint{2.412951in}{0.963013in}}%
\pgfpathlineto{\pgfqpoint{3.344262in}{0.963013in}}%
\pgfusepath{stroke}%
\end{pgfscope}%
\begin{pgfscope}%
\pgfpathrectangle{\pgfqpoint{2.412951in}{0.497358in}}{\pgfqpoint{0.931311in}{0.931311in}}%
\pgfusepath{clip}%
\pgfsetrectcap%
\pgfsetroundjoin%
\pgfsetlinewidth{0.803000pt}%
\definecolor{currentstroke}{rgb}{0.000000,0.000000,0.000000}%
\pgfsetstrokecolor{currentstroke}%
\pgfsetdash{}{0pt}%
\pgfpathmoveto{\pgfqpoint{2.878606in}{0.497358in}}%
\pgfpathlineto{\pgfqpoint{2.878606in}{1.428668in}}%
\pgfusepath{stroke}%
\end{pgfscope}%
\begin{pgfscope}%
\pgfsetrectcap%
\pgfsetmiterjoin%
\pgfsetlinewidth{1.505625pt}%
\definecolor{currentstroke}{rgb}{0.000000,0.000000,0.000000}%
\pgfsetstrokecolor{currentstroke}%
\pgfsetdash{}{0pt}%
\pgfpathmoveto{\pgfqpoint{2.412951in}{0.497357in}}%
\pgfpathlineto{\pgfqpoint{2.412951in}{1.428668in}}%
\pgfusepath{stroke}%
\end{pgfscope}%
\begin{pgfscope}%
\pgfsetrectcap%
\pgfsetmiterjoin%
\pgfsetlinewidth{1.505625pt}%
\definecolor{currentstroke}{rgb}{0.000000,0.000000,0.000000}%
\pgfsetstrokecolor{currentstroke}%
\pgfsetdash{}{0pt}%
\pgfpathmoveto{\pgfqpoint{3.344262in}{0.497357in}}%
\pgfpathlineto{\pgfqpoint{3.344262in}{1.428668in}}%
\pgfusepath{stroke}%
\end{pgfscope}%
\begin{pgfscope}%
\pgfsetrectcap%
\pgfsetmiterjoin%
\pgfsetlinewidth{1.505625pt}%
\definecolor{currentstroke}{rgb}{0.000000,0.000000,0.000000}%
\pgfsetstrokecolor{currentstroke}%
\pgfsetdash{}{0pt}%
\pgfpathmoveto{\pgfqpoint{2.412951in}{0.497358in}}%
\pgfpathlineto{\pgfqpoint{3.344262in}{0.497358in}}%
\pgfusepath{stroke}%
\end{pgfscope}%
\begin{pgfscope}%
\pgfsetrectcap%
\pgfsetmiterjoin%
\pgfsetlinewidth{1.505625pt}%
\definecolor{currentstroke}{rgb}{0.000000,0.000000,0.000000}%
\pgfsetstrokecolor{currentstroke}%
\pgfsetdash{}{0pt}%
\pgfpathmoveto{\pgfqpoint{2.412951in}{1.428668in}}%
\pgfpathlineto{\pgfqpoint{3.344262in}{1.428668in}}%
\pgfusepath{stroke}%
\end{pgfscope}%
\begin{pgfscope}%
\definecolor{textcolor}{rgb}{0.000000,0.000000,0.000000}%
\pgfsetstrokecolor{textcolor}%
\pgfsetfillcolor{textcolor}%
\pgftext[x=2.645779in,y=1.195841in,,]{\color{textcolor}\rmfamily\fontsize{10.000000}{12.000000}\selectfont 0.00}%
\end{pgfscope}%
\begin{pgfscope}%
\definecolor{textcolor}{rgb}{1.000000,1.000000,1.000000}%
\pgfsetstrokecolor{textcolor}%
\pgfsetfillcolor{textcolor}%
\pgftext[x=3.111434in,y=1.195841in,,]{\color{textcolor}\rmfamily\fontsize{10.000000}{12.000000}\selectfont 1.00}%
\end{pgfscope}%
\begin{pgfscope}%
\definecolor{textcolor}{rgb}{1.000000,1.000000,1.000000}%
\pgfsetstrokecolor{textcolor}%
\pgfsetfillcolor{textcolor}%
\pgftext[x=2.645779in,y=0.730185in,,]{\color{textcolor}\rmfamily\fontsize{10.000000}{12.000000}\selectfont 1.00}%
\end{pgfscope}%
\begin{pgfscope}%
\definecolor{textcolor}{rgb}{0.000000,0.000000,0.000000}%
\pgfsetstrokecolor{textcolor}%
\pgfsetfillcolor{textcolor}%
\pgftext[x=3.111434in,y=0.730185in,,]{\color{textcolor}\rmfamily\fontsize{10.000000}{12.000000}\selectfont 0.00}%
\end{pgfscope}%
\begin{pgfscope}%
\definecolor{textcolor}{rgb}{0.000000,0.000000,0.000000}%
\pgfsetstrokecolor{textcolor}%
\pgfsetfillcolor{textcolor}%
\pgftext[x=2.878606in,y=1.512002in,,base]{\color{textcolor}\rmfamily\fontsize{12.000000}{14.400000}\selectfont \listener}%
\end{pgfscope}%
\end{pgfpicture}%
\makeatother%
\endgroup%

%% file: plots/rsa_all_startlistener_rd0.pgf
\begingroup%
\makeatletter%
\begin{pgfpicture}%
\pgfpathrectangle{\pgfpointorigin}{\pgfqpoint{2.630555in}{2.378626in}}%
\pgfusepath{use as bounding box, clip}%
\begin{pgfscope}%
\pgfsetbuttcap%
\pgfsetmiterjoin%
\definecolor{currentfill}{rgb}{1.000000,1.000000,1.000000}%
\pgfsetfillcolor{currentfill}%
\pgfsetlinewidth{0.000000pt}%
\definecolor{currentstroke}{rgb}{1.000000,1.000000,1.000000}%
\pgfsetstrokecolor{currentstroke}%
\pgfsetdash{}{0pt}%
\pgfpathmoveto{\pgfqpoint{0.000000in}{0.000000in}}%
\pgfpathlineto{\pgfqpoint{2.630555in}{0.000000in}}%
\pgfpathlineto{\pgfqpoint{2.630555in}{2.378626in}}%
\pgfpathlineto{\pgfqpoint{0.000000in}{2.378626in}}%
\pgfpathlineto{\pgfqpoint{0.000000in}{0.000000in}}%
\pgfpathclose%
\pgfusepath{fill}%
\end{pgfscope}%
\begin{pgfscope}%
\pgfsetbuttcap%
\pgfsetmiterjoin%
\definecolor{currentfill}{rgb}{1.000000,1.000000,1.000000}%
\pgfsetfillcolor{currentfill}%
\pgfsetlinewidth{0.000000pt}%
\definecolor{currentstroke}{rgb}{0.000000,0.000000,0.000000}%
\pgfsetstrokecolor{currentstroke}%
\pgfsetstrokeopacity{0.000000}%
\pgfsetdash{}{0pt}%
\pgfpathmoveto{\pgfqpoint{0.554481in}{0.501245in}}%
\pgfpathlineto{\pgfqpoint{2.530555in}{0.501245in}}%
\pgfpathlineto{\pgfqpoint{2.530555in}{2.052917in}}%
\pgfpathlineto{\pgfqpoint{0.554481in}{2.052917in}}%
\pgfpathlineto{\pgfqpoint{0.554481in}{0.501245in}}%
\pgfpathclose%
\pgfusepath{fill}%
\end{pgfscope}%
\begin{pgfscope}%
\pgfpathrectangle{\pgfqpoint{0.554481in}{0.501245in}}{\pgfqpoint{1.976074in}{1.551672in}}%
\pgfusepath{clip}%
\pgfsetrectcap%
\pgfsetroundjoin%
\pgfsetlinewidth{0.803000pt}%
\definecolor{currentstroke}{rgb}{0.690196,0.690196,0.690196}%
\pgfsetstrokecolor{currentstroke}%
\pgfsetdash{}{0pt}%
\pgfpathmoveto{\pgfqpoint{0.644303in}{0.501245in}}%
\pgfpathlineto{\pgfqpoint{0.644303in}{2.052917in}}%
\pgfusepath{stroke}%
\end{pgfscope}%
\begin{pgfscope}%
\pgfsetbuttcap%
\pgfsetroundjoin%
\definecolor{currentfill}{rgb}{0.000000,0.000000,0.000000}%
\pgfsetfillcolor{currentfill}%
\pgfsetlinewidth{0.803000pt}%
\definecolor{currentstroke}{rgb}{0.000000,0.000000,0.000000}%
\pgfsetstrokecolor{currentstroke}%
\pgfsetdash{}{0pt}%
\pgfsys@defobject{currentmarker}{\pgfqpoint{0.000000in}{-0.048611in}}{\pgfqpoint{0.000000in}{0.000000in}}{%
\pgfpathmoveto{\pgfqpoint{0.000000in}{0.000000in}}%
\pgfpathlineto{\pgfqpoint{0.000000in}{-0.048611in}}%
\pgfusepath{stroke,fill}%
}%
\begin{pgfscope}%
\pgfsys@transformshift{0.644303in}{0.501245in}%
\pgfsys@useobject{currentmarker}{}%
\end{pgfscope}%
\end{pgfscope}%
\begin{pgfscope}%
\definecolor{textcolor}{rgb}{0.000000,0.000000,0.000000}%
\pgfsetstrokecolor{textcolor}%
\pgfsetfillcolor{textcolor}%
\pgftext[x=0.644303in,y=0.404023in,,top]{\color{textcolor}\rmfamily\fontsize{10.000000}{12.000000}\selectfont \(\displaystyle {0}\)}%
\end{pgfscope}%
\begin{pgfscope}%
\pgfpathrectangle{\pgfqpoint{0.554481in}{0.501245in}}{\pgfqpoint{1.976074in}{1.551672in}}%
\pgfusepath{clip}%
\pgfsetrectcap%
\pgfsetroundjoin%
\pgfsetlinewidth{0.803000pt}%
\definecolor{currentstroke}{rgb}{0.690196,0.690196,0.690196}%
\pgfsetstrokecolor{currentstroke}%
\pgfsetdash{}{0pt}%
\pgfpathmoveto{\pgfqpoint{1.003589in}{0.501245in}}%
\pgfpathlineto{\pgfqpoint{1.003589in}{2.052917in}}%
\pgfusepath{stroke}%
\end{pgfscope}%
\begin{pgfscope}%
\pgfsetbuttcap%
\pgfsetroundjoin%
\definecolor{currentfill}{rgb}{0.000000,0.000000,0.000000}%
\pgfsetfillcolor{currentfill}%
\pgfsetlinewidth{0.803000pt}%
\definecolor{currentstroke}{rgb}{0.000000,0.000000,0.000000}%
\pgfsetstrokecolor{currentstroke}%
\pgfsetdash{}{0pt}%
\pgfsys@defobject{currentmarker}{\pgfqpoint{0.000000in}{-0.048611in}}{\pgfqpoint{0.000000in}{0.000000in}}{%
\pgfpathmoveto{\pgfqpoint{0.000000in}{0.000000in}}%
\pgfpathlineto{\pgfqpoint{0.000000in}{-0.048611in}}%
\pgfusepath{stroke,fill}%
}%
\begin{pgfscope}%
\pgfsys@transformshift{1.003589in}{0.501245in}%
\pgfsys@useobject{currentmarker}{}%
\end{pgfscope}%
\end{pgfscope}%
\begin{pgfscope}%
\definecolor{textcolor}{rgb}{0.000000,0.000000,0.000000}%
\pgfsetstrokecolor{textcolor}%
\pgfsetfillcolor{textcolor}%
\pgftext[x=1.003589in,y=0.404023in,,top]{\color{textcolor}\rmfamily\fontsize{10.000000}{12.000000}\selectfont \(\displaystyle {1}\)}%
\end{pgfscope}%
\begin{pgfscope}%
\pgfpathrectangle{\pgfqpoint{0.554481in}{0.501245in}}{\pgfqpoint{1.976074in}{1.551672in}}%
\pgfusepath{clip}%
\pgfsetrectcap%
\pgfsetroundjoin%
\pgfsetlinewidth{0.803000pt}%
\definecolor{currentstroke}{rgb}{0.690196,0.690196,0.690196}%
\pgfsetstrokecolor{currentstroke}%
\pgfsetdash{}{0pt}%
\pgfpathmoveto{\pgfqpoint{1.362875in}{0.501245in}}%
\pgfpathlineto{\pgfqpoint{1.362875in}{2.052917in}}%
\pgfusepath{stroke}%
\end{pgfscope}%
\begin{pgfscope}%
\pgfsetbuttcap%
\pgfsetroundjoin%
\definecolor{currentfill}{rgb}{0.000000,0.000000,0.000000}%
\pgfsetfillcolor{currentfill}%
\pgfsetlinewidth{0.803000pt}%
\definecolor{currentstroke}{rgb}{0.000000,0.000000,0.000000}%
\pgfsetstrokecolor{currentstroke}%
\pgfsetdash{}{0pt}%
\pgfsys@defobject{currentmarker}{\pgfqpoint{0.000000in}{-0.048611in}}{\pgfqpoint{0.000000in}{0.000000in}}{%
\pgfpathmoveto{\pgfqpoint{0.000000in}{0.000000in}}%
\pgfpathlineto{\pgfqpoint{0.000000in}{-0.048611in}}%
\pgfusepath{stroke,fill}%
}%
\begin{pgfscope}%
\pgfsys@transformshift{1.362875in}{0.501245in}%
\pgfsys@useobject{currentmarker}{}%
\end{pgfscope}%
\end{pgfscope}%
\begin{pgfscope}%
\definecolor{textcolor}{rgb}{0.000000,0.000000,0.000000}%
\pgfsetstrokecolor{textcolor}%
\pgfsetfillcolor{textcolor}%
\pgftext[x=1.362875in,y=0.404023in,,top]{\color{textcolor}\rmfamily\fontsize{10.000000}{12.000000}\selectfont \(\displaystyle {2}\)}%
\end{pgfscope}%
\begin{pgfscope}%
\pgfpathrectangle{\pgfqpoint{0.554481in}{0.501245in}}{\pgfqpoint{1.976074in}{1.551672in}}%
\pgfusepath{clip}%
\pgfsetrectcap%
\pgfsetroundjoin%
\pgfsetlinewidth{0.803000pt}%
\definecolor{currentstroke}{rgb}{0.690196,0.690196,0.690196}%
\pgfsetstrokecolor{currentstroke}%
\pgfsetdash{}{0pt}%
\pgfpathmoveto{\pgfqpoint{1.722161in}{0.501245in}}%
\pgfpathlineto{\pgfqpoint{1.722161in}{2.052917in}}%
\pgfusepath{stroke}%
\end{pgfscope}%
\begin{pgfscope}%
\pgfsetbuttcap%
\pgfsetroundjoin%
\definecolor{currentfill}{rgb}{0.000000,0.000000,0.000000}%
\pgfsetfillcolor{currentfill}%
\pgfsetlinewidth{0.803000pt}%
\definecolor{currentstroke}{rgb}{0.000000,0.000000,0.000000}%
\pgfsetstrokecolor{currentstroke}%
\pgfsetdash{}{0pt}%
\pgfsys@defobject{currentmarker}{\pgfqpoint{0.000000in}{-0.048611in}}{\pgfqpoint{0.000000in}{0.000000in}}{%
\pgfpathmoveto{\pgfqpoint{0.000000in}{0.000000in}}%
\pgfpathlineto{\pgfqpoint{0.000000in}{-0.048611in}}%
\pgfusepath{stroke,fill}%
}%
\begin{pgfscope}%
\pgfsys@transformshift{1.722161in}{0.501245in}%
\pgfsys@useobject{currentmarker}{}%
\end{pgfscope}%
\end{pgfscope}%
\begin{pgfscope}%
\definecolor{textcolor}{rgb}{0.000000,0.000000,0.000000}%
\pgfsetstrokecolor{textcolor}%
\pgfsetfillcolor{textcolor}%
\pgftext[x=1.722161in,y=0.404023in,,top]{\color{textcolor}\rmfamily\fontsize{10.000000}{12.000000}\selectfont \(\displaystyle {3}\)}%
\end{pgfscope}%
\begin{pgfscope}%
\pgfpathrectangle{\pgfqpoint{0.554481in}{0.501245in}}{\pgfqpoint{1.976074in}{1.551672in}}%
\pgfusepath{clip}%
\pgfsetrectcap%
\pgfsetroundjoin%
\pgfsetlinewidth{0.803000pt}%
\definecolor{currentstroke}{rgb}{0.690196,0.690196,0.690196}%
\pgfsetstrokecolor{currentstroke}%
\pgfsetdash{}{0pt}%
\pgfpathmoveto{\pgfqpoint{2.081448in}{0.501245in}}%
\pgfpathlineto{\pgfqpoint{2.081448in}{2.052917in}}%
\pgfusepath{stroke}%
\end{pgfscope}%
\begin{pgfscope}%
\pgfsetbuttcap%
\pgfsetroundjoin%
\definecolor{currentfill}{rgb}{0.000000,0.000000,0.000000}%
\pgfsetfillcolor{currentfill}%
\pgfsetlinewidth{0.803000pt}%
\definecolor{currentstroke}{rgb}{0.000000,0.000000,0.000000}%
\pgfsetstrokecolor{currentstroke}%
\pgfsetdash{}{0pt}%
\pgfsys@defobject{currentmarker}{\pgfqpoint{0.000000in}{-0.048611in}}{\pgfqpoint{0.000000in}{0.000000in}}{%
\pgfpathmoveto{\pgfqpoint{0.000000in}{0.000000in}}%
\pgfpathlineto{\pgfqpoint{0.000000in}{-0.048611in}}%
\pgfusepath{stroke,fill}%
}%
\begin{pgfscope}%
\pgfsys@transformshift{2.081448in}{0.501245in}%
\pgfsys@useobject{currentmarker}{}%
\end{pgfscope}%
\end{pgfscope}%
\begin{pgfscope}%
\definecolor{textcolor}{rgb}{0.000000,0.000000,0.000000}%
\pgfsetstrokecolor{textcolor}%
\pgfsetfillcolor{textcolor}%
\pgftext[x=2.081448in,y=0.404023in,,top]{\color{textcolor}\rmfamily\fontsize{10.000000}{12.000000}\selectfont \(\displaystyle {4}\)}%
\end{pgfscope}%
\begin{pgfscope}%
\pgfpathrectangle{\pgfqpoint{0.554481in}{0.501245in}}{\pgfqpoint{1.976074in}{1.551672in}}%
\pgfusepath{clip}%
\pgfsetrectcap%
\pgfsetroundjoin%
\pgfsetlinewidth{0.803000pt}%
\definecolor{currentstroke}{rgb}{0.690196,0.690196,0.690196}%
\pgfsetstrokecolor{currentstroke}%
\pgfsetdash{}{0pt}%
\pgfpathmoveto{\pgfqpoint{2.440734in}{0.501245in}}%
\pgfpathlineto{\pgfqpoint{2.440734in}{2.052917in}}%
\pgfusepath{stroke}%
\end{pgfscope}%
\begin{pgfscope}%
\pgfsetbuttcap%
\pgfsetroundjoin%
\definecolor{currentfill}{rgb}{0.000000,0.000000,0.000000}%
\pgfsetfillcolor{currentfill}%
\pgfsetlinewidth{0.803000pt}%
\definecolor{currentstroke}{rgb}{0.000000,0.000000,0.000000}%
\pgfsetstrokecolor{currentstroke}%
\pgfsetdash{}{0pt}%
\pgfsys@defobject{currentmarker}{\pgfqpoint{0.000000in}{-0.048611in}}{\pgfqpoint{0.000000in}{0.000000in}}{%
\pgfpathmoveto{\pgfqpoint{0.000000in}{0.000000in}}%
\pgfpathlineto{\pgfqpoint{0.000000in}{-0.048611in}}%
\pgfusepath{stroke,fill}%
}%
\begin{pgfscope}%
\pgfsys@transformshift{2.440734in}{0.501245in}%
\pgfsys@useobject{currentmarker}{}%
\end{pgfscope}%
\end{pgfscope}%
\begin{pgfscope}%
\definecolor{textcolor}{rgb}{0.000000,0.000000,0.000000}%
\pgfsetstrokecolor{textcolor}%
\pgfsetfillcolor{textcolor}%
\pgftext[x=2.440734in,y=0.404023in,,top]{\color{textcolor}\rmfamily\fontsize{10.000000}{12.000000}\selectfont \(\displaystyle {5}\)}%
\end{pgfscope}%
\begin{pgfscope}%
\definecolor{textcolor}{rgb}{0.000000,0.000000,0.000000}%
\pgfsetstrokecolor{textcolor}%
\pgfsetfillcolor{textcolor}%
\pgftext[x=1.542518in,y=0.224234in,,top]{\color{textcolor}\rmfamily\fontsize{10.000000}{12.000000}\selectfont Parameter \(\displaystyle \alpha\)}%
\end{pgfscope}%
\begin{pgfscope}%
\pgfpathrectangle{\pgfqpoint{0.554481in}{0.501245in}}{\pgfqpoint{1.976074in}{1.551672in}}%
\pgfusepath{clip}%
\pgfsetrectcap%
\pgfsetroundjoin%
\pgfsetlinewidth{0.803000pt}%
\definecolor{currentstroke}{rgb}{0.690196,0.690196,0.690196}%
\pgfsetstrokecolor{currentstroke}%
\pgfsetdash{}{0pt}%
\pgfpathmoveto{\pgfqpoint{0.554481in}{0.926307in}}%
\pgfpathlineto{\pgfqpoint{2.530555in}{0.926307in}}%
\pgfusepath{stroke}%
\end{pgfscope}%
\begin{pgfscope}%
\pgfsetbuttcap%
\pgfsetroundjoin%
\definecolor{currentfill}{rgb}{0.000000,0.000000,0.000000}%
\pgfsetfillcolor{currentfill}%
\pgfsetlinewidth{0.803000pt}%
\definecolor{currentstroke}{rgb}{0.000000,0.000000,0.000000}%
\pgfsetstrokecolor{currentstroke}%
\pgfsetdash{}{0pt}%
\pgfsys@defobject{currentmarker}{\pgfqpoint{-0.048611in}{0.000000in}}{\pgfqpoint{-0.000000in}{0.000000in}}{%
\pgfpathmoveto{\pgfqpoint{-0.000000in}{0.000000in}}%
\pgfpathlineto{\pgfqpoint{-0.048611in}{0.000000in}}%
\pgfusepath{stroke,fill}%
}%
\begin{pgfscope}%
\pgfsys@transformshift{0.554481in}{0.926307in}%
\pgfsys@useobject{currentmarker}{}%
\end{pgfscope}%
\end{pgfscope}%
\begin{pgfscope}%
\definecolor{textcolor}{rgb}{0.000000,0.000000,0.000000}%
\pgfsetstrokecolor{textcolor}%
\pgfsetfillcolor{textcolor}%
\pgftext[x=0.279789in, y=0.879225in, left, base]{\color{textcolor}\rmfamily\fontsize{10.000000}{12.000000}\selectfont \(\displaystyle {0.8}\)}%
\end{pgfscope}%
\begin{pgfscope}%
\pgfpathrectangle{\pgfqpoint{0.554481in}{0.501245in}}{\pgfqpoint{1.976074in}{1.551672in}}%
\pgfusepath{clip}%
\pgfsetrectcap%
\pgfsetroundjoin%
\pgfsetlinewidth{0.803000pt}%
\definecolor{currentstroke}{rgb}{0.690196,0.690196,0.690196}%
\pgfsetstrokecolor{currentstroke}%
\pgfsetdash{}{0pt}%
\pgfpathmoveto{\pgfqpoint{0.554481in}{1.489612in}}%
\pgfpathlineto{\pgfqpoint{2.530555in}{1.489612in}}%
\pgfusepath{stroke}%
\end{pgfscope}%
\begin{pgfscope}%
\pgfsetbuttcap%
\pgfsetroundjoin%
\definecolor{currentfill}{rgb}{0.000000,0.000000,0.000000}%
\pgfsetfillcolor{currentfill}%
\pgfsetlinewidth{0.803000pt}%
\definecolor{currentstroke}{rgb}{0.000000,0.000000,0.000000}%
\pgfsetstrokecolor{currentstroke}%
\pgfsetdash{}{0pt}%
\pgfsys@defobject{currentmarker}{\pgfqpoint{-0.048611in}{0.000000in}}{\pgfqpoint{-0.000000in}{0.000000in}}{%
\pgfpathmoveto{\pgfqpoint{-0.000000in}{0.000000in}}%
\pgfpathlineto{\pgfqpoint{-0.048611in}{0.000000in}}%
\pgfusepath{stroke,fill}%
}%
\begin{pgfscope}%
\pgfsys@transformshift{0.554481in}{1.489612in}%
\pgfsys@useobject{currentmarker}{}%
\end{pgfscope}%
\end{pgfscope}%
\begin{pgfscope}%
\definecolor{textcolor}{rgb}{0.000000,0.000000,0.000000}%
\pgfsetstrokecolor{textcolor}%
\pgfsetfillcolor{textcolor}%
\pgftext[x=0.279789in, y=1.442530in, left, base]{\color{textcolor}\rmfamily\fontsize{10.000000}{12.000000}\selectfont \(\displaystyle {0.9}\)}%
\end{pgfscope}%
\begin{pgfscope}%
\pgfpathrectangle{\pgfqpoint{0.554481in}{0.501245in}}{\pgfqpoint{1.976074in}{1.551672in}}%
\pgfusepath{clip}%
\pgfsetrectcap%
\pgfsetroundjoin%
\pgfsetlinewidth{0.803000pt}%
\definecolor{currentstroke}{rgb}{0.690196,0.690196,0.690196}%
\pgfsetstrokecolor{currentstroke}%
\pgfsetdash{}{0pt}%
\pgfpathmoveto{\pgfqpoint{0.554481in}{2.052917in}}%
\pgfpathlineto{\pgfqpoint{2.530555in}{2.052917in}}%
\pgfusepath{stroke}%
\end{pgfscope}%
\begin{pgfscope}%
\pgfsetbuttcap%
\pgfsetroundjoin%
\definecolor{currentfill}{rgb}{0.000000,0.000000,0.000000}%
\pgfsetfillcolor{currentfill}%
\pgfsetlinewidth{0.803000pt}%
\definecolor{currentstroke}{rgb}{0.000000,0.000000,0.000000}%
\pgfsetstrokecolor{currentstroke}%
\pgfsetdash{}{0pt}%
\pgfsys@defobject{currentmarker}{\pgfqpoint{-0.048611in}{0.000000in}}{\pgfqpoint{-0.000000in}{0.000000in}}{%
\pgfpathmoveto{\pgfqpoint{-0.000000in}{0.000000in}}%
\pgfpathlineto{\pgfqpoint{-0.048611in}{0.000000in}}%
\pgfusepath{stroke,fill}%
}%
\begin{pgfscope}%
\pgfsys@transformshift{0.554481in}{2.052917in}%
\pgfsys@useobject{currentmarker}{}%
\end{pgfscope}%
\end{pgfscope}%
\begin{pgfscope}%
\definecolor{textcolor}{rgb}{0.000000,0.000000,0.000000}%
\pgfsetstrokecolor{textcolor}%
\pgfsetfillcolor{textcolor}%
\pgftext[x=0.279789in, y=2.005835in, left, base]{\color{textcolor}\rmfamily\fontsize{10.000000}{12.000000}\selectfont \(\displaystyle {1.0}\)}%
\end{pgfscope}%
\begin{pgfscope}%
\definecolor{textcolor}{rgb}{0.000000,0.000000,0.000000}%
\pgfsetstrokecolor{textcolor}%
\pgfsetfillcolor{textcolor}%
\pgftext[x=0.224234in,y=1.277081in,,bottom,rotate=90.000000]{\color{textcolor}\rmfamily\fontsize{10.000000}{12.000000}\selectfont Pearson correlation}%
\end{pgfscope}%
\begin{pgfscope}%
\pgfpathrectangle{\pgfqpoint{0.554481in}{0.501245in}}{\pgfqpoint{1.976074in}{1.551672in}}%
\pgfusepath{clip}%
\pgfsetbuttcap%
\pgfsetroundjoin%
\pgfsetlinewidth{0.301125pt}%
\definecolor{currentstroke}{rgb}{0.215686,0.494118,0.721569}%
\pgfsetstrokecolor{currentstroke}%
\pgfsetdash{{1.110000pt}{0.480000pt}}{0.000000pt}%
\pgfpathmoveto{\pgfqpoint{0.644303in}{0.564310in}}%
\pgfpathlineto{\pgfqpoint{2.440734in}{0.564310in}}%
\pgfpathlineto{\pgfqpoint{2.440734in}{0.564310in}}%
\pgfusepath{stroke}%
\end{pgfscope}%
\begin{pgfscope}%
\pgfpathrectangle{\pgfqpoint{0.554481in}{0.501245in}}{\pgfqpoint{1.976074in}{1.551672in}}%
\pgfusepath{clip}%
\pgfsetbuttcap%
\pgfsetbeveljoin%
\definecolor{currentfill}{rgb}{0.000000,0.000000,0.000000}%
\pgfsetfillcolor{currentfill}%
\pgfsetfillopacity{0.000000}%
\pgfsetlinewidth{0.501875pt}%
\definecolor{currentstroke}{rgb}{0.215686,0.494118,0.721569}%
\pgfsetstrokecolor{currentstroke}%
\pgfsetdash{}{0pt}%
\pgfsys@defobject{currentmarker}{\pgfqpoint{-0.033023in}{-0.028091in}}{\pgfqpoint{0.033023in}{0.034722in}}{%
\pgfpathmoveto{\pgfqpoint{0.000000in}{0.034722in}}%
\pgfpathlineto{\pgfqpoint{-0.007796in}{0.010730in}}%
\pgfpathlineto{\pgfqpoint{-0.033023in}{0.010730in}}%
\pgfpathlineto{\pgfqpoint{-0.012614in}{-0.004098in}}%
\pgfpathlineto{\pgfqpoint{-0.020409in}{-0.028091in}}%
\pgfpathlineto{\pgfqpoint{-0.000000in}{-0.013263in}}%
\pgfpathlineto{\pgfqpoint{0.020409in}{-0.028091in}}%
\pgfpathlineto{\pgfqpoint{0.012614in}{-0.004098in}}%
\pgfpathlineto{\pgfqpoint{0.033023in}{0.010730in}}%
\pgfpathlineto{\pgfqpoint{0.007796in}{0.010730in}}%
\pgfpathlineto{\pgfqpoint{0.000000in}{0.034722in}}%
\pgfpathclose%
\pgfusepath{stroke,fill}%
}%
\begin{pgfscope}%
\pgfsys@transformshift{0.644303in}{0.564310in}%
\pgfsys@useobject{currentmarker}{}%
\end{pgfscope}%
\end{pgfscope}%
\begin{pgfscope}%
\pgfpathrectangle{\pgfqpoint{0.554481in}{0.501245in}}{\pgfqpoint{1.976074in}{1.551672in}}%
\pgfusepath{clip}%
\pgfsetrectcap%
\pgfsetroundjoin%
\pgfsetlinewidth{0.602250pt}%
\definecolor{currentstroke}{rgb}{1.000000,0.498039,0.000000}%
\pgfsetstrokecolor{currentstroke}%
\pgfsetdash{}{0pt}%
\pgfpathmoveto{\pgfqpoint{0.644303in}{0.564311in}}%
\pgfpathlineto{\pgfqpoint{0.671385in}{0.676825in}}%
\pgfpathlineto{\pgfqpoint{0.707494in}{0.829918in}}%
\pgfpathlineto{\pgfqpoint{0.752631in}{1.021082in}}%
\pgfpathlineto{\pgfqpoint{0.770685in}{1.095788in}}%
\pgfpathlineto{\pgfqpoint{0.788740in}{1.168630in}}%
\pgfpathlineto{\pgfqpoint{0.806794in}{1.239026in}}%
\pgfpathlineto{\pgfqpoint{0.824849in}{1.306406in}}%
\pgfpathlineto{\pgfqpoint{0.842903in}{1.370233in}}%
\pgfpathlineto{\pgfqpoint{0.851931in}{1.400660in}}%
\pgfpathlineto{\pgfqpoint{0.860958in}{1.430020in}}%
\pgfpathlineto{\pgfqpoint{0.869985in}{1.458264in}}%
\pgfpathlineto{\pgfqpoint{0.879013in}{1.485345in}}%
\pgfpathlineto{\pgfqpoint{0.888040in}{1.511224in}}%
\pgfpathlineto{\pgfqpoint{0.897067in}{1.535867in}}%
\pgfpathlineto{\pgfqpoint{0.906094in}{1.559245in}}%
\pgfpathlineto{\pgfqpoint{0.915122in}{1.581337in}}%
\pgfpathlineto{\pgfqpoint{0.924149in}{1.602127in}}%
\pgfpathlineto{\pgfqpoint{0.933176in}{1.621606in}}%
\pgfpathlineto{\pgfqpoint{0.942204in}{1.639772in}}%
\pgfpathlineto{\pgfqpoint{0.951231in}{1.656627in}}%
\pgfpathlineto{\pgfqpoint{0.960258in}{1.672181in}}%
\pgfpathlineto{\pgfqpoint{0.969285in}{1.686449in}}%
\pgfpathlineto{\pgfqpoint{0.978313in}{1.699453in}}%
\pgfpathlineto{\pgfqpoint{0.987340in}{1.711219in}}%
\pgfpathlineto{\pgfqpoint{0.996367in}{1.721776in}}%
\pgfpathlineto{\pgfqpoint{1.005395in}{1.731161in}}%
\pgfpathlineto{\pgfqpoint{1.014422in}{1.739412in}}%
\pgfpathlineto{\pgfqpoint{1.023449in}{1.746571in}}%
\pgfpathlineto{\pgfqpoint{1.032477in}{1.752684in}}%
\pgfpathlineto{\pgfqpoint{1.041504in}{1.757799in}}%
\pgfpathlineto{\pgfqpoint{1.050531in}{1.761963in}}%
\pgfpathlineto{\pgfqpoint{1.059558in}{1.765230in}}%
\pgfpathlineto{\pgfqpoint{1.068586in}{1.767650in}}%
\pgfpathlineto{\pgfqpoint{1.077613in}{1.769276in}}%
\pgfpathlineto{\pgfqpoint{1.086640in}{1.770160in}}%
\pgfpathlineto{\pgfqpoint{1.095668in}{1.770355in}}%
\pgfpathlineto{\pgfqpoint{1.104695in}{1.769913in}}%
\pgfpathlineto{\pgfqpoint{1.113722in}{1.768884in}}%
\pgfpathlineto{\pgfqpoint{1.122749in}{1.767318in}}%
\pgfpathlineto{\pgfqpoint{1.131777in}{1.765264in}}%
\pgfpathlineto{\pgfqpoint{1.140804in}{1.762768in}}%
\pgfpathlineto{\pgfqpoint{1.149831in}{1.759876in}}%
\pgfpathlineto{\pgfqpoint{1.158859in}{1.756630in}}%
\pgfpathlineto{\pgfqpoint{1.176913in}{1.749242in}}%
\pgfpathlineto{\pgfqpoint{1.194968in}{1.740909in}}%
\pgfpathlineto{\pgfqpoint{1.213022in}{1.731901in}}%
\pgfpathlineto{\pgfqpoint{1.240104in}{1.717635in}}%
\pgfpathlineto{\pgfqpoint{1.303295in}{1.683952in}}%
\pgfpathlineto{\pgfqpoint{1.330377in}{1.670379in}}%
\pgfpathlineto{\pgfqpoint{1.348432in}{1.661827in}}%
\pgfpathlineto{\pgfqpoint{1.366486in}{1.653733in}}%
\pgfpathlineto{\pgfqpoint{1.384541in}{1.646130in}}%
\pgfpathlineto{\pgfqpoint{1.402595in}{1.639040in}}%
\pgfpathlineto{\pgfqpoint{1.420650in}{1.632473in}}%
\pgfpathlineto{\pgfqpoint{1.438705in}{1.626433in}}%
\pgfpathlineto{\pgfqpoint{1.456759in}{1.620913in}}%
\pgfpathlineto{\pgfqpoint{1.474814in}{1.615900in}}%
\pgfpathlineto{\pgfqpoint{1.492868in}{1.611377in}}%
\pgfpathlineto{\pgfqpoint{1.510923in}{1.607324in}}%
\pgfpathlineto{\pgfqpoint{1.528978in}{1.603715in}}%
\pgfpathlineto{\pgfqpoint{1.547032in}{1.600524in}}%
\pgfpathlineto{\pgfqpoint{1.574114in}{1.596457in}}%
\pgfpathlineto{\pgfqpoint{1.601196in}{1.593168in}}%
\pgfpathlineto{\pgfqpoint{1.628278in}{1.590555in}}%
\pgfpathlineto{\pgfqpoint{1.655360in}{1.588524in}}%
\pgfpathlineto{\pgfqpoint{1.682441in}{1.586984in}}%
\pgfpathlineto{\pgfqpoint{1.718551in}{1.585554in}}%
\pgfpathlineto{\pgfqpoint{1.754660in}{1.584679in}}%
\pgfpathlineto{\pgfqpoint{1.799796in}{1.584149in}}%
\pgfpathlineto{\pgfqpoint{1.862987in}{1.584081in}}%
\pgfpathlineto{\pgfqpoint{1.953260in}{1.584648in}}%
\pgfpathlineto{\pgfqpoint{2.287270in}{1.587188in}}%
\pgfpathlineto{\pgfqpoint{2.440734in}{1.587862in}}%
\pgfpathlineto{\pgfqpoint{2.440734in}{1.587862in}}%
\pgfusepath{stroke}%
\end{pgfscope}%
\begin{pgfscope}%
\pgfpathrectangle{\pgfqpoint{0.554481in}{0.501245in}}{\pgfqpoint{1.976074in}{1.551672in}}%
\pgfusepath{clip}%
\pgfsetbuttcap%
\pgfsetbeveljoin%
\definecolor{currentfill}{rgb}{0.000000,0.000000,0.000000}%
\pgfsetfillcolor{currentfill}%
\pgfsetfillopacity{0.000000}%
\pgfsetlinewidth{0.501875pt}%
\definecolor{currentstroke}{rgb}{1.000000,0.498039,0.000000}%
\pgfsetstrokecolor{currentstroke}%
\pgfsetdash{}{0pt}%
\pgfsys@defobject{currentmarker}{\pgfqpoint{-0.033023in}{-0.028091in}}{\pgfqpoint{0.033023in}{0.034722in}}{%
\pgfpathmoveto{\pgfqpoint{0.000000in}{0.034722in}}%
\pgfpathlineto{\pgfqpoint{-0.007796in}{0.010730in}}%
\pgfpathlineto{\pgfqpoint{-0.033023in}{0.010730in}}%
\pgfpathlineto{\pgfqpoint{-0.012614in}{-0.004098in}}%
\pgfpathlineto{\pgfqpoint{-0.020409in}{-0.028091in}}%
\pgfpathlineto{\pgfqpoint{-0.000000in}{-0.013263in}}%
\pgfpathlineto{\pgfqpoint{0.020409in}{-0.028091in}}%
\pgfpathlineto{\pgfqpoint{0.012614in}{-0.004098in}}%
\pgfpathlineto{\pgfqpoint{0.033023in}{0.010730in}}%
\pgfpathlineto{\pgfqpoint{0.007796in}{0.010730in}}%
\pgfpathlineto{\pgfqpoint{0.000000in}{0.034722in}}%
\pgfpathclose%
\pgfusepath{stroke,fill}%
}%
\begin{pgfscope}%
\pgfsys@transformshift{1.095668in}{1.770355in}%
\pgfsys@useobject{currentmarker}{}%
\end{pgfscope}%
\end{pgfscope}%
\begin{pgfscope}%
\pgfpathrectangle{\pgfqpoint{0.554481in}{0.501245in}}{\pgfqpoint{1.976074in}{1.551672in}}%
\pgfusepath{clip}%
\pgfsetbuttcap%
\pgfsetroundjoin%
\pgfsetlinewidth{0.903375pt}%
\definecolor{currentstroke}{rgb}{0.301961,0.686275,0.290196}%
\pgfsetstrokecolor{currentstroke}%
\pgfsetdash{{0.900000pt}{1.485000pt}}{0.000000pt}%
\pgfpathmoveto{\pgfqpoint{0.644303in}{0.564311in}}%
\pgfpathlineto{\pgfqpoint{0.671385in}{0.676991in}}%
\pgfpathlineto{\pgfqpoint{0.698467in}{0.792830in}}%
\pgfpathlineto{\pgfqpoint{0.725549in}{0.911412in}}%
\pgfpathlineto{\pgfqpoint{0.761658in}{1.072414in}}%
\pgfpathlineto{\pgfqpoint{0.797767in}{1.233617in}}%
\pgfpathlineto{\pgfqpoint{0.815822in}{1.312649in}}%
\pgfpathlineto{\pgfqpoint{0.833876in}{1.389407in}}%
\pgfpathlineto{\pgfqpoint{0.851931in}{1.462740in}}%
\pgfpathlineto{\pgfqpoint{0.860958in}{1.497718in}}%
\pgfpathlineto{\pgfqpoint{0.869985in}{1.531338in}}%
\pgfpathlineto{\pgfqpoint{0.879013in}{1.563420in}}%
\pgfpathlineto{\pgfqpoint{0.888040in}{1.593782in}}%
\pgfpathlineto{\pgfqpoint{0.897067in}{1.622245in}}%
\pgfpathlineto{\pgfqpoint{0.906094in}{1.648637in}}%
\pgfpathlineto{\pgfqpoint{0.915122in}{1.672797in}}%
\pgfpathlineto{\pgfqpoint{0.924149in}{1.694581in}}%
\pgfpathlineto{\pgfqpoint{0.933176in}{1.713867in}}%
\pgfpathlineto{\pgfqpoint{0.942204in}{1.730559in}}%
\pgfpathlineto{\pgfqpoint{0.951231in}{1.744594in}}%
\pgfpathlineto{\pgfqpoint{0.960258in}{1.755943in}}%
\pgfpathlineto{\pgfqpoint{0.969285in}{1.764616in}}%
\pgfpathlineto{\pgfqpoint{0.978313in}{1.770665in}}%
\pgfpathlineto{\pgfqpoint{0.987340in}{1.774182in}}%
\pgfpathlineto{\pgfqpoint{0.996367in}{1.775295in}}%
\pgfpathlineto{\pgfqpoint{1.005395in}{1.774174in}}%
\pgfpathlineto{\pgfqpoint{1.014422in}{1.771017in}}%
\pgfpathlineto{\pgfqpoint{1.023449in}{1.766050in}}%
\pgfpathlineto{\pgfqpoint{1.032477in}{1.759518in}}%
\pgfpathlineto{\pgfqpoint{1.041504in}{1.751681in}}%
\pgfpathlineto{\pgfqpoint{1.050531in}{1.742799in}}%
\pgfpathlineto{\pgfqpoint{1.059558in}{1.733134in}}%
\pgfpathlineto{\pgfqpoint{1.077613in}{1.712435in}}%
\pgfpathlineto{\pgfqpoint{1.095668in}{1.691364in}}%
\pgfpathlineto{\pgfqpoint{1.113722in}{1.671319in}}%
\pgfpathlineto{\pgfqpoint{1.122749in}{1.661996in}}%
\pgfpathlineto{\pgfqpoint{1.131777in}{1.653253in}}%
\pgfpathlineto{\pgfqpoint{1.140804in}{1.645144in}}%
\pgfpathlineto{\pgfqpoint{1.149831in}{1.637702in}}%
\pgfpathlineto{\pgfqpoint{1.158859in}{1.630937in}}%
\pgfpathlineto{\pgfqpoint{1.167886in}{1.624846in}}%
\pgfpathlineto{\pgfqpoint{1.176913in}{1.619411in}}%
\pgfpathlineto{\pgfqpoint{1.185940in}{1.614607in}}%
\pgfpathlineto{\pgfqpoint{1.194968in}{1.610396in}}%
\pgfpathlineto{\pgfqpoint{1.203995in}{1.606739in}}%
\pgfpathlineto{\pgfqpoint{1.213022in}{1.603590in}}%
\pgfpathlineto{\pgfqpoint{1.222050in}{1.600905in}}%
\pgfpathlineto{\pgfqpoint{1.231077in}{1.598636in}}%
\pgfpathlineto{\pgfqpoint{1.240104in}{1.596737in}}%
\pgfpathlineto{\pgfqpoint{1.249131in}{1.595164in}}%
\pgfpathlineto{\pgfqpoint{1.267186in}{1.592827in}}%
\pgfpathlineto{\pgfqpoint{1.285241in}{1.591323in}}%
\pgfpathlineto{\pgfqpoint{1.303295in}{1.590401in}}%
\pgfpathlineto{\pgfqpoint{1.330377in}{1.589697in}}%
\pgfpathlineto{\pgfqpoint{1.375514in}{1.589362in}}%
\pgfpathlineto{\pgfqpoint{1.538005in}{1.589339in}}%
\pgfpathlineto{\pgfqpoint{2.440734in}{1.589339in}}%
\pgfpathlineto{\pgfqpoint{2.440734in}{1.589339in}}%
\pgfusepath{stroke}%
\end{pgfscope}%
\begin{pgfscope}%
\pgfpathrectangle{\pgfqpoint{0.554481in}{0.501245in}}{\pgfqpoint{1.976074in}{1.551672in}}%
\pgfusepath{clip}%
\pgfsetbuttcap%
\pgfsetbeveljoin%
\definecolor{currentfill}{rgb}{0.000000,0.000000,0.000000}%
\pgfsetfillcolor{currentfill}%
\pgfsetfillopacity{0.000000}%
\pgfsetlinewidth{0.501875pt}%
\definecolor{currentstroke}{rgb}{0.301961,0.686275,0.290196}%
\pgfsetstrokecolor{currentstroke}%
\pgfsetdash{}{0pt}%
\pgfsys@defobject{currentmarker}{\pgfqpoint{-0.033023in}{-0.028091in}}{\pgfqpoint{0.033023in}{0.034722in}}{%
\pgfpathmoveto{\pgfqpoint{0.000000in}{0.034722in}}%
\pgfpathlineto{\pgfqpoint{-0.007796in}{0.010730in}}%
\pgfpathlineto{\pgfqpoint{-0.033023in}{0.010730in}}%
\pgfpathlineto{\pgfqpoint{-0.012614in}{-0.004098in}}%
\pgfpathlineto{\pgfqpoint{-0.020409in}{-0.028091in}}%
\pgfpathlineto{\pgfqpoint{-0.000000in}{-0.013263in}}%
\pgfpathlineto{\pgfqpoint{0.020409in}{-0.028091in}}%
\pgfpathlineto{\pgfqpoint{0.012614in}{-0.004098in}}%
\pgfpathlineto{\pgfqpoint{0.033023in}{0.010730in}}%
\pgfpathlineto{\pgfqpoint{0.007796in}{0.010730in}}%
\pgfpathlineto{\pgfqpoint{0.000000in}{0.034722in}}%
\pgfpathclose%
\pgfusepath{stroke,fill}%
}%
\begin{pgfscope}%
\pgfsys@transformshift{0.996367in}{1.775295in}%
\pgfsys@useobject{currentmarker}{}%
\end{pgfscope}%
\end{pgfscope}%
\begin{pgfscope}%
\pgfpathrectangle{\pgfqpoint{0.554481in}{0.501245in}}{\pgfqpoint{1.976074in}{1.551672in}}%
\pgfusepath{clip}%
\pgfsetbuttcap%
\pgfsetroundjoin%
\pgfsetlinewidth{1.204500pt}%
\definecolor{currentstroke}{rgb}{0.968627,0.505882,0.749020}%
\pgfsetstrokecolor{currentstroke}%
\pgfsetdash{{7.680000pt}{1.920000pt}{1.200000pt}{1.920000pt}}{0.000000pt}%
\pgfpathmoveto{\pgfqpoint{0.644303in}{0.564311in}}%
\pgfpathlineto{\pgfqpoint{0.671385in}{0.676992in}}%
\pgfpathlineto{\pgfqpoint{0.698467in}{0.792839in}}%
\pgfpathlineto{\pgfqpoint{0.725549in}{0.911490in}}%
\pgfpathlineto{\pgfqpoint{0.761658in}{1.072930in}}%
\pgfpathlineto{\pgfqpoint{0.815822in}{1.316122in}}%
\pgfpathlineto{\pgfqpoint{0.833876in}{1.394977in}}%
\pgfpathlineto{\pgfqpoint{0.851931in}{1.471082in}}%
\pgfpathlineto{\pgfqpoint{0.860958in}{1.507673in}}%
\pgfpathlineto{\pgfqpoint{0.869985in}{1.543016in}}%
\pgfpathlineto{\pgfqpoint{0.879013in}{1.576880in}}%
\pgfpathlineto{\pgfqpoint{0.888040in}{1.609009in}}%
\pgfpathlineto{\pgfqpoint{0.897067in}{1.639130in}}%
\pgfpathlineto{\pgfqpoint{0.906094in}{1.666952in}}%
\pgfpathlineto{\pgfqpoint{0.915122in}{1.692175in}}%
\pgfpathlineto{\pgfqpoint{0.924149in}{1.714499in}}%
\pgfpathlineto{\pgfqpoint{0.933176in}{1.733632in}}%
\pgfpathlineto{\pgfqpoint{0.942204in}{1.749311in}}%
\pgfpathlineto{\pgfqpoint{0.951231in}{1.761317in}}%
\pgfpathlineto{\pgfqpoint{0.960258in}{1.769495in}}%
\pgfpathlineto{\pgfqpoint{0.969285in}{1.773779in}}%
\pgfpathlineto{\pgfqpoint{0.978313in}{1.774204in}}%
\pgfpathlineto{\pgfqpoint{0.987340in}{1.770929in}}%
\pgfpathlineto{\pgfqpoint{0.996367in}{1.764241in}}%
\pgfpathlineto{\pgfqpoint{1.005395in}{1.754555in}}%
\pgfpathlineto{\pgfqpoint{1.014422in}{1.742398in}}%
\pgfpathlineto{\pgfqpoint{1.023449in}{1.728384in}}%
\pgfpathlineto{\pgfqpoint{1.041504in}{1.697446in}}%
\pgfpathlineto{\pgfqpoint{1.050531in}{1.681819in}}%
\pgfpathlineto{\pgfqpoint{1.059558in}{1.666842in}}%
\pgfpathlineto{\pgfqpoint{1.068586in}{1.652952in}}%
\pgfpathlineto{\pgfqpoint{1.077613in}{1.640455in}}%
\pgfpathlineto{\pgfqpoint{1.086640in}{1.629530in}}%
\pgfpathlineto{\pgfqpoint{1.095668in}{1.620239in}}%
\pgfpathlineto{\pgfqpoint{1.104695in}{1.612547in}}%
\pgfpathlineto{\pgfqpoint{1.113722in}{1.606348in}}%
\pgfpathlineto{\pgfqpoint{1.122749in}{1.601485in}}%
\pgfpathlineto{\pgfqpoint{1.131777in}{1.597776in}}%
\pgfpathlineto{\pgfqpoint{1.140804in}{1.595029in}}%
\pgfpathlineto{\pgfqpoint{1.149831in}{1.593057in}}%
\pgfpathlineto{\pgfqpoint{1.158859in}{1.591688in}}%
\pgfpathlineto{\pgfqpoint{1.167886in}{1.590769in}}%
\pgfpathlineto{\pgfqpoint{1.185940in}{1.589808in}}%
\pgfpathlineto{\pgfqpoint{1.213022in}{1.589400in}}%
\pgfpathlineto{\pgfqpoint{1.303295in}{1.589339in}}%
\pgfpathlineto{\pgfqpoint{2.440734in}{1.589339in}}%
\pgfpathlineto{\pgfqpoint{2.440734in}{1.589339in}}%
\pgfusepath{stroke}%
\end{pgfscope}%
\begin{pgfscope}%
\pgfpathrectangle{\pgfqpoint{0.554481in}{0.501245in}}{\pgfqpoint{1.976074in}{1.551672in}}%
\pgfusepath{clip}%
\pgfsetbuttcap%
\pgfsetbeveljoin%
\definecolor{currentfill}{rgb}{0.000000,0.000000,0.000000}%
\pgfsetfillcolor{currentfill}%
\pgfsetfillopacity{0.000000}%
\pgfsetlinewidth{0.501875pt}%
\definecolor{currentstroke}{rgb}{0.968627,0.505882,0.749020}%
\pgfsetstrokecolor{currentstroke}%
\pgfsetdash{}{0pt}%
\pgfsys@defobject{currentmarker}{\pgfqpoint{-0.033023in}{-0.028091in}}{\pgfqpoint{0.033023in}{0.034722in}}{%
\pgfpathmoveto{\pgfqpoint{0.000000in}{0.034722in}}%
\pgfpathlineto{\pgfqpoint{-0.007796in}{0.010730in}}%
\pgfpathlineto{\pgfqpoint{-0.033023in}{0.010730in}}%
\pgfpathlineto{\pgfqpoint{-0.012614in}{-0.004098in}}%
\pgfpathlineto{\pgfqpoint{-0.020409in}{-0.028091in}}%
\pgfpathlineto{\pgfqpoint{-0.000000in}{-0.013263in}}%
\pgfpathlineto{\pgfqpoint{0.020409in}{-0.028091in}}%
\pgfpathlineto{\pgfqpoint{0.012614in}{-0.004098in}}%
\pgfpathlineto{\pgfqpoint{0.033023in}{0.010730in}}%
\pgfpathlineto{\pgfqpoint{0.007796in}{0.010730in}}%
\pgfpathlineto{\pgfqpoint{0.000000in}{0.034722in}}%
\pgfpathclose%
\pgfusepath{stroke,fill}%
}%
\begin{pgfscope}%
\pgfsys@transformshift{0.978313in}{1.774204in}%
\pgfsys@useobject{currentmarker}{}%
\end{pgfscope}%
\end{pgfscope}%
\begin{pgfscope}%
\pgfpathrectangle{\pgfqpoint{0.554481in}{0.501245in}}{\pgfqpoint{1.976074in}{1.551672in}}%
\pgfusepath{clip}%
\pgfsetbuttcap%
\pgfsetroundjoin%
\pgfsetlinewidth{1.505625pt}%
\definecolor{currentstroke}{rgb}{0.650980,0.337255,0.156863}%
\pgfsetstrokecolor{currentstroke}%
\pgfsetdash{{5.550000pt}{2.400000pt}}{0.000000pt}%
\pgfpathmoveto{\pgfqpoint{0.644303in}{0.564311in}}%
\pgfpathlineto{\pgfqpoint{0.671385in}{0.676992in}}%
\pgfpathlineto{\pgfqpoint{0.698467in}{0.792839in}}%
\pgfpathlineto{\pgfqpoint{0.725549in}{0.911491in}}%
\pgfpathlineto{\pgfqpoint{0.761658in}{1.072951in}}%
\pgfpathlineto{\pgfqpoint{0.815822in}{1.316460in}}%
\pgfpathlineto{\pgfqpoint{0.833876in}{1.395660in}}%
\pgfpathlineto{\pgfqpoint{0.851931in}{1.472350in}}%
\pgfpathlineto{\pgfqpoint{0.869985in}{1.545179in}}%
\pgfpathlineto{\pgfqpoint{0.879013in}{1.579612in}}%
\pgfpathlineto{\pgfqpoint{0.888040in}{1.612381in}}%
\pgfpathlineto{\pgfqpoint{0.897067in}{1.643182in}}%
\pgfpathlineto{\pgfqpoint{0.906094in}{1.671680in}}%
\pgfpathlineto{\pgfqpoint{0.915122in}{1.697501in}}%
\pgfpathlineto{\pgfqpoint{0.924149in}{1.720240in}}%
\pgfpathlineto{\pgfqpoint{0.933176in}{1.739471in}}%
\pgfpathlineto{\pgfqpoint{0.942204in}{1.754765in}}%
\pgfpathlineto{\pgfqpoint{0.951231in}{1.765714in}}%
\pgfpathlineto{\pgfqpoint{0.960258in}{1.771969in}}%
\pgfpathlineto{\pgfqpoint{0.969285in}{1.773298in}}%
\pgfpathlineto{\pgfqpoint{0.978313in}{1.769637in}}%
\pgfpathlineto{\pgfqpoint{0.987340in}{1.761161in}}%
\pgfpathlineto{\pgfqpoint{0.996367in}{1.748330in}}%
\pgfpathlineto{\pgfqpoint{1.005395in}{1.731917in}}%
\pgfpathlineto{\pgfqpoint{1.014422in}{1.712981in}}%
\pgfpathlineto{\pgfqpoint{1.032477in}{1.672638in}}%
\pgfpathlineto{\pgfqpoint{1.041504in}{1.653755in}}%
\pgfpathlineto{\pgfqpoint{1.050531in}{1.637066in}}%
\pgfpathlineto{\pgfqpoint{1.059558in}{1.623136in}}%
\pgfpathlineto{\pgfqpoint{1.068586in}{1.612144in}}%
\pgfpathlineto{\pgfqpoint{1.077613in}{1.603949in}}%
\pgfpathlineto{\pgfqpoint{1.086640in}{1.598186in}}%
\pgfpathlineto{\pgfqpoint{1.095668in}{1.594375in}}%
\pgfpathlineto{\pgfqpoint{1.104695in}{1.592017in}}%
\pgfpathlineto{\pgfqpoint{1.113722in}{1.590659in}}%
\pgfpathlineto{\pgfqpoint{1.122749in}{1.589938in}}%
\pgfpathlineto{\pgfqpoint{1.140804in}{1.589431in}}%
\pgfpathlineto{\pgfqpoint{1.194968in}{1.589339in}}%
\pgfpathlineto{\pgfqpoint{2.440734in}{1.589339in}}%
\pgfpathlineto{\pgfqpoint{2.440734in}{1.589339in}}%
\pgfusepath{stroke}%
\end{pgfscope}%
\begin{pgfscope}%
\pgfpathrectangle{\pgfqpoint{0.554481in}{0.501245in}}{\pgfqpoint{1.976074in}{1.551672in}}%
\pgfusepath{clip}%
\pgfsetbuttcap%
\pgfsetbeveljoin%
\definecolor{currentfill}{rgb}{0.000000,0.000000,0.000000}%
\pgfsetfillcolor{currentfill}%
\pgfsetfillopacity{0.000000}%
\pgfsetlinewidth{0.501875pt}%
\definecolor{currentstroke}{rgb}{0.650980,0.337255,0.156863}%
\pgfsetstrokecolor{currentstroke}%
\pgfsetdash{}{0pt}%
\pgfsys@defobject{currentmarker}{\pgfqpoint{-0.033023in}{-0.028091in}}{\pgfqpoint{0.033023in}{0.034722in}}{%
\pgfpathmoveto{\pgfqpoint{0.000000in}{0.034722in}}%
\pgfpathlineto{\pgfqpoint{-0.007796in}{0.010730in}}%
\pgfpathlineto{\pgfqpoint{-0.033023in}{0.010730in}}%
\pgfpathlineto{\pgfqpoint{-0.012614in}{-0.004098in}}%
\pgfpathlineto{\pgfqpoint{-0.020409in}{-0.028091in}}%
\pgfpathlineto{\pgfqpoint{-0.000000in}{-0.013263in}}%
\pgfpathlineto{\pgfqpoint{0.020409in}{-0.028091in}}%
\pgfpathlineto{\pgfqpoint{0.012614in}{-0.004098in}}%
\pgfpathlineto{\pgfqpoint{0.033023in}{0.010730in}}%
\pgfpathlineto{\pgfqpoint{0.007796in}{0.010730in}}%
\pgfpathlineto{\pgfqpoint{0.000000in}{0.034722in}}%
\pgfpathclose%
\pgfusepath{stroke,fill}%
}%
\begin{pgfscope}%
\pgfsys@transformshift{0.969285in}{1.773298in}%
\pgfsys@useobject{currentmarker}{}%
\end{pgfscope}%
\end{pgfscope}%
\begin{pgfscope}%
\pgfpathrectangle{\pgfqpoint{0.554481in}{0.501245in}}{\pgfqpoint{1.976074in}{1.551672in}}%
\pgfusepath{clip}%
\pgfsetbuttcap%
\pgfsetroundjoin%
\pgfsetlinewidth{0.803000pt}%
\definecolor{currentstroke}{rgb}{1.000000,0.000000,0.000000}%
\pgfsetstrokecolor{currentstroke}%
\pgfsetdash{{2.960000pt}{1.280000pt}}{0.000000pt}%
\pgfpathmoveto{\pgfqpoint{0.554481in}{1.825608in}}%
\pgfpathlineto{\pgfqpoint{2.530555in}{1.825608in}}%
\pgfusepath{stroke}%
\end{pgfscope}%
\begin{pgfscope}%
\pgfsetrectcap%
\pgfsetmiterjoin%
\pgfsetlinewidth{0.803000pt}%
\definecolor{currentstroke}{rgb}{0.000000,0.000000,0.000000}%
\pgfsetstrokecolor{currentstroke}%
\pgfsetdash{}{0pt}%
\pgfpathmoveto{\pgfqpoint{0.554481in}{0.501245in}}%
\pgfpathlineto{\pgfqpoint{0.554481in}{2.052917in}}%
\pgfusepath{stroke}%
\end{pgfscope}%
\begin{pgfscope}%
\pgfsetrectcap%
\pgfsetmiterjoin%
\pgfsetlinewidth{0.803000pt}%
\definecolor{currentstroke}{rgb}{0.000000,0.000000,0.000000}%
\pgfsetstrokecolor{currentstroke}%
\pgfsetdash{}{0pt}%
\pgfpathmoveto{\pgfqpoint{2.530555in}{0.501245in}}%
\pgfpathlineto{\pgfqpoint{2.530555in}{2.052917in}}%
\pgfusepath{stroke}%
\end{pgfscope}%
\begin{pgfscope}%
\pgfsetrectcap%
\pgfsetmiterjoin%
\pgfsetlinewidth{0.803000pt}%
\definecolor{currentstroke}{rgb}{0.000000,0.000000,0.000000}%
\pgfsetstrokecolor{currentstroke}%
\pgfsetdash{}{0pt}%
\pgfpathmoveto{\pgfqpoint{0.554481in}{0.501245in}}%
\pgfpathlineto{\pgfqpoint{2.530555in}{0.501245in}}%
\pgfusepath{stroke}%
\end{pgfscope}%
\begin{pgfscope}%
\pgfsetrectcap%
\pgfsetmiterjoin%
\pgfsetlinewidth{0.803000pt}%
\definecolor{currentstroke}{rgb}{0.000000,0.000000,0.000000}%
\pgfsetstrokecolor{currentstroke}%
\pgfsetdash{}{0pt}%
\pgfpathmoveto{\pgfqpoint{0.554481in}{2.052917in}}%
\pgfpathlineto{\pgfqpoint{2.530555in}{2.052917in}}%
\pgfusepath{stroke}%
\end{pgfscope}%
\begin{pgfscope}%
\definecolor{textcolor}{rgb}{1.000000,0.000000,0.000000}%
\pgfsetstrokecolor{textcolor}%
\pgfsetfillcolor{textcolor}%
\pgftext[x=2.332948in,y=1.959817in,right,top]{\color{textcolor}\rmfamily\fontsize{8.330000}{9.996000}\selectfont \contour{white}{\ourmethod{}\(\displaystyle ^\star\)}}%
\end{pgfscope}%
\begin{pgfscope}%
\definecolor{textcolor}{rgb}{0.000000,0.000000,0.000000}%
\pgfsetstrokecolor{textcolor}%
\pgfsetfillcolor{textcolor}%
\pgftext[x=1.542518in,y=2.136251in,,base]{\color{textcolor}\rmfamily\fontsize{12.000000}{14.400000}\selectfont \taskname{all} RSA}%
\end{pgfscope}%
\begin{pgfscope}%
\pgfsetbuttcap%
\pgfsetmiterjoin%
\definecolor{currentfill}{rgb}{1.000000,1.000000,1.000000}%
\pgfsetfillcolor{currentfill}%
\pgfsetfillopacity{0.800000}%
\pgfsetlinewidth{1.003750pt}%
\definecolor{currentstroke}{rgb}{0.800000,0.800000,0.800000}%
\pgfsetstrokecolor{currentstroke}%
\pgfsetstrokeopacity{0.800000}%
\pgfsetdash{}{0pt}%
\pgfpathmoveto{\pgfqpoint{1.001549in}{1.013907in}}%
\pgfpathlineto{\pgfqpoint{2.449569in}{1.013907in}}%
\pgfpathquadraticcurveto{\pgfqpoint{2.472708in}{1.013907in}}{\pgfqpoint{2.472708in}{1.037045in}}%
\pgfpathlineto{\pgfqpoint{2.472708in}{1.517117in}}%
\pgfpathquadraticcurveto{\pgfqpoint{2.472708in}{1.540256in}}{\pgfqpoint{2.449569in}{1.540256in}}%
\pgfpathlineto{\pgfqpoint{1.001549in}{1.540256in}}%
\pgfpathquadraticcurveto{\pgfqpoint{0.978410in}{1.540256in}}{\pgfqpoint{0.978410in}{1.517117in}}%
\pgfpathlineto{\pgfqpoint{0.978410in}{1.037045in}}%
\pgfpathquadraticcurveto{\pgfqpoint{0.978410in}{1.013907in}}{\pgfqpoint{1.001549in}{1.013907in}}%
\pgfpathlineto{\pgfqpoint{1.001549in}{1.013907in}}%
\pgfpathclose%
\pgfusepath{stroke,fill}%
\end{pgfscope}%
\begin{pgfscope}%
\pgfsetbuttcap%
\pgfsetroundjoin%
\pgfsetlinewidth{0.301125pt}%
\definecolor{currentstroke}{rgb}{0.215686,0.494118,0.721569}%
\pgfsetstrokecolor{currentstroke}%
\pgfsetdash{{1.110000pt}{0.480000pt}}{0.000000pt}%
\pgfpathmoveto{\pgfqpoint{1.024688in}{1.453485in}}%
\pgfpathlineto{\pgfqpoint{1.094104in}{1.453485in}}%
\pgfpathlineto{\pgfqpoint{1.163521in}{1.453485in}}%
\pgfusepath{stroke}%
\end{pgfscope}%
\begin{pgfscope}%
\definecolor{textcolor}{rgb}{0.000000,0.000000,0.000000}%
\pgfsetstrokecolor{textcolor}%
\pgfsetfillcolor{textcolor}%
\pgftext[x=1.256076in,y=1.412992in,left,base]{\color{textcolor}\rmfamily\fontsize{8.330000}{9.996000}\selectfont Liter. list.}%
\end{pgfscope}%
\begin{pgfscope}%
\pgfsetrectcap%
\pgfsetroundjoin%
\pgfsetlinewidth{0.602250pt}%
\definecolor{currentstroke}{rgb}{1.000000,0.498039,0.000000}%
\pgfsetstrokecolor{currentstroke}%
\pgfsetdash{}{0pt}%
\pgfpathmoveto{\pgfqpoint{1.024688in}{1.289605in}}%
\pgfpathlineto{\pgfqpoint{1.094104in}{1.289605in}}%
\pgfpathlineto{\pgfqpoint{1.163521in}{1.289605in}}%
\pgfusepath{stroke}%
\end{pgfscope}%
\begin{pgfscope}%
\definecolor{textcolor}{rgb}{0.000000,0.000000,0.000000}%
\pgfsetstrokecolor{textcolor}%
\pgfsetfillcolor{textcolor}%
\pgftext[x=1.256076in,y=1.249112in,left,base]{\color{textcolor}\rmfamily\fontsize{8.330000}{9.996000}\selectfont Depth 1}%
\end{pgfscope}%
\begin{pgfscope}%
\pgfsetbuttcap%
\pgfsetroundjoin%
\pgfsetlinewidth{0.903375pt}%
\definecolor{currentstroke}{rgb}{0.301961,0.686275,0.290196}%
\pgfsetstrokecolor{currentstroke}%
\pgfsetdash{{0.900000pt}{1.485000pt}}{0.000000pt}%
\pgfpathmoveto{\pgfqpoint{1.024688in}{1.125724in}}%
\pgfpathlineto{\pgfqpoint{1.094104in}{1.125724in}}%
\pgfpathlineto{\pgfqpoint{1.163521in}{1.125724in}}%
\pgfusepath{stroke}%
\end{pgfscope}%
\begin{pgfscope}%
\definecolor{textcolor}{rgb}{0.000000,0.000000,0.000000}%
\pgfsetstrokecolor{textcolor}%
\pgfsetfillcolor{textcolor}%
\pgftext[x=1.256076in,y=1.085231in,left,base]{\color{textcolor}\rmfamily\fontsize{8.330000}{9.996000}\selectfont Depth 2}%
\end{pgfscope}%
\begin{pgfscope}%
\pgfsetbuttcap%
\pgfsetroundjoin%
\pgfsetlinewidth{1.204500pt}%
\definecolor{currentstroke}{rgb}{0.968627,0.505882,0.749020}%
\pgfsetstrokecolor{currentstroke}%
\pgfsetdash{{7.680000pt}{1.920000pt}{1.200000pt}{1.920000pt}}{0.000000pt}%
\pgfpathmoveto{\pgfqpoint{1.825516in}{1.453485in}}%
\pgfpathlineto{\pgfqpoint{1.894933in}{1.453485in}}%
\pgfpathlineto{\pgfqpoint{1.964349in}{1.453485in}}%
\pgfusepath{stroke}%
\end{pgfscope}%
\begin{pgfscope}%
\definecolor{textcolor}{rgb}{0.000000,0.000000,0.000000}%
\pgfsetstrokecolor{textcolor}%
\pgfsetfillcolor{textcolor}%
\pgftext[x=2.056905in,y=1.412992in,left,base]{\color{textcolor}\rmfamily\fontsize{8.330000}{9.996000}\selectfont Depth 3}%
\end{pgfscope}%
\begin{pgfscope}%
\pgfsetbuttcap%
\pgfsetroundjoin%
\pgfsetlinewidth{1.505625pt}%
\definecolor{currentstroke}{rgb}{0.650980,0.337255,0.156863}%
\pgfsetstrokecolor{currentstroke}%
\pgfsetdash{{5.550000pt}{2.400000pt}}{0.000000pt}%
\pgfpathmoveto{\pgfqpoint{1.825516in}{1.289605in}}%
\pgfpathlineto{\pgfqpoint{1.894933in}{1.289605in}}%
\pgfpathlineto{\pgfqpoint{1.964349in}{1.289605in}}%
\pgfusepath{stroke}%
\end{pgfscope}%
\begin{pgfscope}%
\definecolor{textcolor}{rgb}{0.000000,0.000000,0.000000}%
\pgfsetstrokecolor{textcolor}%
\pgfsetfillcolor{textcolor}%
\pgftext[x=2.056905in,y=1.249112in,left,base]{\color{textcolor}\rmfamily\fontsize{8.330000}{9.996000}\selectfont Depth 4}%
\end{pgfscope}%
\end{pgfpicture}%
\makeatother%
\endgroup%

%% file: plots/rsa_all_startlistener_rd1.pgf
\begingroup%
\makeatletter%
\begin{pgfpicture}%
\pgfpathrectangle{\pgfpointorigin}{\pgfqpoint{2.630555in}{2.378626in}}%
\pgfusepath{use as bounding box, clip}%
\begin{pgfscope}%
\pgfsetbuttcap%
\pgfsetmiterjoin%
\definecolor{currentfill}{rgb}{1.000000,1.000000,1.000000}%
\pgfsetfillcolor{currentfill}%
\pgfsetlinewidth{0.000000pt}%
\definecolor{currentstroke}{rgb}{1.000000,1.000000,1.000000}%
\pgfsetstrokecolor{currentstroke}%
\pgfsetdash{}{0pt}%
\pgfpathmoveto{\pgfqpoint{0.000000in}{0.000000in}}%
\pgfpathlineto{\pgfqpoint{2.630555in}{0.000000in}}%
\pgfpathlineto{\pgfqpoint{2.630555in}{2.378626in}}%
\pgfpathlineto{\pgfqpoint{0.000000in}{2.378626in}}%
\pgfpathlineto{\pgfqpoint{0.000000in}{0.000000in}}%
\pgfpathclose%
\pgfusepath{fill}%
\end{pgfscope}%
\begin{pgfscope}%
\pgfsetbuttcap%
\pgfsetmiterjoin%
\definecolor{currentfill}{rgb}{1.000000,1.000000,1.000000}%
\pgfsetfillcolor{currentfill}%
\pgfsetlinewidth{0.000000pt}%
\definecolor{currentstroke}{rgb}{0.000000,0.000000,0.000000}%
\pgfsetstrokecolor{currentstroke}%
\pgfsetstrokeopacity{0.000000}%
\pgfsetdash{}{0pt}%
\pgfpathmoveto{\pgfqpoint{0.554481in}{0.501245in}}%
\pgfpathlineto{\pgfqpoint{2.530555in}{0.501245in}}%
\pgfpathlineto{\pgfqpoint{2.530555in}{2.052917in}}%
\pgfpathlineto{\pgfqpoint{0.554481in}{2.052917in}}%
\pgfpathlineto{\pgfqpoint{0.554481in}{0.501245in}}%
\pgfpathclose%
\pgfusepath{fill}%
\end{pgfscope}%
\begin{pgfscope}%
\pgfpathrectangle{\pgfqpoint{0.554481in}{0.501245in}}{\pgfqpoint{1.976074in}{1.551672in}}%
\pgfusepath{clip}%
\pgfsetrectcap%
\pgfsetroundjoin%
\pgfsetlinewidth{0.803000pt}%
\definecolor{currentstroke}{rgb}{0.690196,0.690196,0.690196}%
\pgfsetstrokecolor{currentstroke}%
\pgfsetdash{}{0pt}%
\pgfpathmoveto{\pgfqpoint{0.644303in}{0.501245in}}%
\pgfpathlineto{\pgfqpoint{0.644303in}{2.052917in}}%
\pgfusepath{stroke}%
\end{pgfscope}%
\begin{pgfscope}%
\pgfsetbuttcap%
\pgfsetroundjoin%
\definecolor{currentfill}{rgb}{0.000000,0.000000,0.000000}%
\pgfsetfillcolor{currentfill}%
\pgfsetlinewidth{0.803000pt}%
\definecolor{currentstroke}{rgb}{0.000000,0.000000,0.000000}%
\pgfsetstrokecolor{currentstroke}%
\pgfsetdash{}{0pt}%
\pgfsys@defobject{currentmarker}{\pgfqpoint{0.000000in}{-0.048611in}}{\pgfqpoint{0.000000in}{0.000000in}}{%
\pgfpathmoveto{\pgfqpoint{0.000000in}{0.000000in}}%
\pgfpathlineto{\pgfqpoint{0.000000in}{-0.048611in}}%
\pgfusepath{stroke,fill}%
}%
\begin{pgfscope}%
\pgfsys@transformshift{0.644303in}{0.501245in}%
\pgfsys@useobject{currentmarker}{}%
\end{pgfscope}%
\end{pgfscope}%
\begin{pgfscope}%
\definecolor{textcolor}{rgb}{0.000000,0.000000,0.000000}%
\pgfsetstrokecolor{textcolor}%
\pgfsetfillcolor{textcolor}%
\pgftext[x=0.644303in,y=0.404023in,,top]{\color{textcolor}\rmfamily\fontsize{10.000000}{12.000000}\selectfont \(\displaystyle {0}\)}%
\end{pgfscope}%
\begin{pgfscope}%
\pgfpathrectangle{\pgfqpoint{0.554481in}{0.501245in}}{\pgfqpoint{1.976074in}{1.551672in}}%
\pgfusepath{clip}%
\pgfsetrectcap%
\pgfsetroundjoin%
\pgfsetlinewidth{0.803000pt}%
\definecolor{currentstroke}{rgb}{0.690196,0.690196,0.690196}%
\pgfsetstrokecolor{currentstroke}%
\pgfsetdash{}{0pt}%
\pgfpathmoveto{\pgfqpoint{1.003589in}{0.501245in}}%
\pgfpathlineto{\pgfqpoint{1.003589in}{2.052917in}}%
\pgfusepath{stroke}%
\end{pgfscope}%
\begin{pgfscope}%
\pgfsetbuttcap%
\pgfsetroundjoin%
\definecolor{currentfill}{rgb}{0.000000,0.000000,0.000000}%
\pgfsetfillcolor{currentfill}%
\pgfsetlinewidth{0.803000pt}%
\definecolor{currentstroke}{rgb}{0.000000,0.000000,0.000000}%
\pgfsetstrokecolor{currentstroke}%
\pgfsetdash{}{0pt}%
\pgfsys@defobject{currentmarker}{\pgfqpoint{0.000000in}{-0.048611in}}{\pgfqpoint{0.000000in}{0.000000in}}{%
\pgfpathmoveto{\pgfqpoint{0.000000in}{0.000000in}}%
\pgfpathlineto{\pgfqpoint{0.000000in}{-0.048611in}}%
\pgfusepath{stroke,fill}%
}%
\begin{pgfscope}%
\pgfsys@transformshift{1.003589in}{0.501245in}%
\pgfsys@useobject{currentmarker}{}%
\end{pgfscope}%
\end{pgfscope}%
\begin{pgfscope}%
\definecolor{textcolor}{rgb}{0.000000,0.000000,0.000000}%
\pgfsetstrokecolor{textcolor}%
\pgfsetfillcolor{textcolor}%
\pgftext[x=1.003589in,y=0.404023in,,top]{\color{textcolor}\rmfamily\fontsize{10.000000}{12.000000}\selectfont \(\displaystyle {1}\)}%
\end{pgfscope}%
\begin{pgfscope}%
\pgfpathrectangle{\pgfqpoint{0.554481in}{0.501245in}}{\pgfqpoint{1.976074in}{1.551672in}}%
\pgfusepath{clip}%
\pgfsetrectcap%
\pgfsetroundjoin%
\pgfsetlinewidth{0.803000pt}%
\definecolor{currentstroke}{rgb}{0.690196,0.690196,0.690196}%
\pgfsetstrokecolor{currentstroke}%
\pgfsetdash{}{0pt}%
\pgfpathmoveto{\pgfqpoint{1.362875in}{0.501245in}}%
\pgfpathlineto{\pgfqpoint{1.362875in}{2.052917in}}%
\pgfusepath{stroke}%
\end{pgfscope}%
\begin{pgfscope}%
\pgfsetbuttcap%
\pgfsetroundjoin%
\definecolor{currentfill}{rgb}{0.000000,0.000000,0.000000}%
\pgfsetfillcolor{currentfill}%
\pgfsetlinewidth{0.803000pt}%
\definecolor{currentstroke}{rgb}{0.000000,0.000000,0.000000}%
\pgfsetstrokecolor{currentstroke}%
\pgfsetdash{}{0pt}%
\pgfsys@defobject{currentmarker}{\pgfqpoint{0.000000in}{-0.048611in}}{\pgfqpoint{0.000000in}{0.000000in}}{%
\pgfpathmoveto{\pgfqpoint{0.000000in}{0.000000in}}%
\pgfpathlineto{\pgfqpoint{0.000000in}{-0.048611in}}%
\pgfusepath{stroke,fill}%
}%
\begin{pgfscope}%
\pgfsys@transformshift{1.362875in}{0.501245in}%
\pgfsys@useobject{currentmarker}{}%
\end{pgfscope}%
\end{pgfscope}%
\begin{pgfscope}%
\definecolor{textcolor}{rgb}{0.000000,0.000000,0.000000}%
\pgfsetstrokecolor{textcolor}%
\pgfsetfillcolor{textcolor}%
\pgftext[x=1.362875in,y=0.404023in,,top]{\color{textcolor}\rmfamily\fontsize{10.000000}{12.000000}\selectfont \(\displaystyle {2}\)}%
\end{pgfscope}%
\begin{pgfscope}%
\pgfpathrectangle{\pgfqpoint{0.554481in}{0.501245in}}{\pgfqpoint{1.976074in}{1.551672in}}%
\pgfusepath{clip}%
\pgfsetrectcap%
\pgfsetroundjoin%
\pgfsetlinewidth{0.803000pt}%
\definecolor{currentstroke}{rgb}{0.690196,0.690196,0.690196}%
\pgfsetstrokecolor{currentstroke}%
\pgfsetdash{}{0pt}%
\pgfpathmoveto{\pgfqpoint{1.722161in}{0.501245in}}%
\pgfpathlineto{\pgfqpoint{1.722161in}{2.052917in}}%
\pgfusepath{stroke}%
\end{pgfscope}%
\begin{pgfscope}%
\pgfsetbuttcap%
\pgfsetroundjoin%
\definecolor{currentfill}{rgb}{0.000000,0.000000,0.000000}%
\pgfsetfillcolor{currentfill}%
\pgfsetlinewidth{0.803000pt}%
\definecolor{currentstroke}{rgb}{0.000000,0.000000,0.000000}%
\pgfsetstrokecolor{currentstroke}%
\pgfsetdash{}{0pt}%
\pgfsys@defobject{currentmarker}{\pgfqpoint{0.000000in}{-0.048611in}}{\pgfqpoint{0.000000in}{0.000000in}}{%
\pgfpathmoveto{\pgfqpoint{0.000000in}{0.000000in}}%
\pgfpathlineto{\pgfqpoint{0.000000in}{-0.048611in}}%
\pgfusepath{stroke,fill}%
}%
\begin{pgfscope}%
\pgfsys@transformshift{1.722161in}{0.501245in}%
\pgfsys@useobject{currentmarker}{}%
\end{pgfscope}%
\end{pgfscope}%
\begin{pgfscope}%
\definecolor{textcolor}{rgb}{0.000000,0.000000,0.000000}%
\pgfsetstrokecolor{textcolor}%
\pgfsetfillcolor{textcolor}%
\pgftext[x=1.722161in,y=0.404023in,,top]{\color{textcolor}\rmfamily\fontsize{10.000000}{12.000000}\selectfont \(\displaystyle {3}\)}%
\end{pgfscope}%
\begin{pgfscope}%
\pgfpathrectangle{\pgfqpoint{0.554481in}{0.501245in}}{\pgfqpoint{1.976074in}{1.551672in}}%
\pgfusepath{clip}%
\pgfsetrectcap%
\pgfsetroundjoin%
\pgfsetlinewidth{0.803000pt}%
\definecolor{currentstroke}{rgb}{0.690196,0.690196,0.690196}%
\pgfsetstrokecolor{currentstroke}%
\pgfsetdash{}{0pt}%
\pgfpathmoveto{\pgfqpoint{2.081448in}{0.501245in}}%
\pgfpathlineto{\pgfqpoint{2.081448in}{2.052917in}}%
\pgfusepath{stroke}%
\end{pgfscope}%
\begin{pgfscope}%
\pgfsetbuttcap%
\pgfsetroundjoin%
\definecolor{currentfill}{rgb}{0.000000,0.000000,0.000000}%
\pgfsetfillcolor{currentfill}%
\pgfsetlinewidth{0.803000pt}%
\definecolor{currentstroke}{rgb}{0.000000,0.000000,0.000000}%
\pgfsetstrokecolor{currentstroke}%
\pgfsetdash{}{0pt}%
\pgfsys@defobject{currentmarker}{\pgfqpoint{0.000000in}{-0.048611in}}{\pgfqpoint{0.000000in}{0.000000in}}{%
\pgfpathmoveto{\pgfqpoint{0.000000in}{0.000000in}}%
\pgfpathlineto{\pgfqpoint{0.000000in}{-0.048611in}}%
\pgfusepath{stroke,fill}%
}%
\begin{pgfscope}%
\pgfsys@transformshift{2.081448in}{0.501245in}%
\pgfsys@useobject{currentmarker}{}%
\end{pgfscope}%
\end{pgfscope}%
\begin{pgfscope}%
\definecolor{textcolor}{rgb}{0.000000,0.000000,0.000000}%
\pgfsetstrokecolor{textcolor}%
\pgfsetfillcolor{textcolor}%
\pgftext[x=2.081448in,y=0.404023in,,top]{\color{textcolor}\rmfamily\fontsize{10.000000}{12.000000}\selectfont \(\displaystyle {4}\)}%
\end{pgfscope}%
\begin{pgfscope}%
\pgfpathrectangle{\pgfqpoint{0.554481in}{0.501245in}}{\pgfqpoint{1.976074in}{1.551672in}}%
\pgfusepath{clip}%
\pgfsetrectcap%
\pgfsetroundjoin%
\pgfsetlinewidth{0.803000pt}%
\definecolor{currentstroke}{rgb}{0.690196,0.690196,0.690196}%
\pgfsetstrokecolor{currentstroke}%
\pgfsetdash{}{0pt}%
\pgfpathmoveto{\pgfqpoint{2.440734in}{0.501245in}}%
\pgfpathlineto{\pgfqpoint{2.440734in}{2.052917in}}%
\pgfusepath{stroke}%
\end{pgfscope}%
\begin{pgfscope}%
\pgfsetbuttcap%
\pgfsetroundjoin%
\definecolor{currentfill}{rgb}{0.000000,0.000000,0.000000}%
\pgfsetfillcolor{currentfill}%
\pgfsetlinewidth{0.803000pt}%
\definecolor{currentstroke}{rgb}{0.000000,0.000000,0.000000}%
\pgfsetstrokecolor{currentstroke}%
\pgfsetdash{}{0pt}%
\pgfsys@defobject{currentmarker}{\pgfqpoint{0.000000in}{-0.048611in}}{\pgfqpoint{0.000000in}{0.000000in}}{%
\pgfpathmoveto{\pgfqpoint{0.000000in}{0.000000in}}%
\pgfpathlineto{\pgfqpoint{0.000000in}{-0.048611in}}%
\pgfusepath{stroke,fill}%
}%
\begin{pgfscope}%
\pgfsys@transformshift{2.440734in}{0.501245in}%
\pgfsys@useobject{currentmarker}{}%
\end{pgfscope}%
\end{pgfscope}%
\begin{pgfscope}%
\definecolor{textcolor}{rgb}{0.000000,0.000000,0.000000}%
\pgfsetstrokecolor{textcolor}%
\pgfsetfillcolor{textcolor}%
\pgftext[x=2.440734in,y=0.404023in,,top]{\color{textcolor}\rmfamily\fontsize{10.000000}{12.000000}\selectfont \(\displaystyle {5}\)}%
\end{pgfscope}%
\begin{pgfscope}%
\definecolor{textcolor}{rgb}{0.000000,0.000000,0.000000}%
\pgfsetstrokecolor{textcolor}%
\pgfsetfillcolor{textcolor}%
\pgftext[x=1.542518in,y=0.224234in,,top]{\color{textcolor}\rmfamily\fontsize{10.000000}{12.000000}\selectfont Parameter \(\displaystyle \alpha\)}%
\end{pgfscope}%
\begin{pgfscope}%
\pgfpathrectangle{\pgfqpoint{0.554481in}{0.501245in}}{\pgfqpoint{1.976074in}{1.551672in}}%
\pgfusepath{clip}%
\pgfsetrectcap%
\pgfsetroundjoin%
\pgfsetlinewidth{0.803000pt}%
\definecolor{currentstroke}{rgb}{0.690196,0.690196,0.690196}%
\pgfsetstrokecolor{currentstroke}%
\pgfsetdash{}{0pt}%
\pgfpathmoveto{\pgfqpoint{0.554481in}{0.926307in}}%
\pgfpathlineto{\pgfqpoint{2.530555in}{0.926307in}}%
\pgfusepath{stroke}%
\end{pgfscope}%
\begin{pgfscope}%
\pgfsetbuttcap%
\pgfsetroundjoin%
\definecolor{currentfill}{rgb}{0.000000,0.000000,0.000000}%
\pgfsetfillcolor{currentfill}%
\pgfsetlinewidth{0.803000pt}%
\definecolor{currentstroke}{rgb}{0.000000,0.000000,0.000000}%
\pgfsetstrokecolor{currentstroke}%
\pgfsetdash{}{0pt}%
\pgfsys@defobject{currentmarker}{\pgfqpoint{-0.048611in}{0.000000in}}{\pgfqpoint{-0.000000in}{0.000000in}}{%
\pgfpathmoveto{\pgfqpoint{-0.000000in}{0.000000in}}%
\pgfpathlineto{\pgfqpoint{-0.048611in}{0.000000in}}%
\pgfusepath{stroke,fill}%
}%
\begin{pgfscope}%
\pgfsys@transformshift{0.554481in}{0.926307in}%
\pgfsys@useobject{currentmarker}{}%
\end{pgfscope}%
\end{pgfscope}%
\begin{pgfscope}%
\definecolor{textcolor}{rgb}{0.000000,0.000000,0.000000}%
\pgfsetstrokecolor{textcolor}%
\pgfsetfillcolor{textcolor}%
\pgftext[x=0.279789in, y=0.879225in, left, base]{\color{textcolor}\rmfamily\fontsize{10.000000}{12.000000}\selectfont \(\displaystyle {0.8}\)}%
\end{pgfscope}%
\begin{pgfscope}%
\pgfpathrectangle{\pgfqpoint{0.554481in}{0.501245in}}{\pgfqpoint{1.976074in}{1.551672in}}%
\pgfusepath{clip}%
\pgfsetrectcap%
\pgfsetroundjoin%
\pgfsetlinewidth{0.803000pt}%
\definecolor{currentstroke}{rgb}{0.690196,0.690196,0.690196}%
\pgfsetstrokecolor{currentstroke}%
\pgfsetdash{}{0pt}%
\pgfpathmoveto{\pgfqpoint{0.554481in}{1.489612in}}%
\pgfpathlineto{\pgfqpoint{2.530555in}{1.489612in}}%
\pgfusepath{stroke}%
\end{pgfscope}%
\begin{pgfscope}%
\pgfsetbuttcap%
\pgfsetroundjoin%
\definecolor{currentfill}{rgb}{0.000000,0.000000,0.000000}%
\pgfsetfillcolor{currentfill}%
\pgfsetlinewidth{0.803000pt}%
\definecolor{currentstroke}{rgb}{0.000000,0.000000,0.000000}%
\pgfsetstrokecolor{currentstroke}%
\pgfsetdash{}{0pt}%
\pgfsys@defobject{currentmarker}{\pgfqpoint{-0.048611in}{0.000000in}}{\pgfqpoint{-0.000000in}{0.000000in}}{%
\pgfpathmoveto{\pgfqpoint{-0.000000in}{0.000000in}}%
\pgfpathlineto{\pgfqpoint{-0.048611in}{0.000000in}}%
\pgfusepath{stroke,fill}%
}%
\begin{pgfscope}%
\pgfsys@transformshift{0.554481in}{1.489612in}%
\pgfsys@useobject{currentmarker}{}%
\end{pgfscope}%
\end{pgfscope}%
\begin{pgfscope}%
\definecolor{textcolor}{rgb}{0.000000,0.000000,0.000000}%
\pgfsetstrokecolor{textcolor}%
\pgfsetfillcolor{textcolor}%
\pgftext[x=0.279789in, y=1.442530in, left, base]{\color{textcolor}\rmfamily\fontsize{10.000000}{12.000000}\selectfont \(\displaystyle {0.9}\)}%
\end{pgfscope}%
\begin{pgfscope}%
\pgfpathrectangle{\pgfqpoint{0.554481in}{0.501245in}}{\pgfqpoint{1.976074in}{1.551672in}}%
\pgfusepath{clip}%
\pgfsetrectcap%
\pgfsetroundjoin%
\pgfsetlinewidth{0.803000pt}%
\definecolor{currentstroke}{rgb}{0.690196,0.690196,0.690196}%
\pgfsetstrokecolor{currentstroke}%
\pgfsetdash{}{0pt}%
\pgfpathmoveto{\pgfqpoint{0.554481in}{2.052917in}}%
\pgfpathlineto{\pgfqpoint{2.530555in}{2.052917in}}%
\pgfusepath{stroke}%
\end{pgfscope}%
\begin{pgfscope}%
\pgfsetbuttcap%
\pgfsetroundjoin%
\definecolor{currentfill}{rgb}{0.000000,0.000000,0.000000}%
\pgfsetfillcolor{currentfill}%
\pgfsetlinewidth{0.803000pt}%
\definecolor{currentstroke}{rgb}{0.000000,0.000000,0.000000}%
\pgfsetstrokecolor{currentstroke}%
\pgfsetdash{}{0pt}%
\pgfsys@defobject{currentmarker}{\pgfqpoint{-0.048611in}{0.000000in}}{\pgfqpoint{-0.000000in}{0.000000in}}{%
\pgfpathmoveto{\pgfqpoint{-0.000000in}{0.000000in}}%
\pgfpathlineto{\pgfqpoint{-0.048611in}{0.000000in}}%
\pgfusepath{stroke,fill}%
}%
\begin{pgfscope}%
\pgfsys@transformshift{0.554481in}{2.052917in}%
\pgfsys@useobject{currentmarker}{}%
\end{pgfscope}%
\end{pgfscope}%
\begin{pgfscope}%
\definecolor{textcolor}{rgb}{0.000000,0.000000,0.000000}%
\pgfsetstrokecolor{textcolor}%
\pgfsetfillcolor{textcolor}%
\pgftext[x=0.279789in, y=2.005835in, left, base]{\color{textcolor}\rmfamily\fontsize{10.000000}{12.000000}\selectfont \(\displaystyle {1.0}\)}%
\end{pgfscope}%
\begin{pgfscope}%
\definecolor{textcolor}{rgb}{0.000000,0.000000,0.000000}%
\pgfsetstrokecolor{textcolor}%
\pgfsetfillcolor{textcolor}%
\pgftext[x=0.224234in,y=1.277081in,,bottom,rotate=90.000000]{\color{textcolor}\rmfamily\fontsize{10.000000}{12.000000}\selectfont Pearson correlation}%
\end{pgfscope}%
\begin{pgfscope}%
\pgfpathrectangle{\pgfqpoint{0.554481in}{0.501245in}}{\pgfqpoint{1.976074in}{1.551672in}}%
\pgfusepath{clip}%
\pgfsetbuttcap%
\pgfsetroundjoin%
\pgfsetlinewidth{0.301125pt}%
\definecolor{currentstroke}{rgb}{0.215686,0.494118,0.721569}%
\pgfsetstrokecolor{currentstroke}%
\pgfsetdash{{1.110000pt}{0.480000pt}}{0.000000pt}%
\pgfpathmoveto{\pgfqpoint{0.644303in}{0.564310in}}%
\pgfpathlineto{\pgfqpoint{2.440734in}{0.564310in}}%
\pgfpathlineto{\pgfqpoint{2.440734in}{0.564310in}}%
\pgfusepath{stroke}%
\end{pgfscope}%
\begin{pgfscope}%
\pgfpathrectangle{\pgfqpoint{0.554481in}{0.501245in}}{\pgfqpoint{1.976074in}{1.551672in}}%
\pgfusepath{clip}%
\pgfsetbuttcap%
\pgfsetbeveljoin%
\definecolor{currentfill}{rgb}{0.000000,0.000000,0.000000}%
\pgfsetfillcolor{currentfill}%
\pgfsetfillopacity{0.000000}%
\pgfsetlinewidth{0.501875pt}%
\definecolor{currentstroke}{rgb}{0.215686,0.494118,0.721569}%
\pgfsetstrokecolor{currentstroke}%
\pgfsetdash{}{0pt}%
\pgfsys@defobject{currentmarker}{\pgfqpoint{-0.033023in}{-0.028091in}}{\pgfqpoint{0.033023in}{0.034722in}}{%
\pgfpathmoveto{\pgfqpoint{0.000000in}{0.034722in}}%
\pgfpathlineto{\pgfqpoint{-0.007796in}{0.010730in}}%
\pgfpathlineto{\pgfqpoint{-0.033023in}{0.010730in}}%
\pgfpathlineto{\pgfqpoint{-0.012614in}{-0.004098in}}%
\pgfpathlineto{\pgfqpoint{-0.020409in}{-0.028091in}}%
\pgfpathlineto{\pgfqpoint{-0.000000in}{-0.013263in}}%
\pgfpathlineto{\pgfqpoint{0.020409in}{-0.028091in}}%
\pgfpathlineto{\pgfqpoint{0.012614in}{-0.004098in}}%
\pgfpathlineto{\pgfqpoint{0.033023in}{0.010730in}}%
\pgfpathlineto{\pgfqpoint{0.007796in}{0.010730in}}%
\pgfpathlineto{\pgfqpoint{0.000000in}{0.034722in}}%
\pgfpathclose%
\pgfusepath{stroke,fill}%
}%
\begin{pgfscope}%
\pgfsys@transformshift{0.644303in}{0.564310in}%
\pgfsys@useobject{currentmarker}{}%
\end{pgfscope}%
\end{pgfscope}%
\begin{pgfscope}%
\pgfpathrectangle{\pgfqpoint{0.554481in}{0.501245in}}{\pgfqpoint{1.976074in}{1.551672in}}%
\pgfusepath{clip}%
\pgfsetrectcap%
\pgfsetroundjoin%
\pgfsetlinewidth{0.602250pt}%
\definecolor{currentstroke}{rgb}{1.000000,0.498039,0.000000}%
\pgfsetstrokecolor{currentstroke}%
\pgfsetdash{}{0pt}%
\pgfpathmoveto{\pgfqpoint{0.644303in}{0.564311in}}%
\pgfpathlineto{\pgfqpoint{0.671385in}{0.676825in}}%
\pgfpathlineto{\pgfqpoint{0.707494in}{0.829918in}}%
\pgfpathlineto{\pgfqpoint{0.752631in}{1.021082in}}%
\pgfpathlineto{\pgfqpoint{0.770685in}{1.095788in}}%
\pgfpathlineto{\pgfqpoint{0.788740in}{1.168630in}}%
\pgfpathlineto{\pgfqpoint{0.806794in}{1.239026in}}%
\pgfpathlineto{\pgfqpoint{0.824849in}{1.306406in}}%
\pgfpathlineto{\pgfqpoint{0.842903in}{1.370233in}}%
\pgfpathlineto{\pgfqpoint{0.851931in}{1.400660in}}%
\pgfpathlineto{\pgfqpoint{0.860958in}{1.430020in}}%
\pgfpathlineto{\pgfqpoint{0.869985in}{1.458264in}}%
\pgfpathlineto{\pgfqpoint{0.879013in}{1.485345in}}%
\pgfpathlineto{\pgfqpoint{0.888040in}{1.511224in}}%
\pgfpathlineto{\pgfqpoint{0.897067in}{1.535867in}}%
\pgfpathlineto{\pgfqpoint{0.906094in}{1.559245in}}%
\pgfpathlineto{\pgfqpoint{0.915122in}{1.581337in}}%
\pgfpathlineto{\pgfqpoint{0.924149in}{1.602127in}}%
\pgfpathlineto{\pgfqpoint{0.933176in}{1.621606in}}%
\pgfpathlineto{\pgfqpoint{0.942204in}{1.639772in}}%
\pgfpathlineto{\pgfqpoint{0.951231in}{1.656627in}}%
\pgfpathlineto{\pgfqpoint{0.960258in}{1.672181in}}%
\pgfpathlineto{\pgfqpoint{0.969285in}{1.686449in}}%
\pgfpathlineto{\pgfqpoint{0.978313in}{1.699453in}}%
\pgfpathlineto{\pgfqpoint{0.987340in}{1.711219in}}%
\pgfpathlineto{\pgfqpoint{0.996367in}{1.721776in}}%
\pgfpathlineto{\pgfqpoint{1.005395in}{1.731161in}}%
\pgfpathlineto{\pgfqpoint{1.014422in}{1.739412in}}%
\pgfpathlineto{\pgfqpoint{1.023449in}{1.746571in}}%
\pgfpathlineto{\pgfqpoint{1.032477in}{1.752684in}}%
\pgfpathlineto{\pgfqpoint{1.041504in}{1.757799in}}%
\pgfpathlineto{\pgfqpoint{1.050531in}{1.761963in}}%
\pgfpathlineto{\pgfqpoint{1.059558in}{1.765230in}}%
\pgfpathlineto{\pgfqpoint{1.068586in}{1.767650in}}%
\pgfpathlineto{\pgfqpoint{1.077613in}{1.769276in}}%
\pgfpathlineto{\pgfqpoint{1.086640in}{1.770160in}}%
\pgfpathlineto{\pgfqpoint{1.095668in}{1.770355in}}%
\pgfpathlineto{\pgfqpoint{1.104695in}{1.769913in}}%
\pgfpathlineto{\pgfqpoint{1.113722in}{1.768884in}}%
\pgfpathlineto{\pgfqpoint{1.122749in}{1.767318in}}%
\pgfpathlineto{\pgfqpoint{1.131777in}{1.765264in}}%
\pgfpathlineto{\pgfqpoint{1.140804in}{1.762768in}}%
\pgfpathlineto{\pgfqpoint{1.149831in}{1.759876in}}%
\pgfpathlineto{\pgfqpoint{1.158859in}{1.756630in}}%
\pgfpathlineto{\pgfqpoint{1.176913in}{1.749242in}}%
\pgfpathlineto{\pgfqpoint{1.194968in}{1.740909in}}%
\pgfpathlineto{\pgfqpoint{1.213022in}{1.731901in}}%
\pgfpathlineto{\pgfqpoint{1.240104in}{1.717635in}}%
\pgfpathlineto{\pgfqpoint{1.303295in}{1.683952in}}%
\pgfpathlineto{\pgfqpoint{1.330377in}{1.670379in}}%
\pgfpathlineto{\pgfqpoint{1.348432in}{1.661827in}}%
\pgfpathlineto{\pgfqpoint{1.366486in}{1.653733in}}%
\pgfpathlineto{\pgfqpoint{1.384541in}{1.646130in}}%
\pgfpathlineto{\pgfqpoint{1.402595in}{1.639040in}}%
\pgfpathlineto{\pgfqpoint{1.420650in}{1.632473in}}%
\pgfpathlineto{\pgfqpoint{1.438705in}{1.626433in}}%
\pgfpathlineto{\pgfqpoint{1.456759in}{1.620913in}}%
\pgfpathlineto{\pgfqpoint{1.474814in}{1.615900in}}%
\pgfpathlineto{\pgfqpoint{1.492868in}{1.611377in}}%
\pgfpathlineto{\pgfqpoint{1.510923in}{1.607324in}}%
\pgfpathlineto{\pgfqpoint{1.528978in}{1.603715in}}%
\pgfpathlineto{\pgfqpoint{1.547032in}{1.600524in}}%
\pgfpathlineto{\pgfqpoint{1.574114in}{1.596457in}}%
\pgfpathlineto{\pgfqpoint{1.601196in}{1.593168in}}%
\pgfpathlineto{\pgfqpoint{1.628278in}{1.590555in}}%
\pgfpathlineto{\pgfqpoint{1.655360in}{1.588524in}}%
\pgfpathlineto{\pgfqpoint{1.682441in}{1.586984in}}%
\pgfpathlineto{\pgfqpoint{1.718551in}{1.585554in}}%
\pgfpathlineto{\pgfqpoint{1.754660in}{1.584679in}}%
\pgfpathlineto{\pgfqpoint{1.799796in}{1.584149in}}%
\pgfpathlineto{\pgfqpoint{1.862987in}{1.584081in}}%
\pgfpathlineto{\pgfqpoint{1.953260in}{1.584648in}}%
\pgfpathlineto{\pgfqpoint{2.287270in}{1.587188in}}%
\pgfpathlineto{\pgfqpoint{2.440734in}{1.587862in}}%
\pgfpathlineto{\pgfqpoint{2.440734in}{1.587862in}}%
\pgfusepath{stroke}%
\end{pgfscope}%
\begin{pgfscope}%
\pgfpathrectangle{\pgfqpoint{0.554481in}{0.501245in}}{\pgfqpoint{1.976074in}{1.551672in}}%
\pgfusepath{clip}%
\pgfsetbuttcap%
\pgfsetbeveljoin%
\definecolor{currentfill}{rgb}{0.000000,0.000000,0.000000}%
\pgfsetfillcolor{currentfill}%
\pgfsetfillopacity{0.000000}%
\pgfsetlinewidth{0.501875pt}%
\definecolor{currentstroke}{rgb}{1.000000,0.498039,0.000000}%
\pgfsetstrokecolor{currentstroke}%
\pgfsetdash{}{0pt}%
\pgfsys@defobject{currentmarker}{\pgfqpoint{-0.033023in}{-0.028091in}}{\pgfqpoint{0.033023in}{0.034722in}}{%
\pgfpathmoveto{\pgfqpoint{0.000000in}{0.034722in}}%
\pgfpathlineto{\pgfqpoint{-0.007796in}{0.010730in}}%
\pgfpathlineto{\pgfqpoint{-0.033023in}{0.010730in}}%
\pgfpathlineto{\pgfqpoint{-0.012614in}{-0.004098in}}%
\pgfpathlineto{\pgfqpoint{-0.020409in}{-0.028091in}}%
\pgfpathlineto{\pgfqpoint{-0.000000in}{-0.013263in}}%
\pgfpathlineto{\pgfqpoint{0.020409in}{-0.028091in}}%
\pgfpathlineto{\pgfqpoint{0.012614in}{-0.004098in}}%
\pgfpathlineto{\pgfqpoint{0.033023in}{0.010730in}}%
\pgfpathlineto{\pgfqpoint{0.007796in}{0.010730in}}%
\pgfpathlineto{\pgfqpoint{0.000000in}{0.034722in}}%
\pgfpathclose%
\pgfusepath{stroke,fill}%
}%
\begin{pgfscope}%
\pgfsys@transformshift{1.095668in}{1.770355in}%
\pgfsys@useobject{currentmarker}{}%
\end{pgfscope}%
\end{pgfscope}%
\begin{pgfscope}%
\pgfpathrectangle{\pgfqpoint{0.554481in}{0.501245in}}{\pgfqpoint{1.976074in}{1.551672in}}%
\pgfusepath{clip}%
\pgfsetbuttcap%
\pgfsetroundjoin%
\pgfsetlinewidth{0.903375pt}%
\definecolor{currentstroke}{rgb}{0.301961,0.686275,0.290196}%
\pgfsetstrokecolor{currentstroke}%
\pgfsetdash{{0.900000pt}{1.485000pt}}{0.000000pt}%
\pgfpathmoveto{\pgfqpoint{0.644303in}{0.564311in}}%
\pgfpathlineto{\pgfqpoint{0.653330in}{0.580764in}}%
\pgfpathlineto{\pgfqpoint{0.662358in}{0.597819in}}%
\pgfpathlineto{\pgfqpoint{0.671385in}{0.615507in}}%
\pgfpathlineto{\pgfqpoint{0.680412in}{0.633858in}}%
\pgfpathlineto{\pgfqpoint{0.689439in}{0.652907in}}%
\pgfpathlineto{\pgfqpoint{0.698467in}{0.672687in}}%
\pgfpathlineto{\pgfqpoint{0.707494in}{0.693236in}}%
\pgfpathlineto{\pgfqpoint{0.716521in}{0.714587in}}%
\pgfpathlineto{\pgfqpoint{0.725549in}{0.736779in}}%
\pgfpathlineto{\pgfqpoint{0.734576in}{0.759847in}}%
\pgfpathlineto{\pgfqpoint{0.743603in}{0.783828in}}%
\pgfpathlineto{\pgfqpoint{0.752631in}{0.808758in}}%
\pgfpathlineto{\pgfqpoint{0.761658in}{0.834670in}}%
\pgfpathlineto{\pgfqpoint{0.770685in}{0.861597in}}%
\pgfpathlineto{\pgfqpoint{0.779712in}{0.889567in}}%
\pgfpathlineto{\pgfqpoint{0.788740in}{0.918604in}}%
\pgfpathlineto{\pgfqpoint{0.797767in}{0.948728in}}%
\pgfpathlineto{\pgfqpoint{0.806794in}{0.979950in}}%
\pgfpathlineto{\pgfqpoint{0.815822in}{1.012273in}}%
\pgfpathlineto{\pgfqpoint{0.824849in}{1.045689in}}%
\pgfpathlineto{\pgfqpoint{0.842903in}{1.115695in}}%
\pgfpathlineto{\pgfqpoint{0.860958in}{1.189588in}}%
\pgfpathlineto{\pgfqpoint{0.879013in}{1.266637in}}%
\pgfpathlineto{\pgfqpoint{0.933176in}{1.502253in}}%
\pgfpathlineto{\pgfqpoint{0.942204in}{1.539286in}}%
\pgfpathlineto{\pgfqpoint{0.951231in}{1.574735in}}%
\pgfpathlineto{\pgfqpoint{0.960258in}{1.608200in}}%
\pgfpathlineto{\pgfqpoint{0.969285in}{1.639282in}}%
\pgfpathlineto{\pgfqpoint{0.978313in}{1.667598in}}%
\pgfpathlineto{\pgfqpoint{0.987340in}{1.692796in}}%
\pgfpathlineto{\pgfqpoint{0.996367in}{1.714575in}}%
\pgfpathlineto{\pgfqpoint{1.005395in}{1.732701in}}%
\pgfpathlineto{\pgfqpoint{1.014422in}{1.747024in}}%
\pgfpathlineto{\pgfqpoint{1.023449in}{1.757486in}}%
\pgfpathlineto{\pgfqpoint{1.032477in}{1.764134in}}%
\pgfpathlineto{\pgfqpoint{1.041504in}{1.767115in}}%
\pgfpathlineto{\pgfqpoint{1.050531in}{1.766676in}}%
\pgfpathlineto{\pgfqpoint{1.059558in}{1.763146in}}%
\pgfpathlineto{\pgfqpoint{1.068586in}{1.756929in}}%
\pgfpathlineto{\pgfqpoint{1.077613in}{1.748473in}}%
\pgfpathlineto{\pgfqpoint{1.086640in}{1.738255in}}%
\pgfpathlineto{\pgfqpoint{1.095668in}{1.726758in}}%
\pgfpathlineto{\pgfqpoint{1.113722in}{1.701744in}}%
\pgfpathlineto{\pgfqpoint{1.131777in}{1.676656in}}%
\pgfpathlineto{\pgfqpoint{1.140804in}{1.664852in}}%
\pgfpathlineto{\pgfqpoint{1.149831in}{1.653829in}}%
\pgfpathlineto{\pgfqpoint{1.158859in}{1.643722in}}%
\pgfpathlineto{\pgfqpoint{1.167886in}{1.634616in}}%
\pgfpathlineto{\pgfqpoint{1.176913in}{1.626543in}}%
\pgfpathlineto{\pgfqpoint{1.185940in}{1.619499in}}%
\pgfpathlineto{\pgfqpoint{1.194968in}{1.613445in}}%
\pgfpathlineto{\pgfqpoint{1.203995in}{1.608319in}}%
\pgfpathlineto{\pgfqpoint{1.213022in}{1.604043in}}%
\pgfpathlineto{\pgfqpoint{1.222050in}{1.600530in}}%
\pgfpathlineto{\pgfqpoint{1.231077in}{1.597687in}}%
\pgfpathlineto{\pgfqpoint{1.240104in}{1.595422in}}%
\pgfpathlineto{\pgfqpoint{1.249131in}{1.593649in}}%
\pgfpathlineto{\pgfqpoint{1.258159in}{1.592284in}}%
\pgfpathlineto{\pgfqpoint{1.276213in}{1.590494in}}%
\pgfpathlineto{\pgfqpoint{1.294268in}{1.589567in}}%
\pgfpathlineto{\pgfqpoint{1.321350in}{1.589060in}}%
\pgfpathlineto{\pgfqpoint{1.366486in}{1.589073in}}%
\pgfpathlineto{\pgfqpoint{1.510923in}{1.589332in}}%
\pgfpathlineto{\pgfqpoint{2.440734in}{1.589339in}}%
\pgfpathlineto{\pgfqpoint{2.440734in}{1.589339in}}%
\pgfusepath{stroke}%
\end{pgfscope}%
\begin{pgfscope}%
\pgfpathrectangle{\pgfqpoint{0.554481in}{0.501245in}}{\pgfqpoint{1.976074in}{1.551672in}}%
\pgfusepath{clip}%
\pgfsetbuttcap%
\pgfsetbeveljoin%
\definecolor{currentfill}{rgb}{0.000000,0.000000,0.000000}%
\pgfsetfillcolor{currentfill}%
\pgfsetfillopacity{0.000000}%
\pgfsetlinewidth{0.501875pt}%
\definecolor{currentstroke}{rgb}{0.301961,0.686275,0.290196}%
\pgfsetstrokecolor{currentstroke}%
\pgfsetdash{}{0pt}%
\pgfsys@defobject{currentmarker}{\pgfqpoint{-0.033023in}{-0.028091in}}{\pgfqpoint{0.033023in}{0.034722in}}{%
\pgfpathmoveto{\pgfqpoint{0.000000in}{0.034722in}}%
\pgfpathlineto{\pgfqpoint{-0.007796in}{0.010730in}}%
\pgfpathlineto{\pgfqpoint{-0.033023in}{0.010730in}}%
\pgfpathlineto{\pgfqpoint{-0.012614in}{-0.004098in}}%
\pgfpathlineto{\pgfqpoint{-0.020409in}{-0.028091in}}%
\pgfpathlineto{\pgfqpoint{-0.000000in}{-0.013263in}}%
\pgfpathlineto{\pgfqpoint{0.020409in}{-0.028091in}}%
\pgfpathlineto{\pgfqpoint{0.012614in}{-0.004098in}}%
\pgfpathlineto{\pgfqpoint{0.033023in}{0.010730in}}%
\pgfpathlineto{\pgfqpoint{0.007796in}{0.010730in}}%
\pgfpathlineto{\pgfqpoint{0.000000in}{0.034722in}}%
\pgfpathclose%
\pgfusepath{stroke,fill}%
}%
\begin{pgfscope}%
\pgfsys@transformshift{1.041504in}{1.767115in}%
\pgfsys@useobject{currentmarker}{}%
\end{pgfscope}%
\end{pgfscope}%
\begin{pgfscope}%
\pgfpathrectangle{\pgfqpoint{0.554481in}{0.501245in}}{\pgfqpoint{1.976074in}{1.551672in}}%
\pgfusepath{clip}%
\pgfsetbuttcap%
\pgfsetroundjoin%
\pgfsetlinewidth{1.204500pt}%
\definecolor{currentstroke}{rgb}{0.968627,0.505882,0.749020}%
\pgfsetstrokecolor{currentstroke}%
\pgfsetdash{{7.680000pt}{1.920000pt}{1.200000pt}{1.920000pt}}{0.000000pt}%
\pgfpathmoveto{\pgfqpoint{0.644303in}{0.564310in}}%
\pgfpathlineto{\pgfqpoint{0.653330in}{0.573147in}}%
\pgfpathlineto{\pgfqpoint{0.662358in}{0.582434in}}%
\pgfpathlineto{\pgfqpoint{0.671385in}{0.592209in}}%
\pgfpathlineto{\pgfqpoint{0.680412in}{0.602514in}}%
\pgfpathlineto{\pgfqpoint{0.689439in}{0.613393in}}%
\pgfpathlineto{\pgfqpoint{0.698467in}{0.624894in}}%
\pgfpathlineto{\pgfqpoint{0.707494in}{0.637069in}}%
\pgfpathlineto{\pgfqpoint{0.716521in}{0.649973in}}%
\pgfpathlineto{\pgfqpoint{0.725549in}{0.663666in}}%
\pgfpathlineto{\pgfqpoint{0.734576in}{0.678212in}}%
\pgfpathlineto{\pgfqpoint{0.743603in}{0.693680in}}%
\pgfpathlineto{\pgfqpoint{0.752631in}{0.710143in}}%
\pgfpathlineto{\pgfqpoint{0.761658in}{0.727681in}}%
\pgfpathlineto{\pgfqpoint{0.770685in}{0.746377in}}%
\pgfpathlineto{\pgfqpoint{0.779712in}{0.766320in}}%
\pgfpathlineto{\pgfqpoint{0.788740in}{0.787605in}}%
\pgfpathlineto{\pgfqpoint{0.797767in}{0.810332in}}%
\pgfpathlineto{\pgfqpoint{0.806794in}{0.834602in}}%
\pgfpathlineto{\pgfqpoint{0.815822in}{0.860524in}}%
\pgfpathlineto{\pgfqpoint{0.824849in}{0.888205in}}%
\pgfpathlineto{\pgfqpoint{0.833876in}{0.917754in}}%
\pgfpathlineto{\pgfqpoint{0.842903in}{0.949276in}}%
\pgfpathlineto{\pgfqpoint{0.851931in}{0.982872in}}%
\pgfpathlineto{\pgfqpoint{0.860958in}{1.018628in}}%
\pgfpathlineto{\pgfqpoint{0.869985in}{1.056615in}}%
\pgfpathlineto{\pgfqpoint{0.879013in}{1.096876in}}%
\pgfpathlineto{\pgfqpoint{0.888040in}{1.139416in}}%
\pgfpathlineto{\pgfqpoint{0.897067in}{1.184193in}}%
\pgfpathlineto{\pgfqpoint{0.906094in}{1.231095in}}%
\pgfpathlineto{\pgfqpoint{0.915122in}{1.279927in}}%
\pgfpathlineto{\pgfqpoint{0.933176in}{1.382038in}}%
\pgfpathlineto{\pgfqpoint{0.960258in}{1.537452in}}%
\pgfpathlineto{\pgfqpoint{0.969285in}{1.586301in}}%
\pgfpathlineto{\pgfqpoint{0.978313in}{1.631712in}}%
\pgfpathlineto{\pgfqpoint{0.987340in}{1.672353in}}%
\pgfpathlineto{\pgfqpoint{0.996367in}{1.706910in}}%
\pgfpathlineto{\pgfqpoint{1.005395in}{1.734208in}}%
\pgfpathlineto{\pgfqpoint{1.014422in}{1.753372in}}%
\pgfpathlineto{\pgfqpoint{1.023449in}{1.763965in}}%
\pgfpathlineto{\pgfqpoint{1.032477in}{1.766104in}}%
\pgfpathlineto{\pgfqpoint{1.041504in}{1.760501in}}%
\pgfpathlineto{\pgfqpoint{1.050531in}{1.748416in}}%
\pgfpathlineto{\pgfqpoint{1.059558in}{1.731518in}}%
\pgfpathlineto{\pgfqpoint{1.068586in}{1.711680in}}%
\pgfpathlineto{\pgfqpoint{1.086640in}{1.670323in}}%
\pgfpathlineto{\pgfqpoint{1.095668in}{1.651640in}}%
\pgfpathlineto{\pgfqpoint{1.104695in}{1.635478in}}%
\pgfpathlineto{\pgfqpoint{1.113722in}{1.622192in}}%
\pgfpathlineto{\pgfqpoint{1.122749in}{1.611784in}}%
\pgfpathlineto{\pgfqpoint{1.131777in}{1.604000in}}%
\pgfpathlineto{\pgfqpoint{1.140804in}{1.598444in}}%
\pgfpathlineto{\pgfqpoint{1.149831in}{1.594665in}}%
\pgfpathlineto{\pgfqpoint{1.158859in}{1.592224in}}%
\pgfpathlineto{\pgfqpoint{1.167886in}{1.590736in}}%
\pgfpathlineto{\pgfqpoint{1.176913in}{1.589892in}}%
\pgfpathlineto{\pgfqpoint{1.194968in}{1.589262in}}%
\pgfpathlineto{\pgfqpoint{1.231077in}{1.589261in}}%
\pgfpathlineto{\pgfqpoint{1.348432in}{1.589339in}}%
\pgfpathlineto{\pgfqpoint{2.440734in}{1.589339in}}%
\pgfpathlineto{\pgfqpoint{2.440734in}{1.589339in}}%
\pgfusepath{stroke}%
\end{pgfscope}%
\begin{pgfscope}%
\pgfpathrectangle{\pgfqpoint{0.554481in}{0.501245in}}{\pgfqpoint{1.976074in}{1.551672in}}%
\pgfusepath{clip}%
\pgfsetbuttcap%
\pgfsetbeveljoin%
\definecolor{currentfill}{rgb}{0.000000,0.000000,0.000000}%
\pgfsetfillcolor{currentfill}%
\pgfsetfillopacity{0.000000}%
\pgfsetlinewidth{0.501875pt}%
\definecolor{currentstroke}{rgb}{0.968627,0.505882,0.749020}%
\pgfsetstrokecolor{currentstroke}%
\pgfsetdash{}{0pt}%
\pgfsys@defobject{currentmarker}{\pgfqpoint{-0.033023in}{-0.028091in}}{\pgfqpoint{0.033023in}{0.034722in}}{%
\pgfpathmoveto{\pgfqpoint{0.000000in}{0.034722in}}%
\pgfpathlineto{\pgfqpoint{-0.007796in}{0.010730in}}%
\pgfpathlineto{\pgfqpoint{-0.033023in}{0.010730in}}%
\pgfpathlineto{\pgfqpoint{-0.012614in}{-0.004098in}}%
\pgfpathlineto{\pgfqpoint{-0.020409in}{-0.028091in}}%
\pgfpathlineto{\pgfqpoint{-0.000000in}{-0.013263in}}%
\pgfpathlineto{\pgfqpoint{0.020409in}{-0.028091in}}%
\pgfpathlineto{\pgfqpoint{0.012614in}{-0.004098in}}%
\pgfpathlineto{\pgfqpoint{0.033023in}{0.010730in}}%
\pgfpathlineto{\pgfqpoint{0.007796in}{0.010730in}}%
\pgfpathlineto{\pgfqpoint{0.000000in}{0.034722in}}%
\pgfpathclose%
\pgfusepath{stroke,fill}%
}%
\begin{pgfscope}%
\pgfsys@transformshift{1.032477in}{1.766104in}%
\pgfsys@useobject{currentmarker}{}%
\end{pgfscope}%
\end{pgfscope}%
\begin{pgfscope}%
\pgfpathrectangle{\pgfqpoint{0.554481in}{0.501245in}}{\pgfqpoint{1.976074in}{1.551672in}}%
\pgfusepath{clip}%
\pgfsetbuttcap%
\pgfsetroundjoin%
\pgfsetlinewidth{1.505625pt}%
\definecolor{currentstroke}{rgb}{0.650980,0.337255,0.156863}%
\pgfsetstrokecolor{currentstroke}%
\pgfsetdash{{5.550000pt}{2.400000pt}}{0.000000pt}%
\pgfpathmoveto{\pgfqpoint{0.644303in}{0.564310in}}%
\pgfpathlineto{\pgfqpoint{0.662358in}{0.575715in}}%
\pgfpathlineto{\pgfqpoint{0.671385in}{0.581866in}}%
\pgfpathlineto{\pgfqpoint{0.680412in}{0.588365in}}%
\pgfpathlineto{\pgfqpoint{0.689439in}{0.595256in}}%
\pgfpathlineto{\pgfqpoint{0.698467in}{0.602585in}}%
\pgfpathlineto{\pgfqpoint{0.707494in}{0.610401in}}%
\pgfpathlineto{\pgfqpoint{0.716521in}{0.618761in}}%
\pgfpathlineto{\pgfqpoint{0.725549in}{0.627726in}}%
\pgfpathlineto{\pgfqpoint{0.734576in}{0.637363in}}%
\pgfpathlineto{\pgfqpoint{0.743603in}{0.647744in}}%
\pgfpathlineto{\pgfqpoint{0.752631in}{0.658953in}}%
\pgfpathlineto{\pgfqpoint{0.761658in}{0.671076in}}%
\pgfpathlineto{\pgfqpoint{0.770685in}{0.684213in}}%
\pgfpathlineto{\pgfqpoint{0.779712in}{0.698472in}}%
\pgfpathlineto{\pgfqpoint{0.788740in}{0.713970in}}%
\pgfpathlineto{\pgfqpoint{0.797767in}{0.730838in}}%
\pgfpathlineto{\pgfqpoint{0.806794in}{0.749219in}}%
\pgfpathlineto{\pgfqpoint{0.815822in}{0.769270in}}%
\pgfpathlineto{\pgfqpoint{0.824849in}{0.791160in}}%
\pgfpathlineto{\pgfqpoint{0.833876in}{0.815075in}}%
\pgfpathlineto{\pgfqpoint{0.842903in}{0.841215in}}%
\pgfpathlineto{\pgfqpoint{0.851931in}{0.869793in}}%
\pgfpathlineto{\pgfqpoint{0.860958in}{0.901034in}}%
\pgfpathlineto{\pgfqpoint{0.869985in}{0.935172in}}%
\pgfpathlineto{\pgfqpoint{0.879013in}{0.972442in}}%
\pgfpathlineto{\pgfqpoint{0.888040in}{1.013074in}}%
\pgfpathlineto{\pgfqpoint{0.897067in}{1.057274in}}%
\pgfpathlineto{\pgfqpoint{0.906094in}{1.105206in}}%
\pgfpathlineto{\pgfqpoint{0.915122in}{1.156961in}}%
\pgfpathlineto{\pgfqpoint{0.924149in}{1.212517in}}%
\pgfpathlineto{\pgfqpoint{0.933176in}{1.271681in}}%
\pgfpathlineto{\pgfqpoint{0.942204in}{1.334018in}}%
\pgfpathlineto{\pgfqpoint{0.978313in}{1.592982in}}%
\pgfpathlineto{\pgfqpoint{0.987340in}{1.650216in}}%
\pgfpathlineto{\pgfqpoint{0.996367in}{1.698808in}}%
\pgfpathlineto{\pgfqpoint{1.005395in}{1.735683in}}%
\pgfpathlineto{\pgfqpoint{1.014422in}{1.758400in}}%
\pgfpathlineto{\pgfqpoint{1.023449in}{1.765834in}}%
\pgfpathlineto{\pgfqpoint{1.032477in}{1.758721in}}%
\pgfpathlineto{\pgfqpoint{1.041504in}{1.739790in}}%
\pgfpathlineto{\pgfqpoint{1.050531in}{1.713312in}}%
\pgfpathlineto{\pgfqpoint{1.059558in}{1.684125in}}%
\pgfpathlineto{\pgfqpoint{1.068586in}{1.656511in}}%
\pgfpathlineto{\pgfqpoint{1.077613in}{1.633361in}}%
\pgfpathlineto{\pgfqpoint{1.086640in}{1.615906in}}%
\pgfpathlineto{\pgfqpoint{1.095668in}{1.603989in}}%
\pgfpathlineto{\pgfqpoint{1.104695in}{1.596615in}}%
\pgfpathlineto{\pgfqpoint{1.113722in}{1.592500in}}%
\pgfpathlineto{\pgfqpoint{1.122749in}{1.590460in}}%
\pgfpathlineto{\pgfqpoint{1.131777in}{1.589591in}}%
\pgfpathlineto{\pgfqpoint{1.149831in}{1.589248in}}%
\pgfpathlineto{\pgfqpoint{1.294268in}{1.589339in}}%
\pgfpathlineto{\pgfqpoint{2.440734in}{1.589339in}}%
\pgfpathlineto{\pgfqpoint{2.440734in}{1.589339in}}%
\pgfusepath{stroke}%
\end{pgfscope}%
\begin{pgfscope}%
\pgfpathrectangle{\pgfqpoint{0.554481in}{0.501245in}}{\pgfqpoint{1.976074in}{1.551672in}}%
\pgfusepath{clip}%
\pgfsetbuttcap%
\pgfsetbeveljoin%
\definecolor{currentfill}{rgb}{0.000000,0.000000,0.000000}%
\pgfsetfillcolor{currentfill}%
\pgfsetfillopacity{0.000000}%
\pgfsetlinewidth{0.501875pt}%
\definecolor{currentstroke}{rgb}{0.650980,0.337255,0.156863}%
\pgfsetstrokecolor{currentstroke}%
\pgfsetdash{}{0pt}%
\pgfsys@defobject{currentmarker}{\pgfqpoint{-0.033023in}{-0.028091in}}{\pgfqpoint{0.033023in}{0.034722in}}{%
\pgfpathmoveto{\pgfqpoint{0.000000in}{0.034722in}}%
\pgfpathlineto{\pgfqpoint{-0.007796in}{0.010730in}}%
\pgfpathlineto{\pgfqpoint{-0.033023in}{0.010730in}}%
\pgfpathlineto{\pgfqpoint{-0.012614in}{-0.004098in}}%
\pgfpathlineto{\pgfqpoint{-0.020409in}{-0.028091in}}%
\pgfpathlineto{\pgfqpoint{-0.000000in}{-0.013263in}}%
\pgfpathlineto{\pgfqpoint{0.020409in}{-0.028091in}}%
\pgfpathlineto{\pgfqpoint{0.012614in}{-0.004098in}}%
\pgfpathlineto{\pgfqpoint{0.033023in}{0.010730in}}%
\pgfpathlineto{\pgfqpoint{0.007796in}{0.010730in}}%
\pgfpathlineto{\pgfqpoint{0.000000in}{0.034722in}}%
\pgfpathclose%
\pgfusepath{stroke,fill}%
}%
\begin{pgfscope}%
\pgfsys@transformshift{1.023449in}{1.765834in}%
\pgfsys@useobject{currentmarker}{}%
\end{pgfscope}%
\end{pgfscope}%
\begin{pgfscope}%
\pgfpathrectangle{\pgfqpoint{0.554481in}{0.501245in}}{\pgfqpoint{1.976074in}{1.551672in}}%
\pgfusepath{clip}%
\pgfsetbuttcap%
\pgfsetroundjoin%
\pgfsetlinewidth{0.803000pt}%
\definecolor{currentstroke}{rgb}{1.000000,0.000000,0.000000}%
\pgfsetstrokecolor{currentstroke}%
\pgfsetdash{{2.960000pt}{1.280000pt}}{0.000000pt}%
\pgfpathmoveto{\pgfqpoint{0.554481in}{1.825608in}}%
\pgfpathlineto{\pgfqpoint{2.530555in}{1.825608in}}%
\pgfusepath{stroke}%
\end{pgfscope}%
\begin{pgfscope}%
\pgfsetrectcap%
\pgfsetmiterjoin%
\pgfsetlinewidth{0.803000pt}%
\definecolor{currentstroke}{rgb}{0.000000,0.000000,0.000000}%
\pgfsetstrokecolor{currentstroke}%
\pgfsetdash{}{0pt}%
\pgfpathmoveto{\pgfqpoint{0.554481in}{0.501245in}}%
\pgfpathlineto{\pgfqpoint{0.554481in}{2.052917in}}%
\pgfusepath{stroke}%
\end{pgfscope}%
\begin{pgfscope}%
\pgfsetrectcap%
\pgfsetmiterjoin%
\pgfsetlinewidth{0.803000pt}%
\definecolor{currentstroke}{rgb}{0.000000,0.000000,0.000000}%
\pgfsetstrokecolor{currentstroke}%
\pgfsetdash{}{0pt}%
\pgfpathmoveto{\pgfqpoint{2.530555in}{0.501245in}}%
\pgfpathlineto{\pgfqpoint{2.530555in}{2.052917in}}%
\pgfusepath{stroke}%
\end{pgfscope}%
\begin{pgfscope}%
\pgfsetrectcap%
\pgfsetmiterjoin%
\pgfsetlinewidth{0.803000pt}%
\definecolor{currentstroke}{rgb}{0.000000,0.000000,0.000000}%
\pgfsetstrokecolor{currentstroke}%
\pgfsetdash{}{0pt}%
\pgfpathmoveto{\pgfqpoint{0.554481in}{0.501245in}}%
\pgfpathlineto{\pgfqpoint{2.530555in}{0.501245in}}%
\pgfusepath{stroke}%
\end{pgfscope}%
\begin{pgfscope}%
\pgfsetrectcap%
\pgfsetmiterjoin%
\pgfsetlinewidth{0.803000pt}%
\definecolor{currentstroke}{rgb}{0.000000,0.000000,0.000000}%
\pgfsetstrokecolor{currentstroke}%
\pgfsetdash{}{0pt}%
\pgfpathmoveto{\pgfqpoint{0.554481in}{2.052917in}}%
\pgfpathlineto{\pgfqpoint{2.530555in}{2.052917in}}%
\pgfusepath{stroke}%
\end{pgfscope}%
\begin{pgfscope}%
\definecolor{textcolor}{rgb}{1.000000,0.000000,0.000000}%
\pgfsetstrokecolor{textcolor}%
\pgfsetfillcolor{textcolor}%
\pgftext[x=2.332948in,y=1.959817in,right,top]{\color{textcolor}\rmfamily\fontsize{8.330000}{9.996000}\selectfont \contour{white}{\ourmethod{}\(\displaystyle ^\star\)}}%
\end{pgfscope}%
\begin{pgfscope}%
\definecolor{textcolor}{rgb}{0.000000,0.000000,0.000000}%
\pgfsetstrokecolor{textcolor}%
\pgfsetfillcolor{textcolor}%
\pgftext[x=1.542518in,y=2.136251in,,base]{\color{textcolor}\rmfamily\fontsize{12.000000}{14.400000}\selectfont \taskname{all} RD-RSA}%
\end{pgfscope}%
\begin{pgfscope}%
\pgfsetbuttcap%
\pgfsetmiterjoin%
\definecolor{currentfill}{rgb}{1.000000,1.000000,1.000000}%
\pgfsetfillcolor{currentfill}%
\pgfsetfillopacity{0.800000}%
\pgfsetlinewidth{1.003750pt}%
\definecolor{currentstroke}{rgb}{0.800000,0.800000,0.800000}%
\pgfsetstrokecolor{currentstroke}%
\pgfsetstrokeopacity{0.800000}%
\pgfsetdash{}{0pt}%
\pgfpathmoveto{\pgfqpoint{1.001549in}{1.013907in}}%
\pgfpathlineto{\pgfqpoint{2.449569in}{1.013907in}}%
\pgfpathquadraticcurveto{\pgfqpoint{2.472708in}{1.013907in}}{\pgfqpoint{2.472708in}{1.037045in}}%
\pgfpathlineto{\pgfqpoint{2.472708in}{1.517117in}}%
\pgfpathquadraticcurveto{\pgfqpoint{2.472708in}{1.540256in}}{\pgfqpoint{2.449569in}{1.540256in}}%
\pgfpathlineto{\pgfqpoint{1.001549in}{1.540256in}}%
\pgfpathquadraticcurveto{\pgfqpoint{0.978410in}{1.540256in}}{\pgfqpoint{0.978410in}{1.517117in}}%
\pgfpathlineto{\pgfqpoint{0.978410in}{1.037045in}}%
\pgfpathquadraticcurveto{\pgfqpoint{0.978410in}{1.013907in}}{\pgfqpoint{1.001549in}{1.013907in}}%
\pgfpathlineto{\pgfqpoint{1.001549in}{1.013907in}}%
\pgfpathclose%
\pgfusepath{stroke,fill}%
\end{pgfscope}%
\begin{pgfscope}%
\pgfsetbuttcap%
\pgfsetroundjoin%
\pgfsetlinewidth{0.301125pt}%
\definecolor{currentstroke}{rgb}{0.215686,0.494118,0.721569}%
\pgfsetstrokecolor{currentstroke}%
\pgfsetdash{{1.110000pt}{0.480000pt}}{0.000000pt}%
\pgfpathmoveto{\pgfqpoint{1.024688in}{1.453485in}}%
\pgfpathlineto{\pgfqpoint{1.094104in}{1.453485in}}%
\pgfpathlineto{\pgfqpoint{1.163521in}{1.453485in}}%
\pgfusepath{stroke}%
\end{pgfscope}%
\begin{pgfscope}%
\definecolor{textcolor}{rgb}{0.000000,0.000000,0.000000}%
\pgfsetstrokecolor{textcolor}%
\pgfsetfillcolor{textcolor}%
\pgftext[x=1.256076in,y=1.412992in,left,base]{\color{textcolor}\rmfamily\fontsize{8.330000}{9.996000}\selectfont Liter. list.}%
\end{pgfscope}%
\begin{pgfscope}%
\pgfsetrectcap%
\pgfsetroundjoin%
\pgfsetlinewidth{0.602250pt}%
\definecolor{currentstroke}{rgb}{1.000000,0.498039,0.000000}%
\pgfsetstrokecolor{currentstroke}%
\pgfsetdash{}{0pt}%
\pgfpathmoveto{\pgfqpoint{1.024688in}{1.289605in}}%
\pgfpathlineto{\pgfqpoint{1.094104in}{1.289605in}}%
\pgfpathlineto{\pgfqpoint{1.163521in}{1.289605in}}%
\pgfusepath{stroke}%
\end{pgfscope}%
\begin{pgfscope}%
\definecolor{textcolor}{rgb}{0.000000,0.000000,0.000000}%
\pgfsetstrokecolor{textcolor}%
\pgfsetfillcolor{textcolor}%
\pgftext[x=1.256076in,y=1.249112in,left,base]{\color{textcolor}\rmfamily\fontsize{8.330000}{9.996000}\selectfont Depth 1}%
\end{pgfscope}%
\begin{pgfscope}%
\pgfsetbuttcap%
\pgfsetroundjoin%
\pgfsetlinewidth{0.903375pt}%
\definecolor{currentstroke}{rgb}{0.301961,0.686275,0.290196}%
\pgfsetstrokecolor{currentstroke}%
\pgfsetdash{{0.900000pt}{1.485000pt}}{0.000000pt}%
\pgfpathmoveto{\pgfqpoint{1.024688in}{1.125724in}}%
\pgfpathlineto{\pgfqpoint{1.094104in}{1.125724in}}%
\pgfpathlineto{\pgfqpoint{1.163521in}{1.125724in}}%
\pgfusepath{stroke}%
\end{pgfscope}%
\begin{pgfscope}%
\definecolor{textcolor}{rgb}{0.000000,0.000000,0.000000}%
\pgfsetstrokecolor{textcolor}%
\pgfsetfillcolor{textcolor}%
\pgftext[x=1.256076in,y=1.085231in,left,base]{\color{textcolor}\rmfamily\fontsize{8.330000}{9.996000}\selectfont Depth 2}%
\end{pgfscope}%
\begin{pgfscope}%
\pgfsetbuttcap%
\pgfsetroundjoin%
\pgfsetlinewidth{1.204500pt}%
\definecolor{currentstroke}{rgb}{0.968627,0.505882,0.749020}%
\pgfsetstrokecolor{currentstroke}%
\pgfsetdash{{7.680000pt}{1.920000pt}{1.200000pt}{1.920000pt}}{0.000000pt}%
\pgfpathmoveto{\pgfqpoint{1.825516in}{1.453485in}}%
\pgfpathlineto{\pgfqpoint{1.894933in}{1.453485in}}%
\pgfpathlineto{\pgfqpoint{1.964349in}{1.453485in}}%
\pgfusepath{stroke}%
\end{pgfscope}%
\begin{pgfscope}%
\definecolor{textcolor}{rgb}{0.000000,0.000000,0.000000}%
\pgfsetstrokecolor{textcolor}%
\pgfsetfillcolor{textcolor}%
\pgftext[x=2.056905in,y=1.412992in,left,base]{\color{textcolor}\rmfamily\fontsize{8.330000}{9.996000}\selectfont Depth 3}%
\end{pgfscope}%
\begin{pgfscope}%
\pgfsetbuttcap%
\pgfsetroundjoin%
\pgfsetlinewidth{1.505625pt}%
\definecolor{currentstroke}{rgb}{0.650980,0.337255,0.156863}%
\pgfsetstrokecolor{currentstroke}%
\pgfsetdash{{5.550000pt}{2.400000pt}}{0.000000pt}%
\pgfpathmoveto{\pgfqpoint{1.825516in}{1.289605in}}%
\pgfpathlineto{\pgfqpoint{1.894933in}{1.289605in}}%
\pgfpathlineto{\pgfqpoint{1.964349in}{1.289605in}}%
\pgfusepath{stroke}%
\end{pgfscope}%
\begin{pgfscope}%
\definecolor{textcolor}{rgb}{0.000000,0.000000,0.000000}%
\pgfsetstrokecolor{textcolor}%
\pgfsetfillcolor{textcolor}%
\pgftext[x=2.056905in,y=1.249112in,left,base]{\color{textcolor}\rmfamily\fontsize{8.330000}{9.996000}\selectfont Depth 4}%
\end{pgfscope}%
\end{pgfpicture}%
\makeatother%
\endgroup%

%% file: text/4_conclusion.tex
\section{Conclusion}
We have presented a model of pragmatic understanding based on equilibrium search called \ourmethod. In this model, speakers and listeners solve communicative tasks by searching for utterance-meaning mappings that that simultaneously optimize reward and similarity to a set of default meanings. \ourmethod offers a link between ``algorithmic'' models of pragmatic reasoning and equilibrium-based models, and accurately predicts human judgments across several pragmatic reasoning tasks.